%% file: main.tex
\documentclass[10pt,journal,compsoc]{IEEEtran}

\usepackage{graphicx}

\usepackage{float}
\usepackage{graphicx}
\usepackage{comment}
\usepackage{amsmath,amssymb} 
\usepackage{color}
\usepackage{xspace}
\usepackage{textcomp}

\usepackage{mathtools}
\usepackage{bbm}
\usepackage{makecell, verbatim, sidecap}
\usepackage{subfig}
\usepackage{multirow}
\usepackage{url,nicefrac}

\usepackage[nocompress]{cite}
\usepackage[margin=7pt,font=small,labelfont=bf,labelsep=endash,tableposition=bottom]{caption}

\usepackage{threeparttable}
\usepackage[vlined, ruled, linesnumbered]{algorithm2e}
\usepackage{multirow}
\usepackage{booktabs}
\usepackage{bbm}
\usepackage{pifont}
\usepackage{subfig}
\usepackage{graphicx}
\usepackage{amsmath}
\usepackage{enumitem}
\usepackage{eucal,nicefrac}
\usepackage{bm}
\usepackage{mathrsfs,xspace,xfrac}  
\usepackage{diagbox}
\usepackage[bottom]{footmisc}
\usepackage{xcolor,colortbl}

\renewcommand{\mathscr}[1]{ {{\mathcal #1} } }

\renewcommand{\vec}[1]{\ensuremath{\pmb{#1}}}
\newcommand{\mat}[1]{\ensuremath{\mathbf{#1}}}
\newcommand{\set}[1]{\ensuremath{\mathscr{#1}}}

\makeatletter
\@tfor\next:=abcdefghijklmnopqrstuvwxyxz\do
{\begingroup\edef\x{\endgroup
		\noexpand\@namedef{v\next}{\noexpand\vec{\next}}%
	}\x}
\@tfor\next:=ABCDEFGHIJKLMNOPQRSTUVWXYZ\do
{\begingroup\edef\x{\endgroup
		\noexpand\@namedef{m\next}{\noexpand\mat{\next}}%
	}\x}
\@tfor\next:=ABCDEFGHIJKLMNOPQRSTUVWXYZ\do
{\begingroup\edef\x{\endgroup
		\noexpand\@namedef{s\next}{\noexpand\set{\next}}%
	}\x}
\makeatother

\usepackage[pagebackref=true,breaklinks=true,letterpaper=true,colorlinks,bookmarks=false]{hyperref}
\newcommand{\etal}{\textit{et al.}\xspace}

\usepackage{xcolor}
\usepackage{soul}

\newcommand\red[1]{\textcolor{black}{#1}}
\newcommand\blue[1]{\textcolor{black}{#1}}

\begin{document}

\title{Metric3D v2: A Versatile Monocular Geometric Foundation Model for Zero-shot Metric Depth and Surface Normal Estimation}
\author{\vspace{-0.5em} Mu Hu$^{1 \ast}$,
        Wei Yin$^{2\ast \dag}$,
        Chi Zhang$^{3}$,
        Zhipeng Cai$^{4}$,
        Xiaoxiao Long$^{5 \ddag}$,  Hao Chen$^{6}$,\\
        Kaixuan Wang$^{1}$,
        Gang Yu$^{7}$,
        Chunhua Shen$^6$,
        Shaojie Shen$^1$ \vspace{-0.8em}
\IEEEcompsocitemizethanks{
\IEEEcompsocthanksitem 
$\ast$ Equal contribution.
\IEEEcompsocthanksitem 
$\dag$ WY is the project lead (yvanwy@outlook.com).
\IEEEcompsocthanksitem 
$\ddag$ XL is the corresponding author (xxlong@connect.hku.hk).
\IEEEcompsocthanksitem Contact MH for technical concerns (mhuam@connect.ust.hk).
\IEEEcompsocthanksitem $^1$ HKUST $^2$ Adelaide University 
\IEEEcompsocthanksitem $^3$ Westlake University $^4$ Intel $^5$ HKU $^6$ Zhejiang University $^7$ Tencent
}        
\thanks{
	Submitted for review on Feb. 29th, 2024.
}
 \\[0.152cm]
\small
\vspace{-0.0em}
}

\IEEEtitleabstractindextext{
\vspace{-0.5em}
\begin{abstract}
We introduce Metric3D v2, a geometric foundation model for zero-shot metric depth and surface normal estimation from a single image, which is crucial for metric 3D recovery. While depth and normal are geometrically related and highly complimentary, they present distinct challenges. State-of-the-art (SoTA) monocular depth methods achieve zero-shot generalization by learning affine-invariant depths, which cannot recover real-world metrics. Meanwhile, SoTA normal estimation methods have limited zero-shot performance due to the lack of large-scale labeled data. 
To tackle these issues, we propose solutions for both metric depth estimation and surface normal estimation. For metric depth estimation, we show that the key to a zero-shot single-view model lies in resolving the metric ambiguity from various camera models and large-scale data training. We propose a canonical camera space transformation module, which explicitly addresses the ambiguity problem and can be effortlessly plugged into existing monocular models. For surface normal estimation, we propose a joint depth-normal optimization module to distill diverse data knowledge from metric depth, enabling normal estimators to learn beyond normal labels. Equipped with these modules, our depth-normal models can be stably trained with over $16$ million of images from thousands of camera models with different-type annotations, resulting in zero-shot generalization to in-the-wild images with unseen camera settings. Our method currently\footnote{Feb. tbd,2024} ranks the 1st on various zero-shot and non-zero-shot benchmarks for metric depth, affine-invariant-depth as well as surface-normal prediction, shown in Fig. \ref{Fig: radar-sota}. Notably, we surpassed the ultra-recent MarigoldDepth and DepthAnything on various depth benchmarks including NYUv2 and KITTI. Our method enables the accurate recovery of metric 3D structures on randomly collected internet images, paving the way for plausible single-image metrology. The potential benefits extend to downstream tasks, which can be significantly improved by simply plugging in our model. For example, our model relieves the scale drift issues of monocular-SLAM (Fig.~\ref{Fig: first page fig.}), leading to high-quality metric scale dense mapping. These applications highlight the versatility of Metric3D v2 models as geometric foundation models. 
Our project page is at \url{https://JUGGHM.github.io/Metric3Dv2}. 

\end{abstract}

\begin{IEEEkeywords}
 Monocular metric depth estimation,  surface normal estimation, 3D scene shape estimation \vspace{-0.8em}
\end{IEEEkeywords}

}

\maketitle

\def\PWN{{\rm PWN}}
\def\VNL{{\rm VNL}}
\def\RPNL{{\rm RPNL}}

\vspace{-2em}
\section{Introduction}
\input{introduction}

\section{Related Work}
\input{related_works}

\section{Method}

\input{method}

\section{Experiments}
\input{exps}

\vspace{-0.5cm}

{\small
\bibliographystyle{ieeetr}
\bibliography{TPAMI}
}

\clearpage

\appendices

\pagenumbering{roman}
\setcounter{page}{1}

\setcounter{table}{0}
\renewcommand{\thetable}{A\arabic{table}}

\setcounter{figure}{0}
\renewcommand{\thefigure}{A\arabic{figure}}

\end{document}


\title{Supplementary Materials for Metric3D v2: A Versatile Monocular Geometric Foundation Model for Zero-shot Metric Depth and Surface Normal Estimation 
}

\maketitle

\section{Details for Models}
\noindent\textbf{Details for ConvNet models.} In our work, our encoder employs the ConvNext \cite{liu2022convnet} networks, whose pretrained weight is from the official released ImageNet-22k pretraining. The decoder follows the adabins~\cite{bhat2021adabins}. We set the depth bins number to 256, and the depth range is $[0.3m, 150m]$. We establish 4 flip connections from different levels of encoder blocks to the decoder to merge more low-level features. An hourglass subnetwork is attached to the head of the decoders to enhance background predictions.  

\noindent\textbf{Details for ViT models.} We use dino-v2 transformers \cite{oquab2023dinov2} with registers \cite{darcet2023vision} as our encoders, which are pretrained on a curated dataset with 142M images. DPT \cite{ranftl2021vision} is used as the decoders. For the ViT-S and ViT-L variants, the DPT decoders take only the last-layer normalized encoder features as the input for stabilized training. The giant ViT-g model instead takes varying-layer features, the same as the original DPT settings. Different from the convnets models above, we use depth bins ranging from $[0.1m, 200m]$ for ViT models.  

\begin{figure}[!b]
\centering
\includegraphics[width=0.45\textwidth]{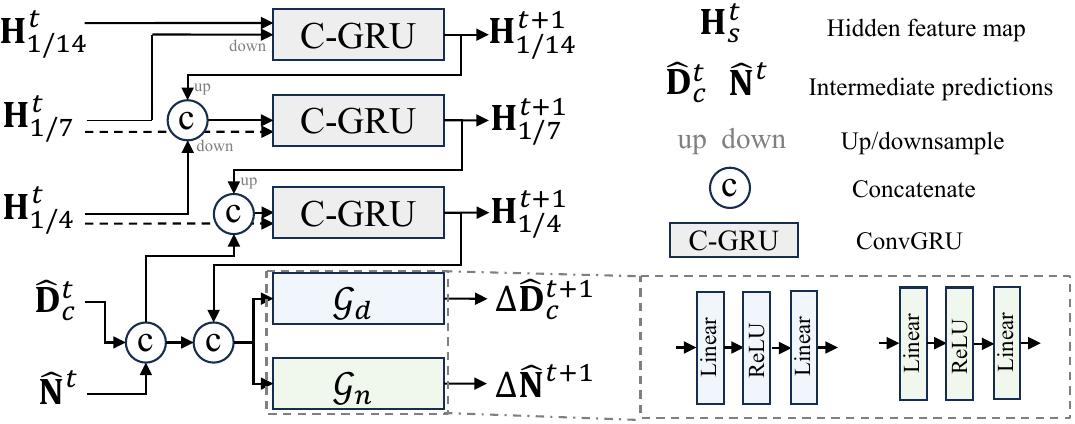}
\vspace{-0.5 em}
\caption{\textbf{Detailed structure of the update block}. Inspired by RAFTStereo \cite{lipson2021raft}, we build slow-fast ConvGRU sub-blocks (denoted as `C-GRU') to refine hierarchical hidden feature maps. Projection heads $\mathcal{G}_d, \mathcal{G}_n$ are attached to the end of the final ConvGRU to prediction update items for the predictions.}
\label{fig: update}
\vspace{0em}
\end{figure}

\noindent\textbf{Details for recurrent blocks.} As illustrated in 
Fig~\ref{fig: update}. Each recurrent block updates hierarchical features maps $\{\mathbf{H}^{t}_{1/14}, \mathbf{H}^{t}_{1/7}, \mathbf{H}^{t}_{1/4}\}$ at $\{\frac{1}{14}, \frac{1}{7}, \frac{1}{4}\}$ scales and the intermediate predictions s$\hat{\mathbf{D}}_c^t$, $\hat{\mathbf{N}}^t$ at each iteration step $t$. This block compromises three ConvGRU sub-blocks to refine feature maps at different scales, and two projection heads $\mathcal{G}_d$ and $\mathcal{G}_n$ to predict updates for depth and normal respectively. The feature maps are gradually refined from the coarsest ($\mathbf{H}^{t}_{1/14}$) to the finest ($\mathbf{H}^{t}_{1/4}$). For instance, the refined feature map at the $\frac{1}{14}$ scale $\mathbf{H}^{t+1}_{1/14}$ is fed into the second ConvGRU sub-block to refine the $\frac{1}{7}$-scale feature map $\mathbf{H}^{t}_{1/7}$. Finally, the projection heads $\mathcal{G}_d, \mathcal{G}_n$ employs a concatenation of original predictions $\hat{\mathbf{D}}_c^t$, $\hat{\mathbf{N}}^t$ and the to finest feature map $\mathbf{H}^{t}_{1/4}$ to predict the update items $\Delta\hat{\mathbf{D}}_c^{t+1}$, $\Delta\hat{\mathbf{N}}^{t+1}$. Both projection heads are composed of two linear layers with a sandwiched ReLU activation layer.

\noindent\textbf{Resource comparison of different models.} We compare the resource and performance among our model families in Tab.~\ref{Table: family.}. All inference-time and GPU memory results are computed on an Nvida-A100 40G GPU with the original pytorch implemented models (No engineering optimization like TensorRT or ONNX). Generally, the enormous ViT-Large/giant-backbone models enjoy better performance, while the others are more deployment-friendly. In addition, our models built in classical en-decoder schemes run much faster than the recent diffusion counterpart \cite{ke2023repurposing}.

\begin{table*}[t]
\centering
\caption{
Comparative analysis of resource and performance across our model families includes evaluation of resource utilization metrics such as inference speed, memory usage, and the proportion of optimization modules. Additionally, we assess metric depth performance on KITTI/NYU datasets and normal performance on NYUv2 dataset, with results derived from checkpoints without fine-tuning. For ViT models, inference speed is measured using 16-bit precision (Bfloat16), which is the same precision as the training setup.
}
\vspace{0.2 em}
\setlength{\tabcolsep}{2pt}
\resizebox{0.9\linewidth}{!}{
\begin{tabular}{ ccc |ccc|ll|ll|ll}
\toprule[1pt]
\multicolumn{3}{c|}{Model} &  \multicolumn{3}{c|}{Resource} & \multicolumn{2}{c|}{KITTI Depth} & \multicolumn{2}{c|}{NYUv2 Depth} & \multicolumn{2}{c}{NYUv2 Normal} \\
 Encoder & Decoder & Optim. & Speed & GPU Memory & Optim. time & AbsRel$\downarrow$     & $\delta_{1}$$\uparrow$     & AbsRel$\downarrow$      & $\delta_{1}$$\uparrow$      & Median$\downarrow$      & $30^\circ$$\uparrow$        \\ \hline
Marigold\cite{ke2023repurposing} VAE+U-net & U-net+VAE & - 
& 0.13 fps & 17.3G & -
& No metric &No metric 
&No metric &No metric  
& - & -   \\ \hline
Ours ConvNeXt-Large & Hourglass & - 
& 10.5 fps & 4.2G & -
&0.053 &0.965 
&0.083 &0.944  
& - & -   \\ 
Ours ViT-Small & DPT & 4 steps 
& 11.6 fps & 2.9G  & 3.4\%
&0.070 &0.937
&0.084 &0.945
& 7.7 & 0.870   \\
Ours ViT-Large & DPT & 8 steps  
& 9.5 fps & 7.0G & 9.5\%
&0.052 &0.974
&0.063 &0.975  
& 7.0 & 0.881  \\
Ours ViT-giant & DPT & 8 steps 
& 5.0 fps & 15.6G & 25\%
&0.051 &0.977
&0.067 &0.980 
& 7.1 & 0.881   \\
 \toprule[1pt]
\end{tabular}}
\label{Table: family.}
\vspace{-1.5em}
\end{table*}

\begin{table}[!b]
\vspace{-2 em}
	\caption{Training and testing datasets used for experiments.}
	\centering
	\begin{threeparttable}
		\scalebox{0.5}{
			\begin{tabular}{ r llllll}
				\toprule[1pt]
				\multicolumn{1}{l|}{Datasets}                     & \multicolumn{1}{l|}{Scenes}    & \multicolumn{1}{l|}{Source}     & \multicolumn{1}{l|}{%
					Label}     &     \multicolumn{1}{l|}{Size}     & \#Cam.          \\ \hline
				\multicolumn{6}{c}{Training Data}                                                                                                                                                          \\ \hline
				\multicolumn{1}{l|}{DDAD~\cite{packnet}}                           & \multicolumn{1}{l|}{Outdoor}     &  \multicolumn{1}{l|}{Real-world}      &  \multicolumn{1}{l|}{Depth}                 & \multicolumn{1}{l|}{$\sim$80K}     & 36+           \\
				\multicolumn{1}{l|}{Lyft~\cite{lyftl5preception}}                           & \multicolumn{1}{l|}{Outdoor}        & 
				\multicolumn{1}{l|}{Real-world}      &  \multicolumn{1}{l|}{Depth}                 & \multicolumn{1}{l|}{$\sim$50K} & 6+              \\
				\multicolumn{1}{l|}{Driving Stereo (DS)~\cite{yang2019drivingstereo}}                 & \multicolumn{1}{l|}{Outdoor}        & \multicolumn{1}{l|}{Real-world}      & \multicolumn{1}{l|}{Depth}       & \multicolumn{1}{l|}{$\sim$181K}  & 1              \\
				\multicolumn{1}{l|}{DIML~\cite{cho2021diml}}                           & \multicolumn{1}{l|}{Outdoor}        & \multicolumn{1}{l|}{Real-world}      & \multicolumn{1}{l|}{Depth}       & \multicolumn{1}{l|}{$\sim$122K}       & 10        \\
				\multicolumn{1}{l|}{Arogoverse2~\cite{Argoverse2}}                    & \multicolumn{1}{l|}{Outdoor}        & \multicolumn{1}{l|}{Real-world}      & \multicolumn{1}{l|}{Depth}                 & \multicolumn{1}{l|}{$\sim$3515K}  & 6+           \\
				\multicolumn{1}{l|}{Cityscapes~\cite{Cordts2016Cityscapes}}                     & \multicolumn{1}{l|}{Outdoor}        & \multicolumn{1}{l|}{Real-world}      & \multicolumn{1}{l|}{Depth}       & \multicolumn{1}{l|}{$\sim$170K}      & 1         \\
				\multicolumn{1}{l|}{DSEC~\cite{Gehrig21ral}}                           & \multicolumn{1}{l|}{Outdoor}        & \multicolumn{1}{l|}{Real-world}      & \multicolumn{1}{l|}{Depth}                 & \multicolumn{1}{l|}{$\sim$26K} & 1                \\
				\multicolumn{1}{l|}{Mapillary PSD~\cite{MapillaryPSD}} & \multicolumn{1}{l|}{Outdoor}   &  \multicolumn{1}{l|}{Real-world}      & \multicolumn{1}{l|}{Depth} & \multicolumn{1}{l|}{750K} & 1000+               \\
				\multicolumn{1}{l|}{Pandaset~\cite{itsc21pandaset}}                       & \multicolumn{1}{l|}{Outdoor}        & \multicolumn{1}{l|}{Real-world}      & \multicolumn{1}{l|}{Depth}   & \multicolumn{1}{l|}{$\sim$48K} & 6                      \\
				\multicolumn{1}{l|}{UASOL~\cite{bauer2019uasol}}                          & \multicolumn{1}{l|}{Outdoor}        & \multicolumn{1}{l|}{Real-world}      & \multicolumn{1}{l|}{Depth}       & \multicolumn{1}{l|}{$\sim$1370K}          & 1             \\
				\multicolumn{1}{l|}{Virtual KITTI~\cite{cabon2020virtual}}                      & \multicolumn{1}{l|}{Outdoor}         & \multicolumn{1}{l|}{Synthesized}      & \multicolumn{1}{l|}{Depth}                 & \multicolumn{1}{l|}{37K}  & 2   \\
				\multicolumn{1}{l|}{Waymo~\cite{sun2020scalability}}                      & \multicolumn{1}{l|}{Outdoor}         & \multicolumn{1}{l|}{Real-world}      & \multicolumn{1}{l|}{Depth}                 & 				\multicolumn{1}{l|}{$\sim${1M}} & 5 \\
				\multicolumn{1}{l|}{Matterport3d~\cite{zamir2018taskonomy}}                      & \multicolumn{1}{l|}{In/Out}         & \multicolumn{1}{l|}{Real-world}      & \multicolumn{1}{l|}{Depth + Normal}                 & \multicolumn{1}{l|}{144K}  & 3 \\
				\multicolumn{1}{l|}{Taskonomy~\cite{zamir2018taskonomy}}                      & \multicolumn{1}{l|}{Indoor}         & \multicolumn{1}{l|}{Real-world}      & \multicolumn{1}{l|}{Depth + Normal}                 & \multicolumn{1}{l|}{$\sim$4M} & $\sim$1M                \\
				\multicolumn{1}{l|}{Replica~\cite{straub2019replica}}                      & \multicolumn{1}{l|}{Indoor}         & \multicolumn{1}{l|}{Real-world}      & \multicolumn{1}{l|}{Depth + Normal}                 & \multicolumn{1}{l|}{$\sim$150K} & 1 \\
				\multicolumn{1}{l|}{ScanNet\tnote{\dag}~\cite{dai2017scannet}}                      &   \multicolumn{1}{l|}{Indoor}         & \multicolumn{1}{l|}{Real-world}      & \multicolumn{1}{l|}{Depth + Normal}                 & \multicolumn{1}{l|}{$\sim$2.5M} & 1 \\
				\multicolumn{1}{l|}{HM3d~\cite{ramakrishnan2021habitat}}                      & \multicolumn{1}{l|}{Indoor}         & \multicolumn{1}{l|}{Real-world}      & \multicolumn{1}{l|}{Depth + Normal}                 &\multicolumn{1}{l|}{$\sim$2000K}  & 1
				\\
				\multicolumn{1}{l|}{Hypersim~\cite{roberts2021hypersim}}                      & \multicolumn{1}{l|}{Indoor}         & \multicolumn{1}{l|}{Synthesized}      & \multicolumn{1}{l|}{Depth + Normal}                 &\multicolumn{1}{l|}{54K} & 1 				
				\\ \hline
				\multicolumn{5}{c}{Testing Data}                                                                                                                                                           \\ \hline
				  \multicolumn{1}{l|}{NYU~\cite{silberman2012indoor}}                            & \multicolumn{1}{l|}{Indoor}      & \multicolumn{1}{l|}{Real-world}         & \multicolumn{1}{l|}{Depth+Normal}                & \multicolumn{1}{l|}{654} & 1                      \\
				\multicolumn{1}{l|}{KITTI~\cite{Geiger2013IJRR}}                          & \multicolumn{1}{l|}{Outdoor}        & \multicolumn{1}{l|}{Real-world}      & \multicolumn{1}{l|}{Depth}                 & \multicolumn{1}{l|}{652} & 4                   \\
				\multicolumn{1}{l|}{ScanNet\tnote{\dag}~\cite{dai2017scannet}}                        & \multicolumn{1}{l|}{Indoor}         & \multicolumn{1}{l|}{Real-world}      & \multicolumn{1}{l|}{Depth+Normal}                & \multicolumn{1}{l|}{700} & 1                      \\
				\multicolumn{1}{l|}{NuScenes (NS)~\cite{caesar2020nuscenes}}                       & \multicolumn{1}{l|}{Outdoor}        & \multicolumn{1}{l|}{Real-world}      & \multicolumn{1}{l|}{Depth}                 & \multicolumn{1}{l|}{10K} & 6                        \\
				\multicolumn{1}{l|}{ETH3D~\cite{schops2017multi}}                          & \multicolumn{1}{l|}{Outdoor}        & \multicolumn{1}{l|}{Real-world}      & \multicolumn{1}{l|}{Depth}                 & \multicolumn{1}{l|}{431} & 1                        \\
				\multicolumn{1}{l|}{DIODE~\cite{vasiljevic2019diode}}                          & \multicolumn{1}{l|}{In/Out} & \multicolumn{1}{l|}{Real-world}      & \multicolumn{1}{l|}{Depth}                 & \multicolumn{1}{l|}{771} & 1                     \\
				\multicolumn{1}{l|}{iBims-1~\cite{koch2018evaluation}}      & \multicolumn{1}{l|}{Indoor}                        & \multicolumn{1}{l|}{Real-world}         & \multicolumn{1}{l|}{Depth}                & \multicolumn{1}{l|}{100} & 1                        \\
				\toprule[1pt]
		\end{tabular}}
		\begin{tablenotes}
		\scriptsize
		\item[\dag] ScanNet is a non-zero-shot testing dataset for our ViT models.
		\end{tablenotes}
		\vspace{-0.2 em}
	\end{threeparttable}
	\label{table: datasetsv2}
\end{table}

\subsection{Datasets and Training and Testing}
We collect over $16$M data from 18 public datasets for training. Datasets are listed in Tab.~\ref{table: datasetsv2}. When training the ConvNeXt-backbone models, we use a smaller collection containing the following 11 datasets with $8$M images: DDAD~\cite{packnet}, Lyft~\cite{lyftl5preception}, DrivingStereo~\cite{yang2019drivingstereo}, DIML~\cite{cho2021diml}, Argoverse2~\cite{Argoverse2}, Cityscapes~\cite{Cordts2016Cityscapes}, DSEC~\cite{Gehrig21ral}, Maplillary PSD~\cite{MapillaryPSD}, Pandaset~\cite{itsc21pandaset}, UASOL\cite{bauer2019uasol}, and Taskonomy~\cite{zamir2018taskonomy}. In the autonomous driving datasets, including DDAD~\cite{packnet}, Lyft~\cite{lyftl5preception}, DrivingStereo~\cite{yang2019drivingstereo}, Argoverse2~\cite{Argoverse2}, DSEC~\cite{Gehrig21ral}, and Pandaset~\cite{itsc21pandaset}, have provided LiDar and camera intrinsic and extrinsic parameters. We project the LiDar to image planes to obtain ground-truth depths. In contrast, Cityscapes~\cite{Cordts2016Cityscapes}, DIML~\cite{cho2021diml}, and UASOL~\cite{bauer2019uasol} only provide calibrated stereo images. We use raftstereo~\cite{lipson2021raft} to achieve pseudo ground-truth depths. Mapillary PSD~\cite{MapillaryPSD} dataset provides paired RGB-D, but the depth maps are achieved from a structure-from-motion method. The camera intrinsic parameters are estimated from the SfM. We believe that such achieved metric information is noisy. Thus we do not enforce learning-metric-depth loss on this data, \textit{i.e.},  $L_{silog}$, to reduce the effect of noises. For the Taskonomy~\cite{zamir2018taskonomy} dataset, we follow LeReS~\cite{yin2022towards} to obtain the instance planes, which are employed in the pair-wise normal regression loss. During training, we employ the training strategy from \cite{yin2020diversedepth_old} to balance all datasets in each training batch.

The testing data is listed in Tab.~\ref{table: datasetsv2}. All of them are captured by high-quality sensors. In testing, we employ their provided camera intrinsic parameters to perform our proposed canonical space transformation.

\subsection{Details for Some Experiments}
\noindent\textbf{Finetuning protocols.} To finetune the large-scale-data trained models on some specific datasets, we use the ADAM optimizer with the initial learning rate beginning at $10^{-6}$ and linear decayed to $10^{-7}$ within $6$K steps. Notably, such finetune does not require a large batch size like large-scale training. We use in practice a batch-size of $16$ for the ViT-g model and $32$ for ViT-L. The models will converge quickly in approximately $2$K steps. To stabilize finetuning, we also leverage the predictions $\tilde{\mathbf{D}}_c$ $\tilde{\mathbf{N}}$ of the pre-finetuned model as alternative pseudo labels. These labels can sufficiently impose supervision upon the annotation-absent regions. The complete loss can be formulated as:
\begin{equation}
\begin{split}
    L_{ft} = 0.01L_{d-n}(\mathbf{D}_c, \mathbf{N}) + L_d(\mathbf{D}_c, \mathbf{D}^\ast_c) + L_n(\mathbf{N}, \mathbf{N^\ast}) \\
    + 0.01(L_d(\mathbf{D}_c, \tilde{\mathbf{D}}_c) + L_n(\mathbf{N}, \tilde{\mathbf{N}})),
\end{split}
\end{equation}
where $\mathbf{D}_c$ and $\mathbf{N}$ are the predicted depth in the canonical space and surface normal, $\mathbf{D}^\ast_c$ and $\mathbf{N}^\ast$ are the groundtruth labels, $L_d$, $L_n$, and $L_{d-n}$ are the losses for depth, normal, and depth-normal consistency introduced in the main text.

\noindent\textbf{Evaluation of zero-shot 3D scene reconstruction.} In this experiment, we use all methods' released models to predict each frame's depth and use the ground-truth poses and camera intrinsic parameters to reconstruct point clouds. When evaluating the reconstructed point cloud, we employ the iterative closest point (ICP)~\cite{besl1992method} algorithm to match the predicted point clouds with ground truth by a pose transformation matrix. Finally, we evaluate the Chamfer $\ell_1$ distance and F-score on the point cloud.

\begin{figure*}[]
\centering
\includegraphics[width=0.8\textwidth]{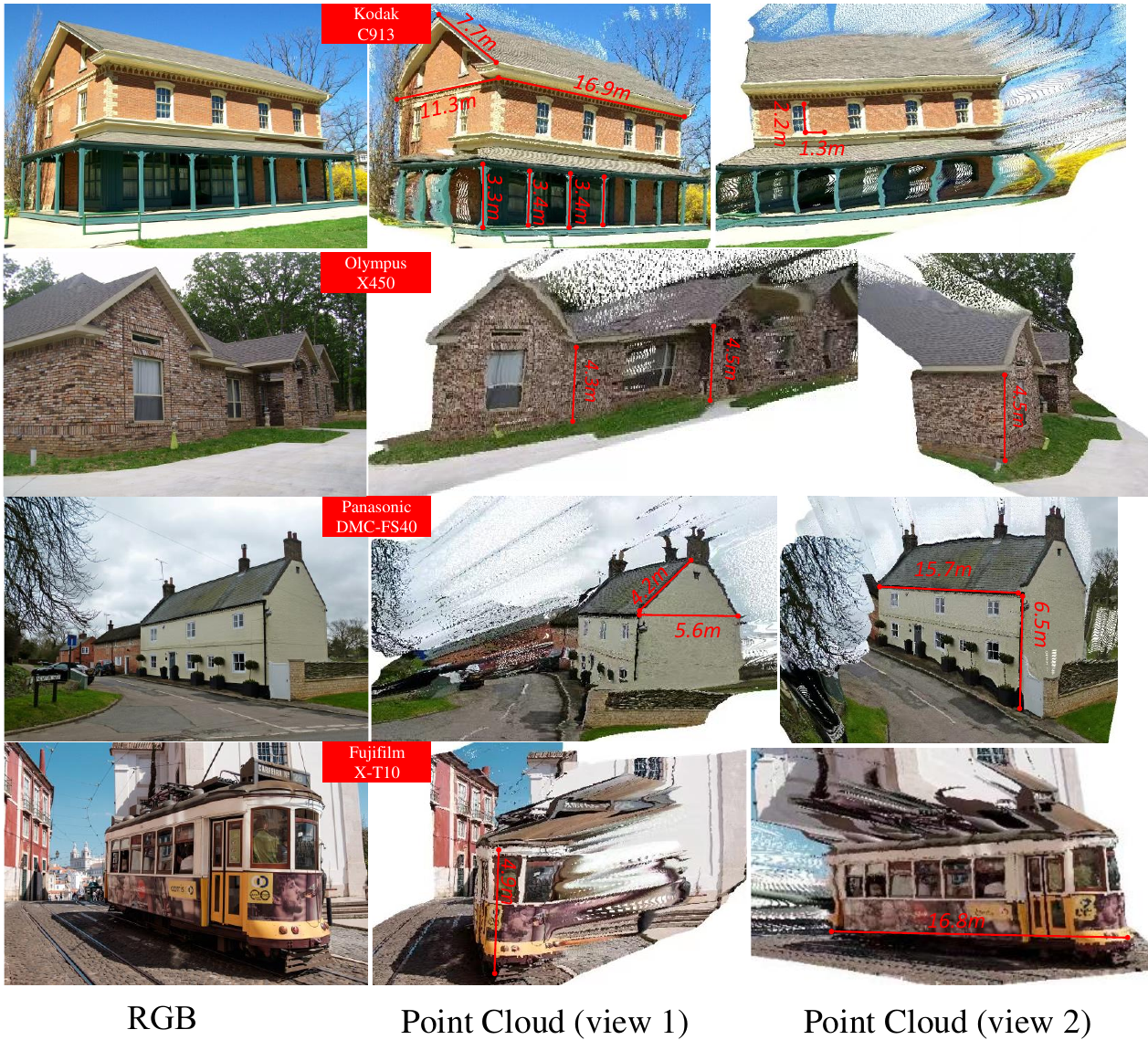}
\caption{\textbf{3D metric reconstruction of in-the-wild images.} We collect several Flickr images and use our model to reconstruct the scene. The focal length information is collected from the photo's metadata. From the reconstructed point cloud, we can measure some structures' sizes. We can observe that sizes are in a reasonable range.}
\label{fig: recon in the wild.}
\end{figure*}

\noindent\textbf{Reconstruction of in-the-wild scenes.} We collect several photos from Flickr. From their associated camera metadata, we can obtain the focal length $\hat{f}$ and the pixel size $\delta$. According to $\nicefrac{\hat{f}}{\delta}$, we can obtain the pixel-represented focal length for 3D reconstruction and achieve the metric information. We use meshlab software to measure some structures' size on point clouds. More visual results are shown in Fig. \ref{fig: recon in the wild.}.

\noindent\textbf{Generalization of metric depth estimation.} To evaluate our method's robustness of metric recovery, we test on 7 zero-shot datasets, i.e. NYU, KITTI, DIODE (indoor and outdoor parts), ETH3D, iBims-1, and NuScenes. Details are reported in Tab.~\ref{table: datasetsv2}.  We use the officially provided focal length to predict the metric depths. All benchmarks use the same depth model for evaluation. We don't perform any scale alignment.  

\noindent\textbf{Evaluation on affine-invariant depth benchmarks.} We follow existing affine-invariant depth estimation methods to evaluate 5 zero-shot datasets. Before evaluation, we employ the least square fitting to align the scale and shift with ground truth~\cite{leres}. Previous methods' performance is cited from their papers. 

\begin{figure*}[]
\centering
\includegraphics[width=0.95\textwidth]{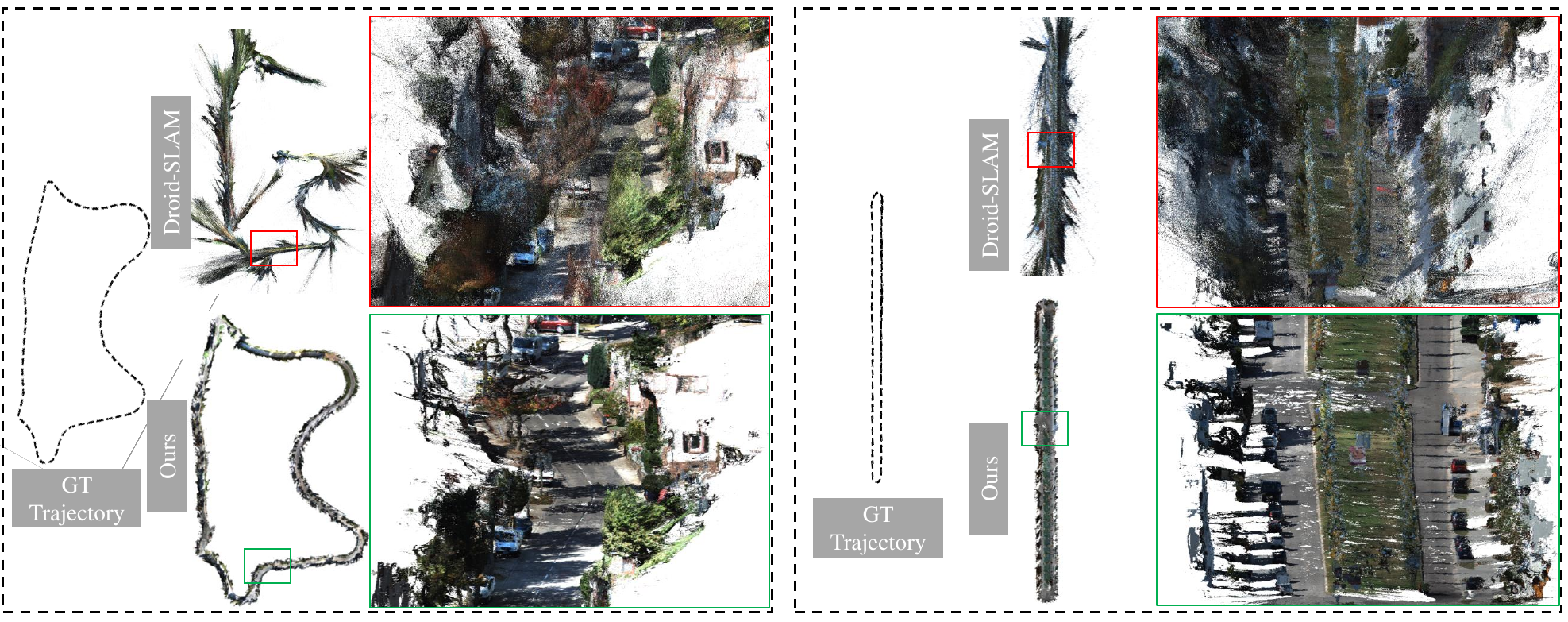}
\includegraphics[width=0.95\textwidth]{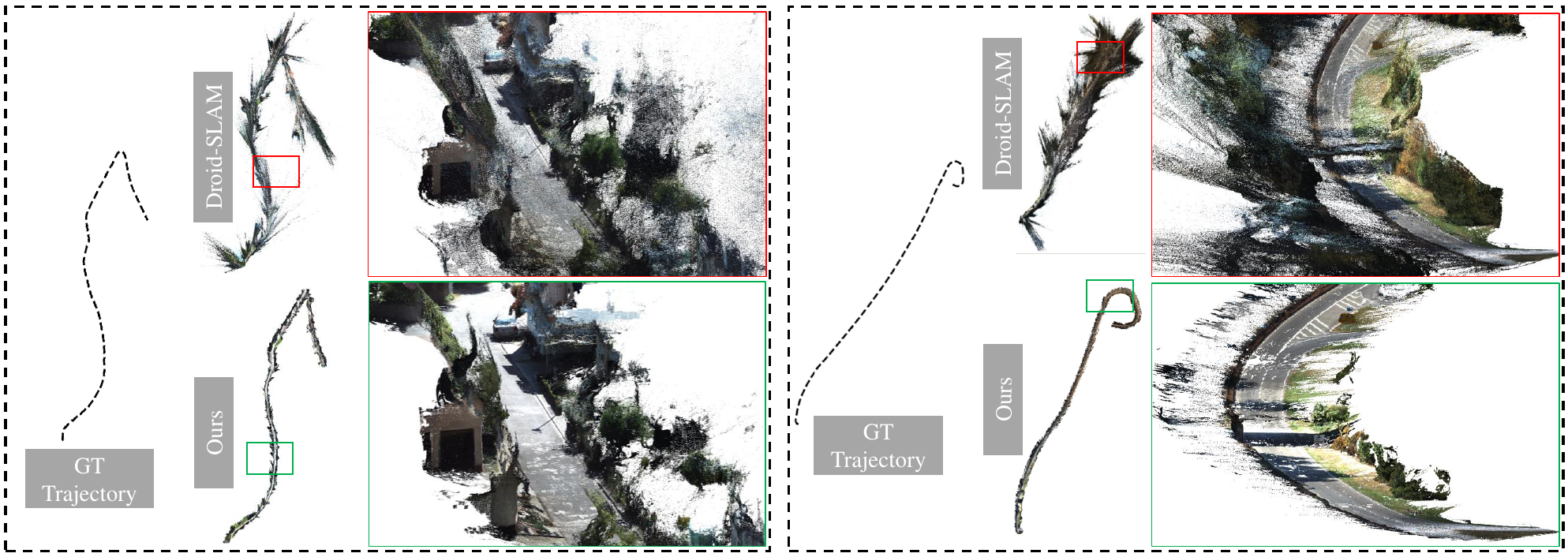}
\caption{\textbf{Dense-SLAM Mapping.} Existing SOTA mono-SLAM methods usually face scale drift problems in large-scale scenes and are unable to achieve the metric scale. We show the ground-truth trajectory and Droid-SLAM~\cite{teed2021droid} predicted trajectory and their dense mapping. Then, we naively input our metric depth to Droid-SLAM, which can recover a much more accurate trajectory and perform the \textit{metric} dense mapping.}
\label{fig: dense_slam1.}
\end{figure*}

\noindent\textbf{Dense-SLAM Mapping.} This experiment is conducted on the KITTI odometry benchmark. We use our model to predict metric depths, and then naively input them to the Droid-SLAM system as an initial depth. We do not perform any finetuning but directly run their released codes on KITTI. With Droid-SLAM predicted poses, we unproject depths to the 3D point clouds and fuse them together to achieve dense metric mapping. More qualitative results are shown in Fig.~\ref{fig: dense_slam1.}.

\begin{figure}[!h]
	\centering
	\includegraphics[width=0.48\textwidth]{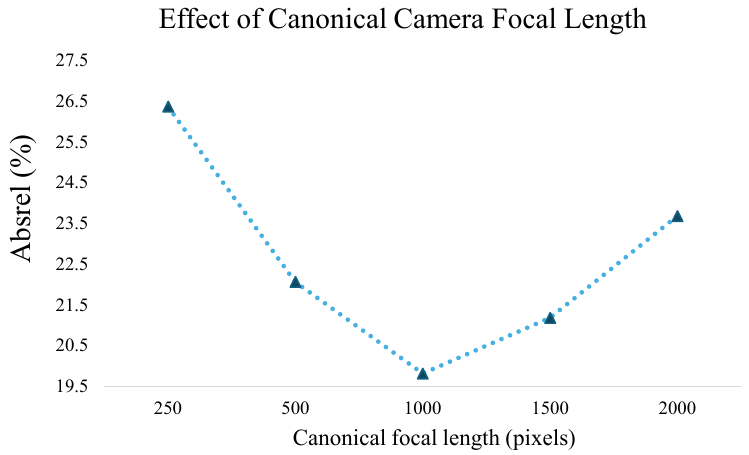}
	\vspace{-0.5 em}
	\caption{\red{\textbf{Effect of varying canonical focal lengths.} We
		apply different canonical focal lengths and find that an intermediate focal length leads to the best performance.}}
	\label{fig: focal}
	\vspace{-0.5em}
\end{figure}

\noindent\red{\textbf{Ablation on canonical space.} To study the effect of different focal lengths in the canonical space, we train the ConvNext models on the small subset of varying datasets and test their performance on the validation set. We compare the absrel error using \{250, 500, 1000, 1500, and 2000\}-pixel canonical camera focal lengths. As shown in Fig. \ref{fig: focal}, the canonical focal length of 1000 achieves the lowest depth error. We apply this setting for all other experiments.}

\subsection{More Visual Results}
\begin{figure*}[]
\centering
\includegraphics[width=0.9\textwidth]{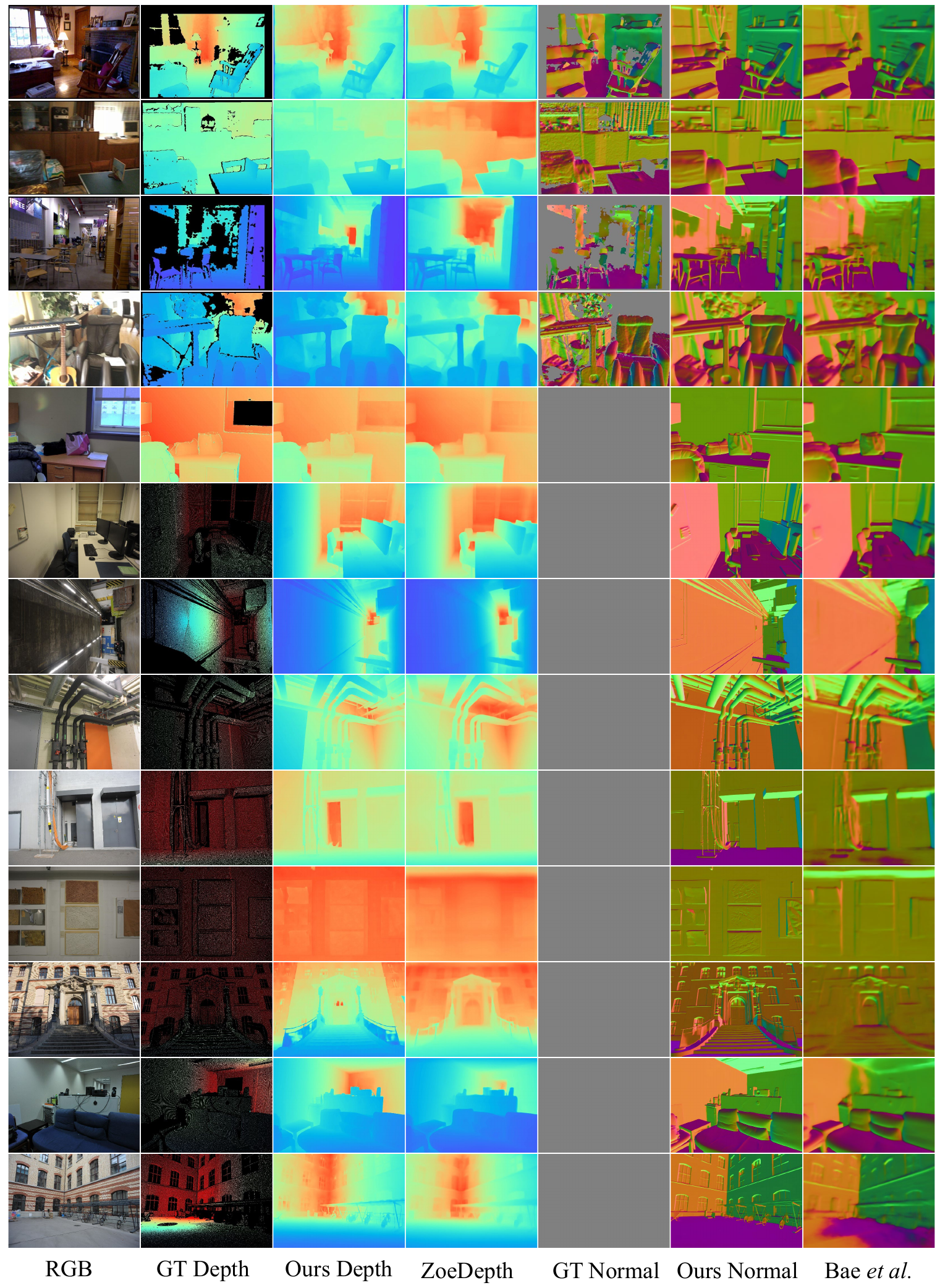}
\caption{\textbf{Depth and normal estimation.} The visual comparison of predicted depth and normal on indoor/outdoor scenes from NYUv2, iBims, Eth3d, and ScanNet. Our depth and normal maps come from the ViT-g CSTM\_label model.}
\label{fig: dn_cmp1.}
\end{figure*}

\begin{figure*}[]
\centering
\includegraphics[width=0.9\textwidth]{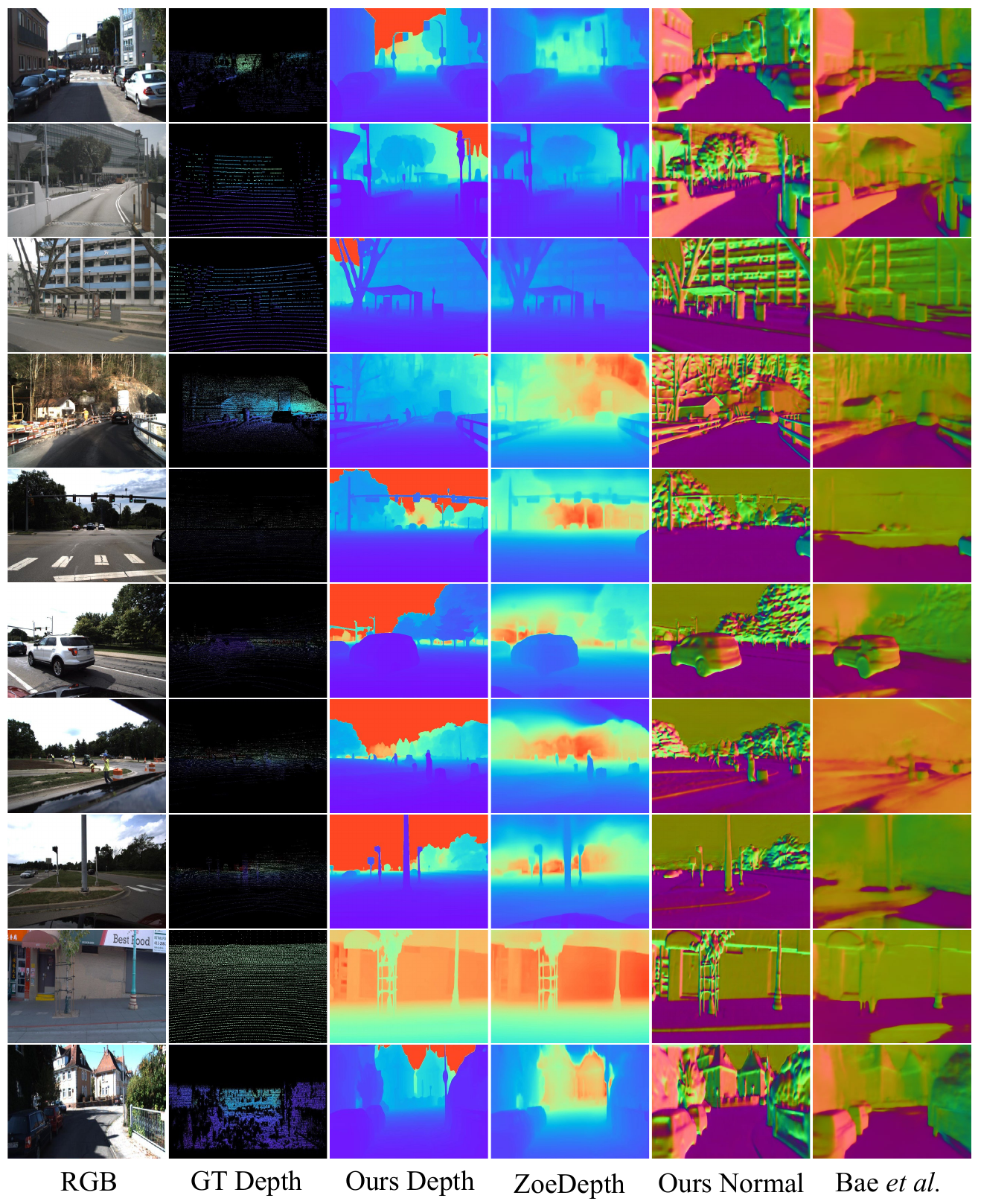}
\caption{\textbf{Depth and normal estimation.} The visual comparison of predicted depth and normal on driving scenes from KITTI, Nuscenes, DIML, DDAD, and Waymo. Our depth and normal maps come from the ViT-g CSTM\_label model.}
\label{fig: dn_cmp2.}
\end{figure*}

\begin{figure*}[]
\centering
\includegraphics[width=1\textwidth]{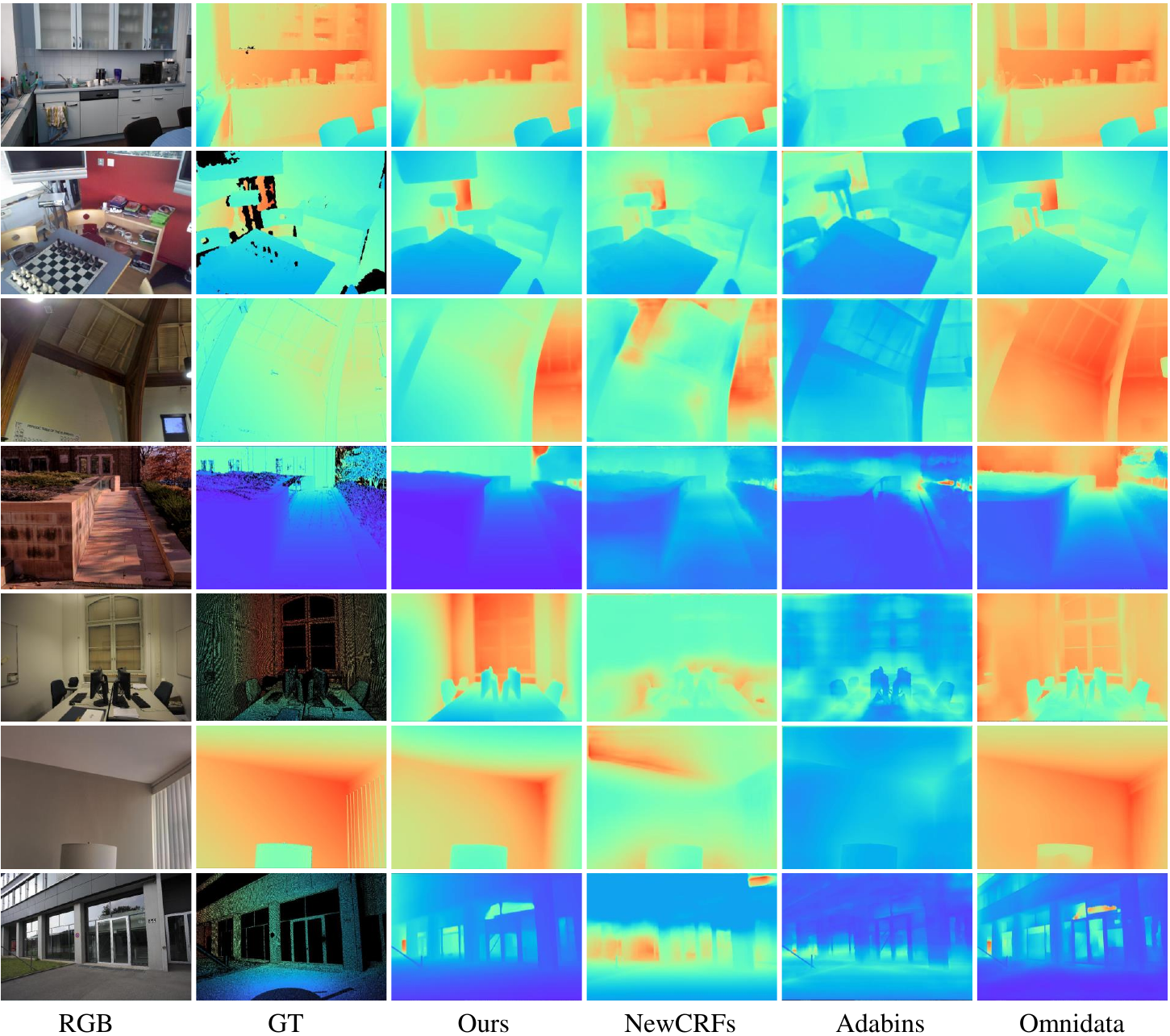}
\caption{The visual comparison of predicted depth on iBims, ETH3D, and DIODE. Our depth maps come from the ConvNeXt-L CSTM\_label model.}
\label{fig: depth_cmp1.}
\end{figure*}

\begin{figure*}[]
\centering
\includegraphics[width=0.9\textwidth]{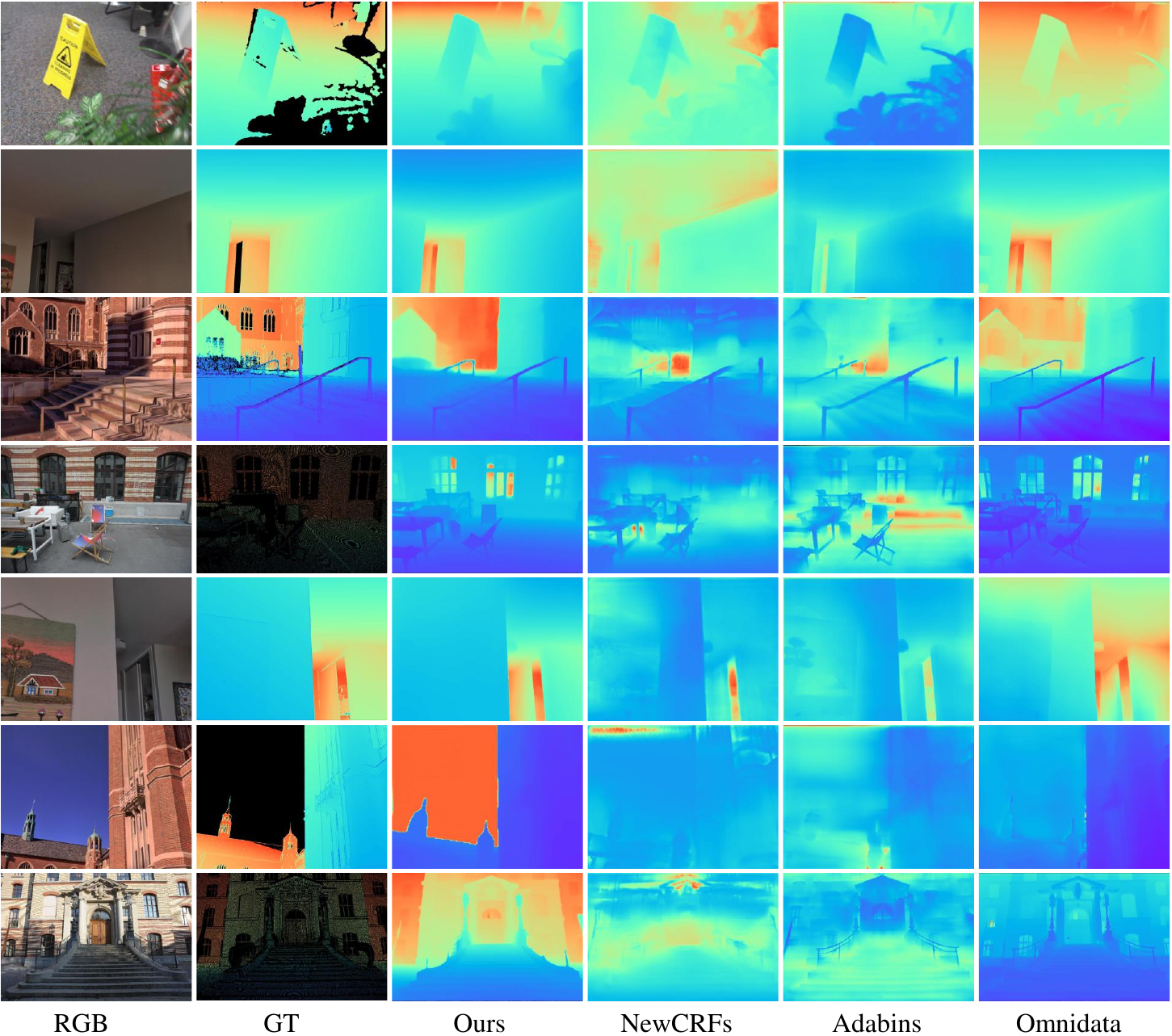}
\caption{\textbf{Depth estimation.} The visual comparison of predicted depth on iBims, ETH3D, and DIODE. Our depth maps come from the ConvNeXt-L CSTM\_label model.} 
\label{fig: depth_cmp2.}
\end{figure*}

\begin{figure*}[]
\centering
\includegraphics[width=0.9\textwidth]{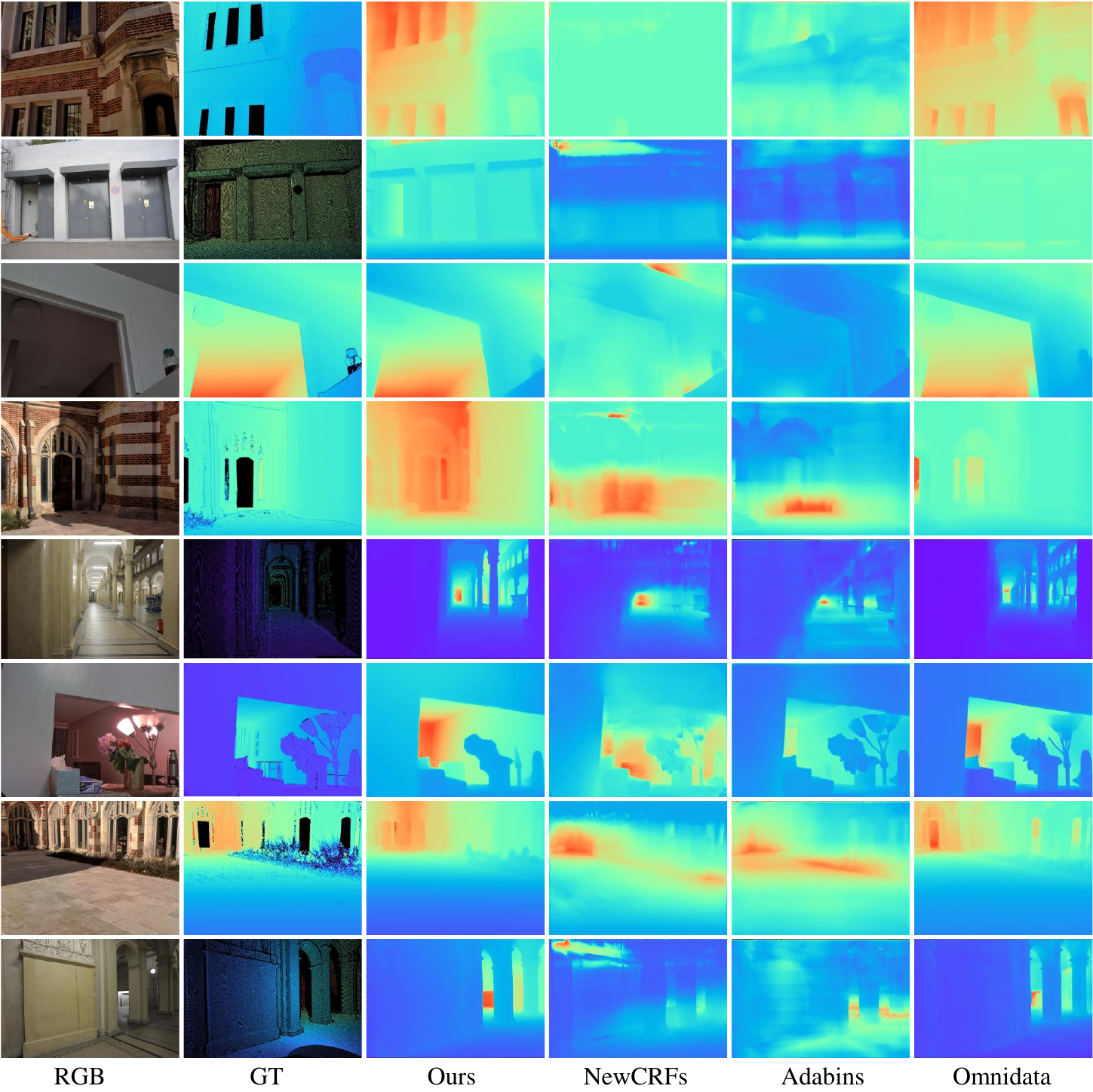}
\caption{\textbf{Depth estimation.} The visual comparison of predicted depth on iBims, ETH3D, and DIODE. Our depth maps come from the ConvNeXt-L CSTM\_label model.}
\label{fig: depth_cmp3.}
\end{figure*}

\begin{figure*}[]
\centering
\includegraphics[width=0.9\textwidth]{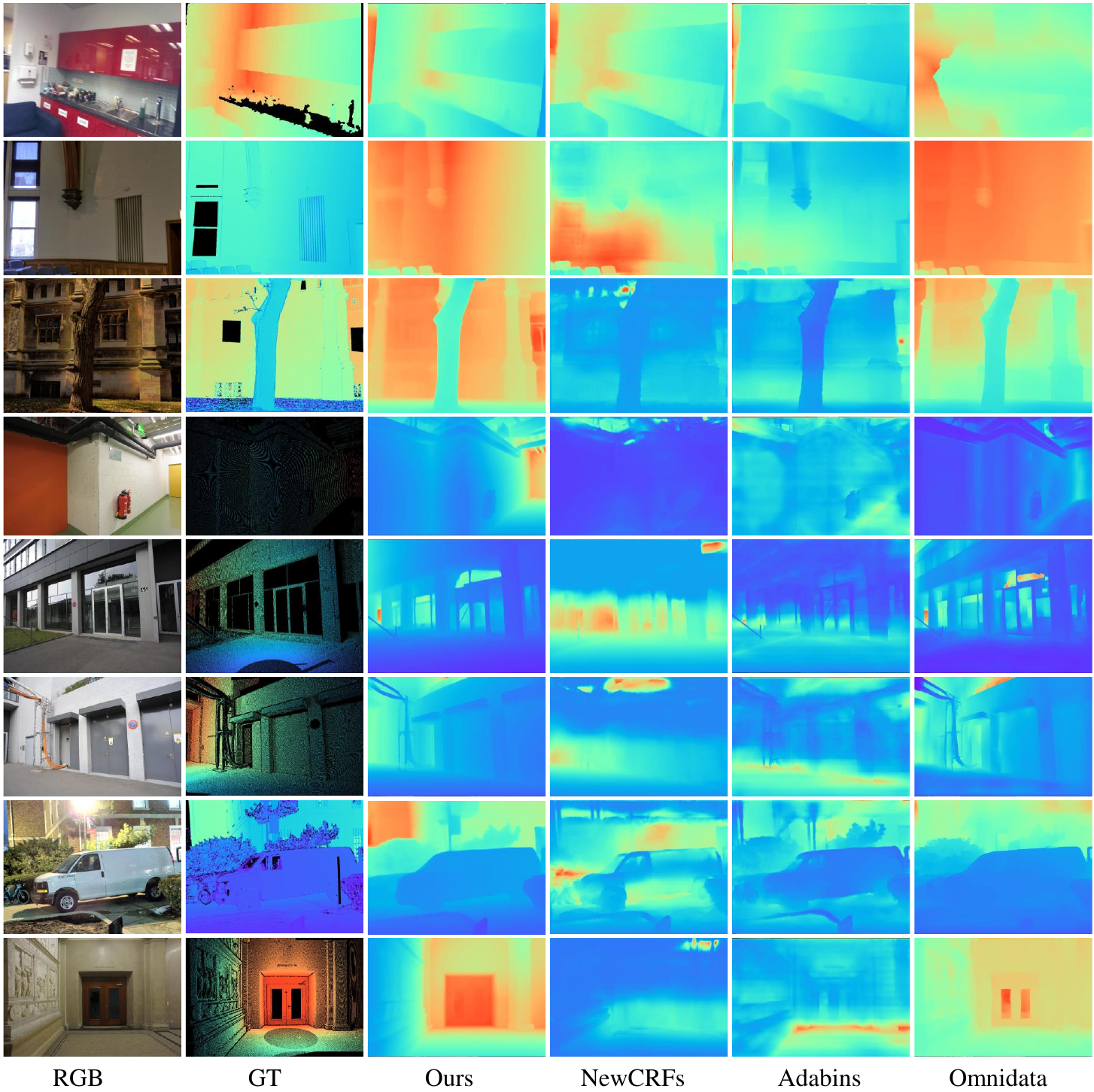}
\caption{\textbf{Depth estimation.} The visual comparison of predicted depth on iBims, ETH3D, and DIODE.  Our depth maps come from the ConvNeXt-L CSTM\_label model.}
\label{fig: depth_cmp4.}
\end{figure*}

\noindent\textbf{Qualitative comparison of depth and normal estimation.} In Figs ~\ref{fig: dn_cmp1.}, ~\ref{fig: dn_cmp2.}, we compare visualized depth and normal maps from the Vit-g CSTM\_label model with ZoeDepth~\cite{bhat2023zoedepth}, Bae \textit{etal} \cite{bae2021estimating}, and Omnidata~\cite{eftekhar2021omnidata}. In Figs.~\ref{fig: depth_cmp1.},~\ref{fig: depth_cmp2.},~\ref{fig: depth_cmp3.}, and ~\ref{fig: depth_cmp4.}, We show the qualitative comparison of our depth maps from the ConvNeXt-L CSTM\_label model with Adabins~\cite{bhat2021adabins}, NewCRFs~\cite{yuan2022new}, and Omnidata~\cite{eftekhar2021omnidata}. Our results have much fine-grained details and less artifacts.

\begin{figure*}[!htbp]
	\centering
	\includegraphics[width=0.8\textwidth]{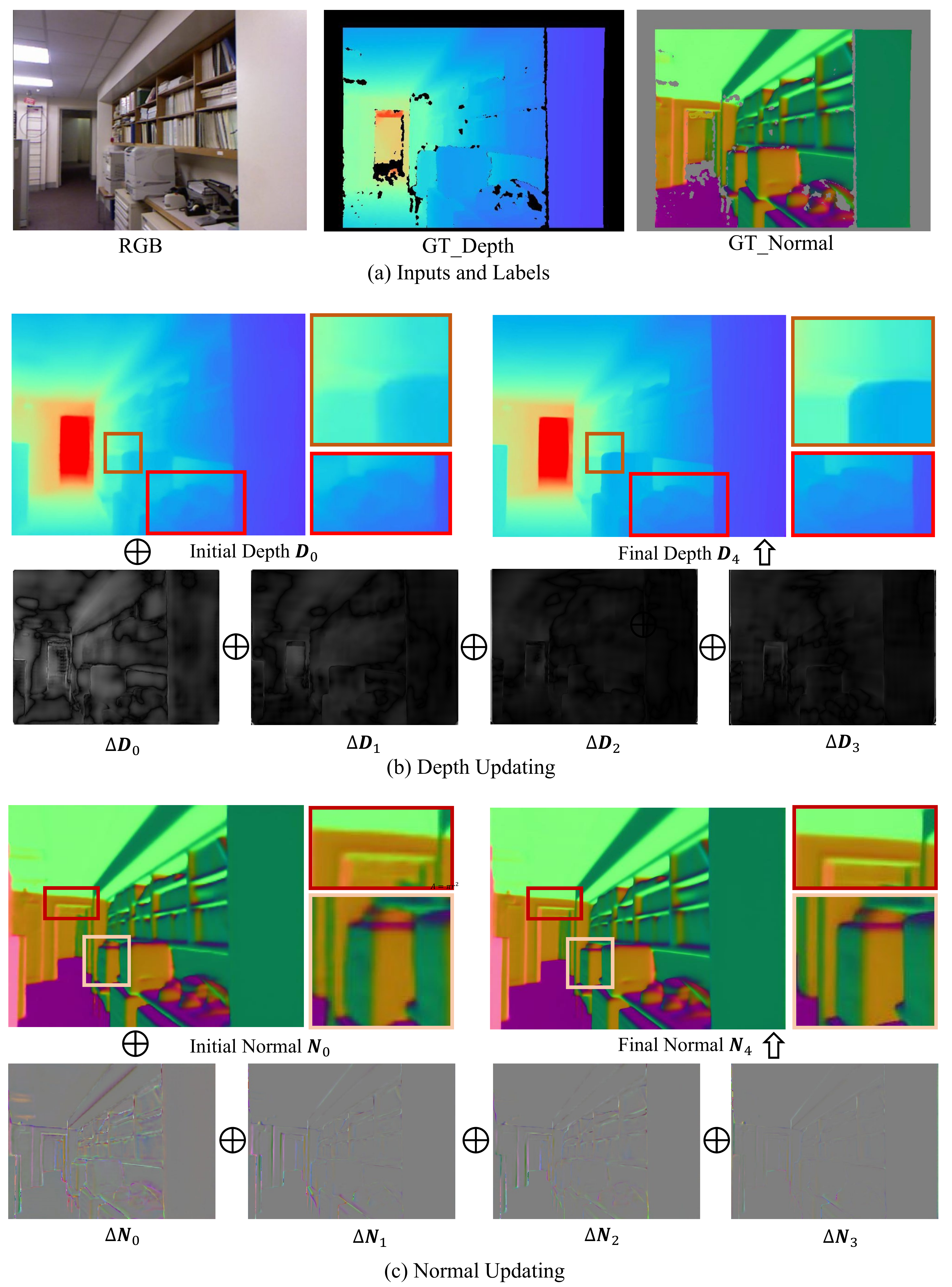}
	\caption{\red{\textbf{Visualization for iterative optimization.} We use our publicly available ViT-S model (with 4 refinement steps) to estimate zero-shot depth and normal maps. The initial and final predictions, as well as their sequential updating items,  are presented in (b) and (c). }}
 \label{fig: viz_opt}
	\vspace{-1.5em}
\end{figure*}

\noindent\red{\textbf{Visualization of iterative refinement.} To comprehensively understand the usage of iterative refinement modules, we visualize the predictions before/after optimization and the updates for different steps in Fig. \ref{fig: viz_opt}. Here we use our publicly released 4-step ViT-small model for visualization. Initially, the network produces a coarse prediction. The first step updates the most drastically, while sub-sequential steps focus mainly on object boundaries. Finally, the refined predictions have clearer shapes and sharper edges.}

\begin{figure*}[]
\centering
\includegraphics[width=1\textwidth]{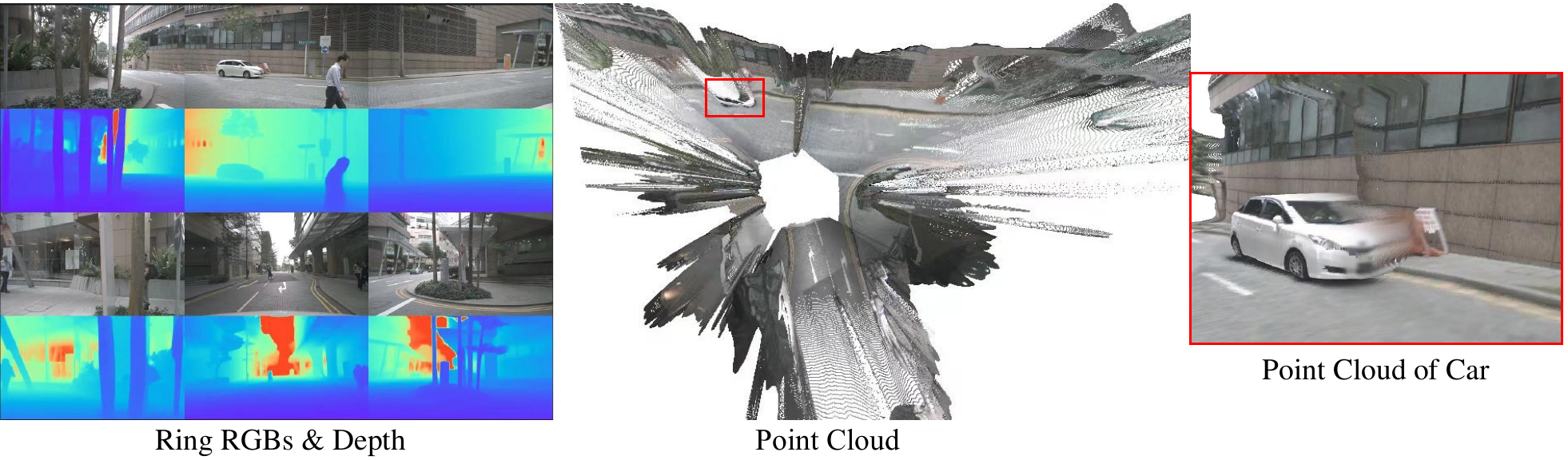}
\caption{\textbf{3D reconstruction of 360$^{\circ}$ views.} Current autonomous driving cars are equipped with several pin-hole cameras to capture 360$^{\circ}$ views. With our model, we can reconstruct each view and smoothly fuse them together. We can see that all views can be well merged together without scale inconsistency problems. Testing data are from NuScenes. Note that the front view camera has a different focal length from other views. }
\label{fig: recon nuscenes.}
\end{figure*}

\noindent\textbf{Reconstructing 360$^{\circ}$ NuScenes scenes.} Current autonomous driving cars are equipped with several pin-hole cameras to capture 360$^{\circ}$ views. Capturing the surround-view depth is important for autonomous driving.  We sampled some scenes from the testing data of NuScenes. With our depth model, we can obtain the metric depths for 6-ring cameras. With the provided camera intrinsic and extrinsic parameters, we unproject the depths to the 3D point cloud and merge all views together. See Fig. \ref{fig: recon nuscenes.} for details. Note that 6-ring cameras have different camera intrinsic parameters. We can observe that all views' point clouds can be fused together consistently.

\maketitle

\def\PWN{{\rm PWN}}
\def\VNL{{\rm VNL}}
\def\RPNL{{\rm RPNL}}

{\small
\bibliographystyle{ieeetr}
\bibliography{TPAMI}
}

%% file: introduction.tex
\begin{figure}[htbp]
	\centering
	\includegraphics[width=0.51\textwidth]{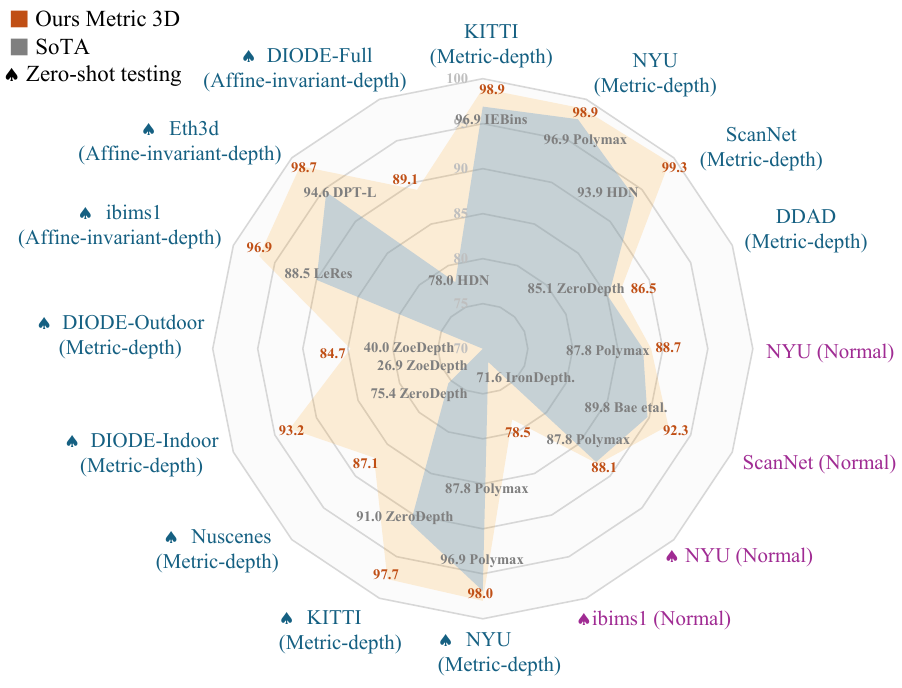}
	\vspace{-2em}
	\caption{\textbf{Comparisons with SoTA methods on 16 depth and normal benchmarks.}
 Radar-map of our Metric3D V2 v.s. SoTA methods from different works, on (1) Metric depth benchmarks, see `(Metric-depth)'. (2) Affine-invariant depth benchmarks, see `(Affine-invariant-depth)'. (3) Surface normal benchmarks, see `(Normal)'. Zero-shot testing is denoted by `\ding{171}'. Here $\delta_1$ percentage accuracy is used for depth benchmarks and $30^\circ$ percentage accuracy is for normal. Both higher values are for better performance. We establish new SoTA on a wide range of depth and normal benchmarks.}
	\label{Fig: radar-sota}
	\vspace{-1em}
\end{figure}

Monocular metric depth and surface normal estimation is the task of predicting absolute distance and surface direction from a single image. 
As crucial 3D representations, depth and normals are geometrically related and highly complementary. While metric depth excels in capturing data at scale, surface normals offer superior preservation of local geometry and are devoid of metric ambiguity compared to metric depth. These unique attributes render both depth and surface normals indispensable in various computer vision applications, including 3D reconstruction \cite{yang2018dense, ju2023dg, mescheder2019occupancy}, neural rendering (NeRF) \cite{deng2022depth, roessle2022dense, yu2022monosdf, jiang2023h2}, autonomous driving \cite{li2022bevdepth, li2023fb, fan2020sne}, and robotics \cite{behley2018efficient, murORB2, schops2019bad}. Currently, the community still lacks a robust, generalizable geometry foundation model \cite{zhu2023ponderv2, zhou2023uni3d, xu2023unifying} capable of producing high-quality metric depth and surface normal from a single image.

\begin{figure*}[]
	\centering
	\includegraphics[width=0.95\textwidth]{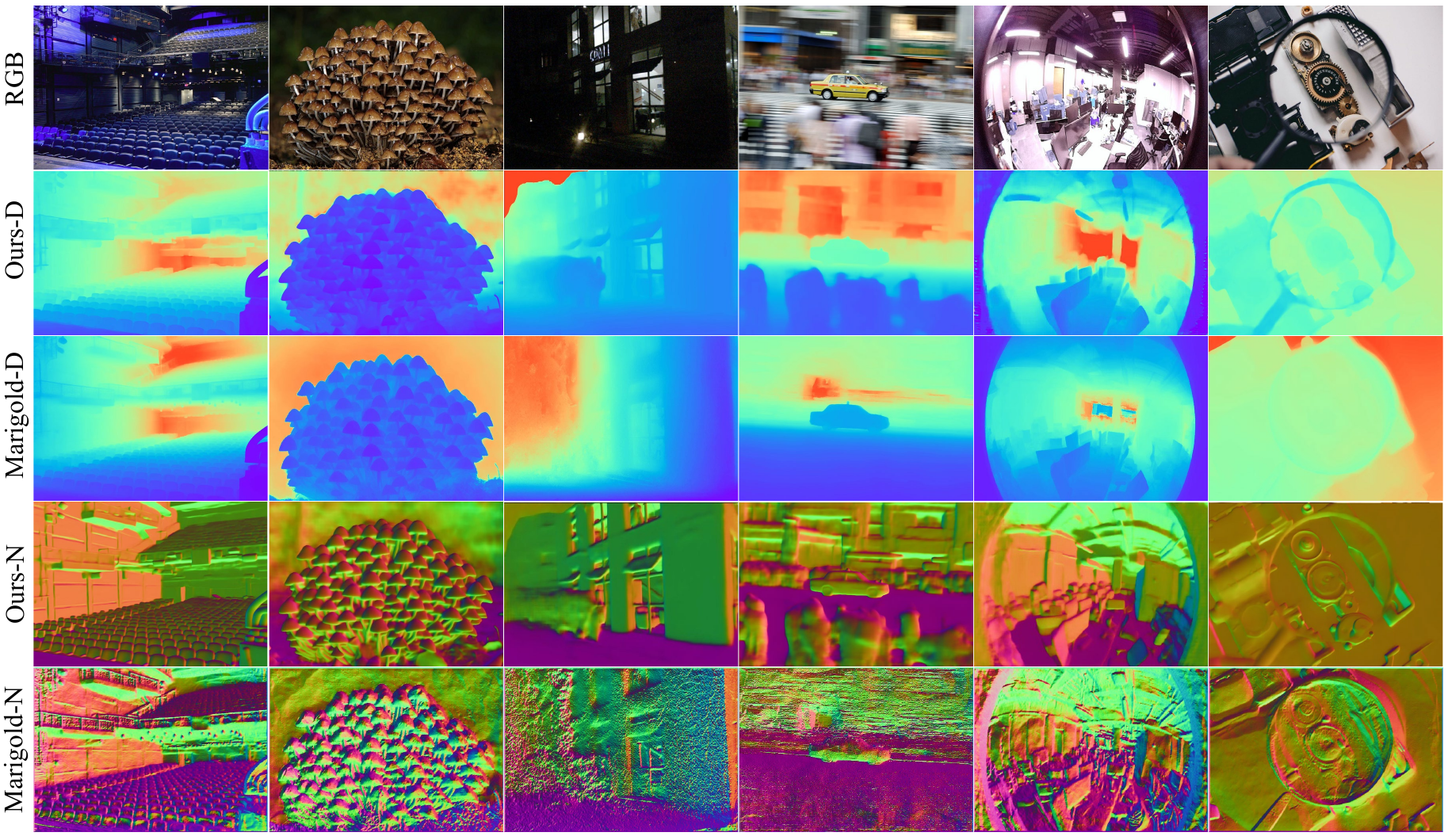}
	\vspace{-1.0 em}
	\caption{\textbf{Surface normal (N) and monocular depth (D) comparisons on diverse web images.}
Our method, directly estimating metric depths and surface normals, shows powerful generalization in a variety of scenarios, including indoor, outdoor, poor-visibility, motion blurred, and fisheye images. Visualized results come from our ViT-large-backbone estimator. Marigold is a strong and robust diffusion-based monocular depth estimation method, but its recovered surface normals from the depth show various artifacts.}
	\label{Fig: page 2.}
	\vspace{-1em}
\end{figure*}

Metric depth estimation and surface normal estimation confront distinct challenges. Existing depth estimation methods are categorized into learning metric depth \cite{yuan2022new, yin2021virtual, bhat2021adabins, yang2021transformers}, relative depth \cite{xian2018monocular, xian2020structure, chen2020oasis, chen2016single}, and affine-invariant depth~\cite{leres, yin2022towards, Ranftl2020, ranftl2021vision, zhang2022hierarchical}. Although the metric depth methods~\cite{yuan2022new, yin2021virtual, Yin2019enforcing, bhat2021adabins, yang2021transformers} have achieved impressive accuracy on various benchmarks, they must train and test on the dataset with the same camera intrinsics. Therefore, the training datasets of metric depth methods are often small, as it is hard to collect a large dataset covering diverse scenes using one identical camera. The consequence is that all these models generalize poorly in zero-shot testing, not to mention the camera parameters of test images can vary too.
A compromise is to learn the relative depth~\cite{chen2020oasis, xian2018monocular}, which only represents one point being further or closer to another one. The application of relative depth is very limited. Learning \emph{affine-invariant depth} finds a trade-off between the above two categories of methods, i.e. the depth is up to an unknown scale and shift.
With large-scale data, they decouple the metric information during training and achieve impressive robustness and generalization ability, such as MiDaS~\cite{Ranftl2020}, DPT~\cite{ranftl2021vision}, LeReS~\cite{leres, yin2022towards}, HDN~\cite{zhang2022hierarchical}. The problem is the unknown shift will cause 3D reconstruction distortions ~\cite{yin2022towards} and non-metric depth cannot satisfy various downstream applications.

In the meantime, these models cannot generate surface normals. Although lifting depths to 3D point clouds can do so, it places high demands on the accuracy and fine details of predicted depths. Otherwise, various artifacts will remain in such transformed normals. For example, Fig.~\ref{Fig: page 2.} shows noisy normals from Marigold~\cite{ke2023repurposing} depths, which excels in producing high-resolution fine depths. Instead of direct transformation, state-of-the-art (SoTA) surface normal estimation methods \cite{do2020surface, bae2021estimating, yang2024polymax} tend to train estimators on high-quality normal annotations. These annotations,  unlike sensor-captured ground-truth (GT), are derived from meticulously and densely reconstructed scenes, which have extremely rigorous requirements for both the capturing equipment and the scene. Consequently, data sources primarily consist of either synthetic creation or 3D indoor reconstruction~\cite{eftekhar2021omnidata}. Real and diverse outdoor scenes are exceedingly rare. (refer to our data statistics in Tab.~\ref{table: datasetsv2}). 
Limited by this label deficiency, SoTA surface normal methods \cite{do2020surface, bae2021estimating, yang2024polymax} typically struggle with strong zero-shot generalization. 
This work endeavors to tackle these challenges by developing a multi-task foundation model for \emph{zero-shot, single view, metric depth, and surface normal} estimation.

\begin{figure*}[]
	\centering
	\includegraphics[width=0.92\textwidth]{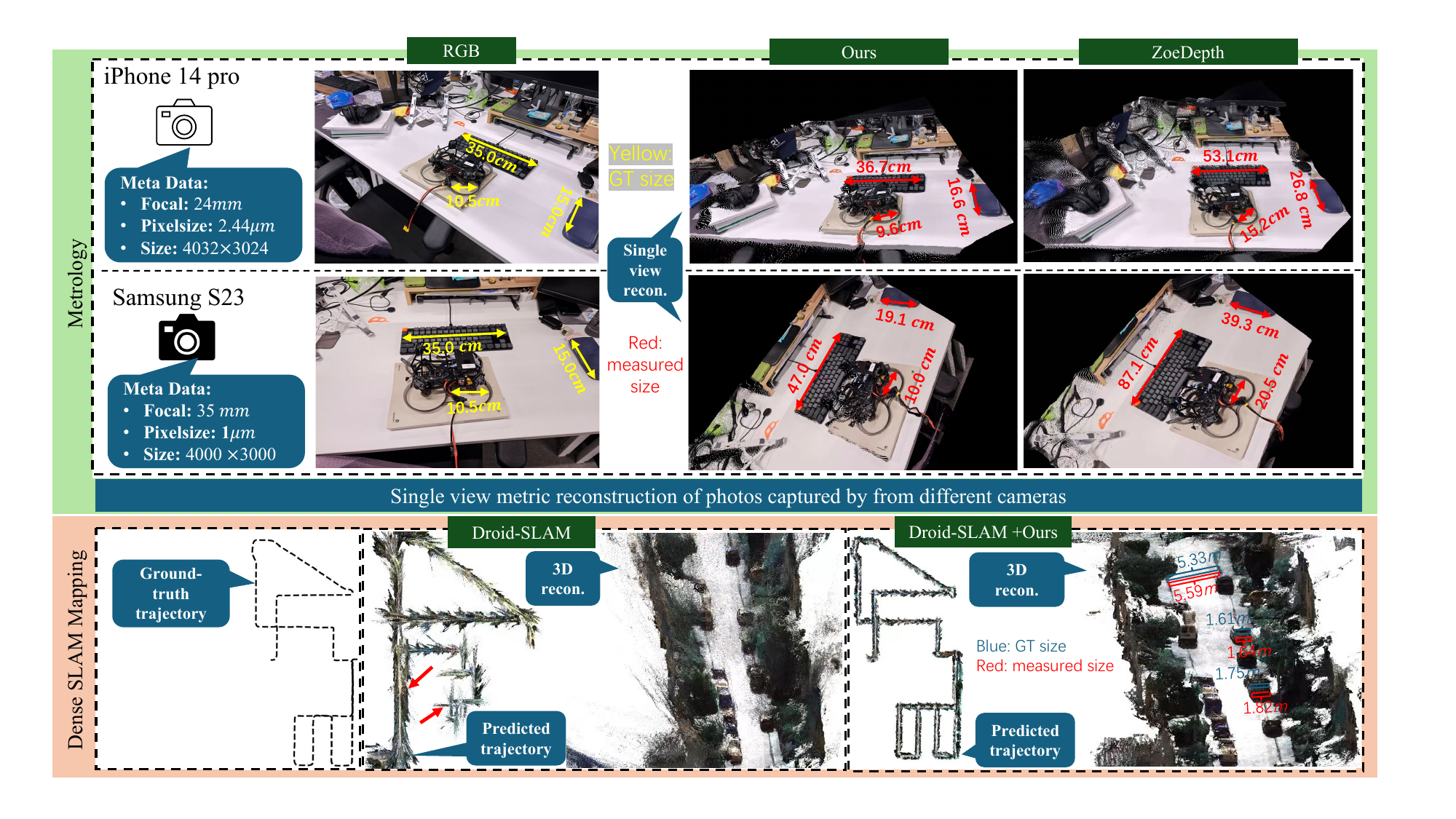}
	\vspace{-0.5 em}
	\caption{
		Top (metrology for a complex scene):  we use two phones (iPhone14 pro and Samsung Galaxy S23) to capture the scene and measure the size of several objects, including a drone which has never occurred in the whole training set. With the photos' metadata, we perform 3D metric reconstruction and then measure object sizes (marked in red), which are close to the ground truth (marked in yellow). %
        Compared with ZoeDepth~\cite{bhat2023zoedepth}, our measured sizes are closer to ground truth.
		Bottom  (dense SLAM mapping): existing SoTA mono-SLAM methods usually face scale drift problems (see the red arrows) in large-scale scenes and are unable to achieve the metric scale, while, naively inputting our metric depth, Droid-SLAM~\cite{teed2021droid} can recover much more accurate trajectory and perform the \textit{metric} dense mapping (see the red measurements). 
		Note that all testing data are unseen to our model.}
	\label{Fig: first page fig.}
	\vspace{-1.5em}
\end{figure*}

We propose targeted solutions for the challenges of zero-shot metric depth and surface normal estimation. For metric-scale recovery, we first analyze the metric ambiguity issues in monocular depth estimation and study different camera parameters in depth, including the pixel size, focal length, and sensor size. We observe that 
the focal length is the critical factor for accurate metric recovery. %
By design, affine-invariant depth methods do not take the focal length information into account during training. As shown in Sec.~\ref{sec:ambiguity}, only from the image appearance, various focal lengths may cause metric ambiguity, thus they decouple the depth scale in training.
To solve the problem of varying focal lengths, CamConv~\cite{facil2019cam}  encodes the camera model in the network, which enforces the network to implicitly understand camera models from the image appearance and then bridges the imaging size to the real-world size. However, training data contains limited images and types of cameras, which challenges data diversity and network capacity.
We propose a canonical camera transformation method in training, inspired by the canonical pose space from human body reconstruction methods ~\cite{peng2022animatable}. We transform all training data to a canonical camera space where the processed images are coarsely regarded as captured by the same camera. To achieve such transformation, we propose two different methods. The first one tries to adjust the image appearance to simulate the canonical camera, while the other one transforms the ground-truth labels for supervision. Camera models are not encoded in the network, making our method easily applicable to existing architectures. During inference, a de-canonical transformation is employed to recover metric information. 
To further boost the depth accuracy, we propose a random proposal normalization loss. It is inspired by the scale-shift invariant loss~\cite{leres, Ranftl2020,zhang2022hierarchical} decoupling the depth scale to emphasize the single image's distribution. However, they perform on the whole image, which inevitably squeezes the fine-grained depth difference. We propose to randomly crop several patches from images and enforce the scale-shift invariant loss~\cite{leres, Ranftl2020} on them. Our loss emphasizes the local geometry and distribution of the single image.

\begin{figure*}[t]
	\centering
	\includegraphics[width=0.93\textwidth]{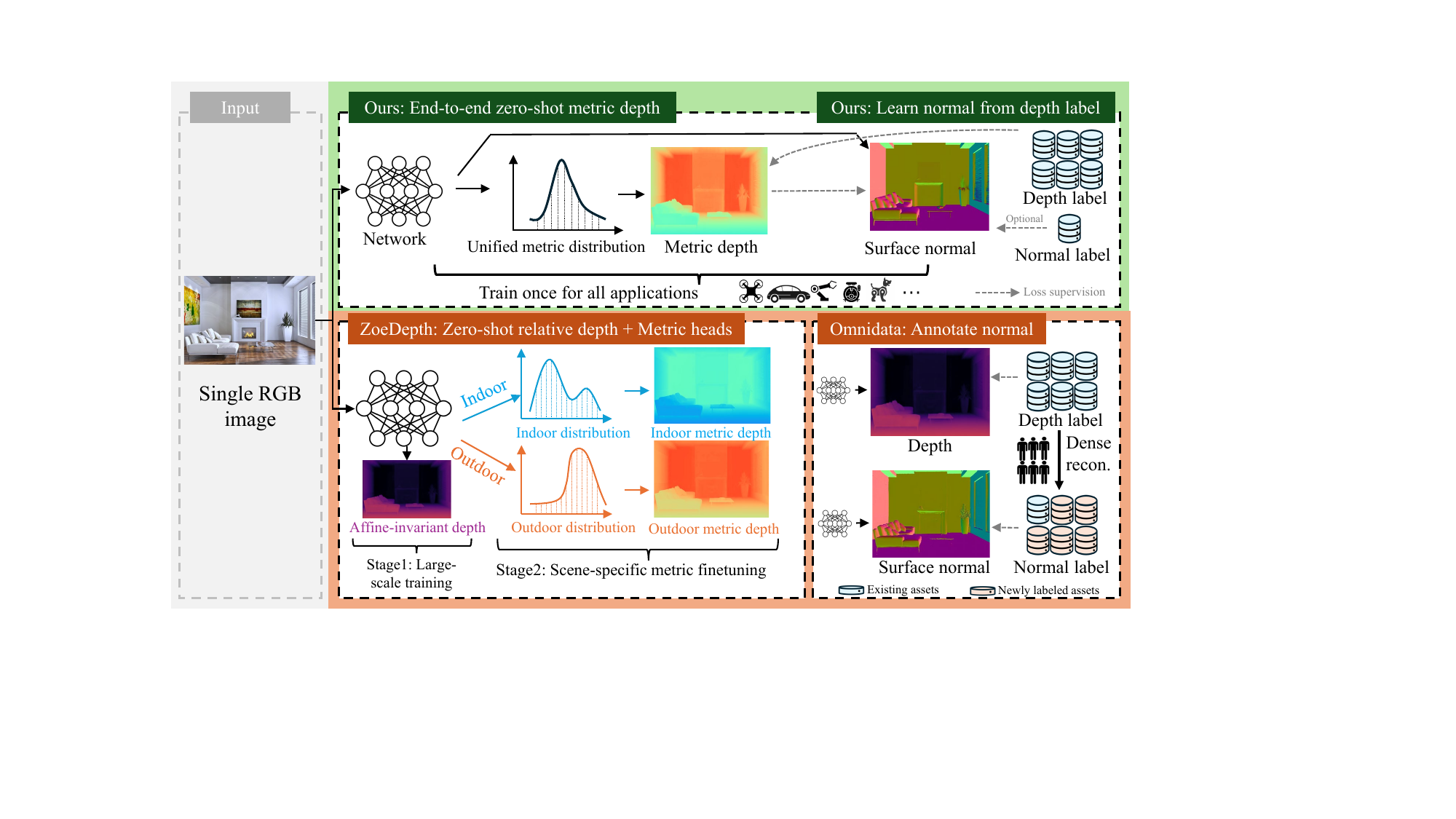}
	\vspace{-0.5 em}
	\caption{\textbf{Overall methodology.} Our method takes a single image to predict the metric depth and surface normal simultaneously. 
 We apply large-scale data training directly for metric depth estimation rather than affine invariant depth, enabling end-to-end zero-shot metric depth estimation for various applications using a single model. For normals, we enable learning from depth labels only, alleviating the demand for dense reconstruction to generate large-scale normal labels.}
	\label{fig: ours}
	\vspace{-1.5em}
\end{figure*}

For surface normal, the biggest challenge is the lack of diverse (outdoor) annotations. Compared to reconstruction-based annotation methods \cite{eftekhar2021omnidata, huang2019framenet}, directly producing normal labels from network-predicted depth is more efficient and scalable. The quality of such pseudo-normal labels, however, is bounded by the accuracy of the depth network. Fortunately, we observe that robust metric depth models are scalable geometric learners, containing abundant information for normal estimation. Weak supervision from the pseudo normal annotations transformed by learned metric depth can effectively prevent the normal estimator from collapsing caused by GT absence. Furthermore, this supervision can guide the normal estimator to generalize on large-scale unlabeled data. Based on such observation, we propose a joint depth-normal optimization module to distill knowledge from diverse depth datasets. During optimization, our normal estimator learns from three sources: (1) Groundtruth normal labels, though they are much fewer compared to depth annotations (2) An explicit learning objective to constrain depth-normal consistency. (3) Implicit and thorough knowledge transfer from depth to normal through feature fusion, which is more tolerant to unsatisfactory initial prediction than the explicit counterparts \cite{qi2018geonet}, \cite{bae2022irondepth}. To achieve this, we implement the optimization module using deep recurrent blocks. While previous researchers have employed similar recurrent modules to optimize depth \cite{bae2022irondepth, park2020non, shao2023iebins}, disparity \cite{lipson2021raft}, ego-motion \cite{teed2021droid}, or optical flows\cite{teed2020raft}, it is the first time that normal is iteratively optimized together with depth in a learning-based scheme. Benefiting from the joint optimization module, our models can efficiently learn normal knowledge from large-scale depth datasets even without labels.

With the proposed method, we can stably scale up model training to \emph{$16$ million} images from 16 datasets of diverse scene types (indoor and outdoor, real or synthetic data), camera models (tens of thousands of different cameras), and annotation categories (with or without normal), leading to zero-shot transferability and significantly improved accuracy. %
Fig.~\ref{fig: ours} illustrates how the large-scale data with depth annotations directly facilitate metric depth and surface normal learning.
The metric depth and normal given by our model directly broaden the applications in downstream tasks. We achieve state-of-the-art performance on over 16 depth and normal benchmarks, see Fig.~\ref{Fig: radar-sota}. Our model can accurately reconstruct metric 3D from randomly collected Internet images, enabling plausible single-image metrology. %
For examples (Fig.~\ref{Fig: first page fig.}), we recovery \emph{real-world metric} to improve monocular SLAM~\cite{teed2021droid, sun2022improving} and facilitate large-scale 3D reconstruction~\cite{im2019dpsnet}.
Our main contributions can be summarized as:
\begin{itemize}
\itemsep -0.0cm 
    \item \blue{We propose a canonical camera transformation method to address metric depth ambiguity across different camera settings. This approach facilitates training zero-shot monocular metric depth models using large-scale datasets.}
    \item \blue{We design a random proposal normalization loss to effectively improve metric depth.}
    \item We propose a joint depth-normal optimization module to learn normal on large-scale datasets without normal annotation, distilling knowledge from the metric depth estimator.
    \item Our models \textbf{rank 1st} on a wide variety of depth and surface normal benchmarks. It can perform high-quality 3D metric structure recovery in the wild and benefit several downstream tasks, such as mono-SLAM~\cite{teed2021droid, mur2017orb}, 3D scene reconstruction~\cite{im2019dpsnet}, and metrology~\cite{zhu2020single}. %
\end{itemize}
\vspace{-1em}

%% file: related_works.tex
\noindent\textbf{3D reconstruction from a single image.} 
\blue{The reconstruction of diverse objects from a singular image has been extensively investigated in prior research~\cite{barron2014shape, wang2018pixel2mesh, wu2018learning}. These methodologies exhibit proficiency in generating high-fidelity 3D models encompassing various entities such as cars, planes, tables, and human bodies~\cite{saito2019pifu, saito2020pifuhd}. The primary challenge lies in optimizing the recovery of object details, devising efficient representations within constrained memory resources, and achieving generalization across a broader spectrum of objects. However, these approaches typically hinge upon learning object-specific or instance-specific priors, often derived from 3D supervision, thereby rendering them unsuitable for comprehensive scene reconstruction. In addition to the aforementioned efforts on object reconstruction, several studies focus on scene reconstruction from single images. Saxena et al.\cite{saxena2008make3d} adopt an approach that segments the entire scene into multiple small planes, with the 3D structure represented based on the orientation and positioning of these planes. More recently, LeReS\cite{leres} proposed employing a robust monocular depth estimation model for scene reconstruction. Nonetheless, their method is limited to recovering shapes up to a certain scale. Zhang et al.~\cite{Zhang_2023_ICCV} recently introduced a zero-shot geometry-preserving depth estimation model capable of providing depth predictions up to an unknown scale.
In contrast to the aforementioned methodologies, our approach excels in recovering the metric 3D structure of scenes.}

\noindent\textbf{Supervised monocular depth estimation.}
\blue{Following the establishment of several benchmarks~\cite{silberman2012indoor, Geiger2013IJRR}, neural network-based methods~\cite{yuan2022new, Yin2019enforcing, bhat2021adabins} have dominated this task. These approaches often regress continuous depth by aggregating the information from an image~\cite{eigen2014depth}. However, since depth distribution varies significantly with different RGB values, some methods tend to discretize depth and reformulate the problem as a classification task~\cite{yin2021virtual} for better performance.
The generalization of deep models for 3D metric recovery faces two challenges: adapting to diverse scenes and predicting accurate metric information under various camera settings. Recent methods~\cite{xian2020structure, xian2018monocular, yin2021virtual} have effectively addressed the first challenge by creating large-scale relative depth datasets like DIW~\cite{chen2016single} and OASIS~\cite{chen2020oasis} to learn relative relations, which lose geometric structure information. To enhance geometry, methods like MiDaS~\cite{Ranftl2020}, LeReS~\cite{leres}, and HDN~\cite{zhang2022hierarchical} employ affine-invariant depth learning. These approaches, utilizing large-scale data, have continuously improved performance and scene generalization. However, they inherently struggle to recover metric information. Thus, achieving both strong generalization and accurate metric data across diverse scenes remains a key challenge to be addressed.} \red{Con-currently, ZoeDepth \cite{bhat2023zoedepth}, ZeroDepth \cite{guizilini2023towards}, and
UniDepth \cite{piccinelli2024unidepth} apply varying strategies to tackle this challenge.}

\noindent\textbf{Surface normal estimation.} Compared to metric depth, surface normal suffers no metric ambiguity and preserves local geometry better. These properties attract researchers to apply normal in various vision tasks like localization \cite{behley2018efficient}, mapping \cite{wang2019real}, and 3D scene reconstruction \cite{yu2022monosdf,wang2022neuris}. Currently, learning-based methods \cite{bae2021estimating, yang2024polymax, bae2022irondepth, qi2020geonet++, wang2015designing, eigen2015predicting, liao2019spherical, do2020surface, ladicky2014discriminatively, li2015depth, long2024adaptive, long2021adaptive, wang2022neuris} have dominated monocular surface normal estimation. Since normal labels required for training cannot be directly captured by sensors, previous works use \cite{silberman2012indoor, eigen2015predicting, qi2018geonet, qi2020geonet++} kernel functions to annotate normal from dense indoor depth maps \cite{silberman2012indoor}. These annotations become incomplete on reflective surfaces and inaccurate at object boundaries. To learn from such imperfect annotations, GeoNet \cite{qi2018geonet} proposes to enforce depth-normal consistency with mutual transformation modules, ASN~\cite{long2021adaptive,long2024adaptive} propose a novel adaptive surface normal constraint to facilitate joint depth-normal learning,
and Bae~\etal\cite{bae2021estimating} propose an uncertainty-based learning objective. Nonetheless, it is challenging for such methods to further increase their generalization, due to the limited %
dataset size and the diversity of %
scenes, especially for outdoor scenarios. Omni-data \cite{eftekhar2021omnidata} advances to fill this gap by building $1300$M frames of normal annotation. Normal-in-the-wild \cite{chen2017surface} proposes a pipeline for efficient normal labeling. \red{A con-current work DSINE \cite{bae2024dsine} also employs diverse datasets to train generalizable surface normal estimators.} However, further scaling up normal labels remains difficult. This underscores research significance in finding an efficient way to distill prior from other types of annotation. 

\noindent\textbf{Deep iterative refinement for geometry.} Iterative refinement enables multi-step coarse-to-fine prediction and benefits a wide range of geometry estimation tasks, such as optical flow estimation \cite{sun2018pwc, teed2020raft, liu2023learning}, depth completion \cite{cheng2018depth, park2020non, hu2021penet}, and stereo matching \cite{lipson2021raft, ma2022multiview, xu2023iterative}. Classical iterative refinements \cite{sun2018pwc, cheng2018depth} optimize directly on high-resolution outputs using high-computing-cost operators, limiting researchers from applying more iterations for better predictions. To address this limitation, RAFT\cite{teed2020raft} proposes to optimize an intermediate low-resolution prediction using ConvGRU modules. For monocular depth estimation, IEBins\cite{shao2023iebins} employs similar methods to optimize depth-bin distribution. Differently, IronDepth \cite{bae2022irondepth} propagates depth on pre-computed local surfaces. Regarding surface normal refinement, Lenssen \etal \cite{lenssen2020deep} propose a deep iterative method to optimize normal from point clouds. Zhao \etal \cite{zhao2021confidence} design a solver to refine depth and normal jointly, but it requires multi-view prior and per-sample post optimization. Without multi-view prior, such a non-learnable optimization method could fail due to unsatisfactory initial predictions. All the monocular methods \cite{shao2023iebins, bae2022irondepth, lenssen2020deep}, however, iterate over either depth or normal independently. In contrast, our joint optimization module tightly couples depth and normal with each other.

\noindent\textbf{Large-scale data training.}
Recently, various natural language problems and computer vision problems~\cite{
yin2022devil, radford2021learning, lambert2020mseg} have achieved impressive progress with large-scale data training. CLIP~\cite{radford2021learning} is a promising classification model trained on billions of paired image-language data pairs, achieving achieves state-of-the-art performance zero-shot classification benchmarks. Dinov2 \cite{oquab2023dinov2} collects $142$M images to conduct vision-only self-supervised learning for vision transformers\cite{dosovitskiy2020an}. Generative models like LDM \cite{rombach2022high} have also undergone billion-level data pre-training.
For depth prediction, large-scale data training has been widely applied. Ranft~\etal~\cite{Ranftl2020} mix over 2 million data in training, LeReS~\cite{yin2022towards} collects over $300$ thousands data, Eftekhar~\etal~\cite{eftekhar2021omnidata} also merge millions of data to build a strong depth prediction model. To train a zero-shot surface normal estimator, Omni-data \cite{eftekhar2021omnidata} performs dense reconstruction to generate $14$M frames with surface normal annotations.
\vspace{-1 em}

%% file: method.tex
\noindent\textbf{Preliminaries.}
We consider the pin-hole camera model with intrinsic parameters formulated as:
[[$\nicefrac{\hat{f}}{\delta}, 0, u_{0}$], [$0,  \nicefrac{\hat{f}}{\delta},  v_{0}$], [$0, 0,  1$]],
where $\hat{f}$ is the 
focal length (in micrometers), $\delta$ is the pixel size
(in micrometers), and
$(u_{0}, v_{0})$ is the principle center. $f = \nicefrac{\hat{f}}{\delta}$ is the pixel-represented focal length.
\vspace{-1em}
\subsection{Ambiguity Issues in Metric Depth Estimation}\label{sec:ambiguity}

\begin{figure}[!bt]
\centering
\includegraphics[width=0.45\textwidth]{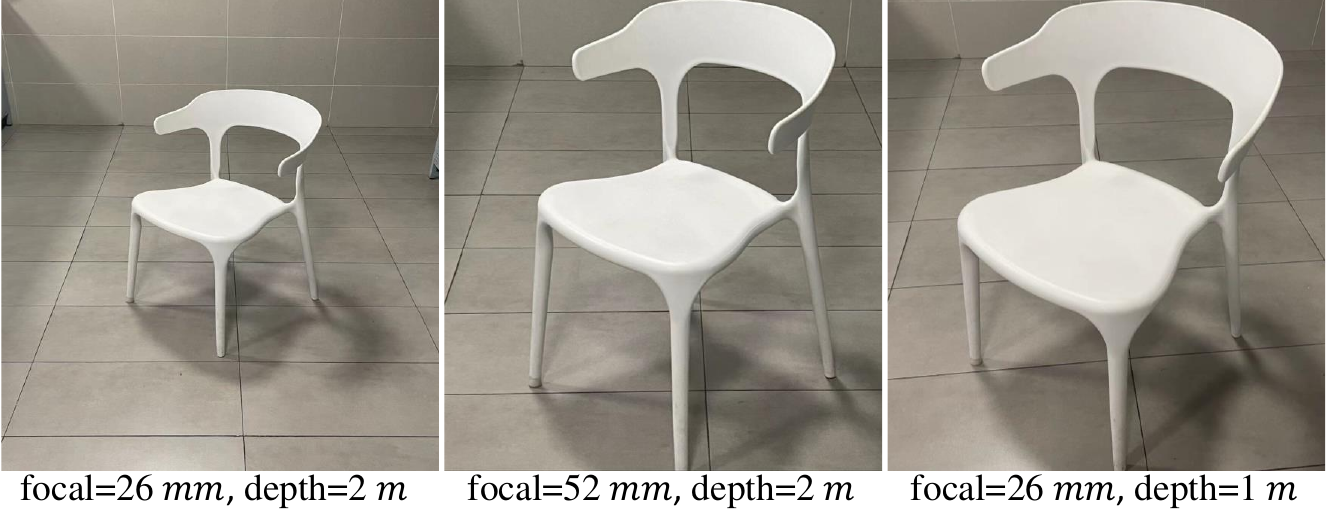}
\vspace{-0.5 em}
\caption{\textbf{Photos of a chair captured at different distances with different cameras}. The first two photos are captured at the same distance but with different cameras, 
while the last one is taken at a closer distance with the same camera as the first one.}
\label{fig: inspiration}
\vspace{-1.5em}
\end{figure}

\blue{Figure \ref{fig: inspiration} illustrates an instance of photographs captured using diverse cameras and at varying distances. Solely based on visual inspection, one might erroneously infer that the last two images originate from a comparable location and are captured by the same camera. However, due to differing focal lengths, these images are indeed captured from distinct locations. Consequently, accurate knowledge of camera intrinsic parameters becomes imperative for metric estimation from a single image; otherwise, the problem becomes ill-posed. Recent methodologies, such as MiDaS~\cite{Ranftl2020} and LeReS~\cite{leres}, mitigate this metric ambiguity by decoupling metric estimation from direct supervision and instead prioritize learning affine-invariant depth.}

\begin{figure}[!b]
\vspace{-2em}
\centering
\includegraphics[width=0.45\textwidth]{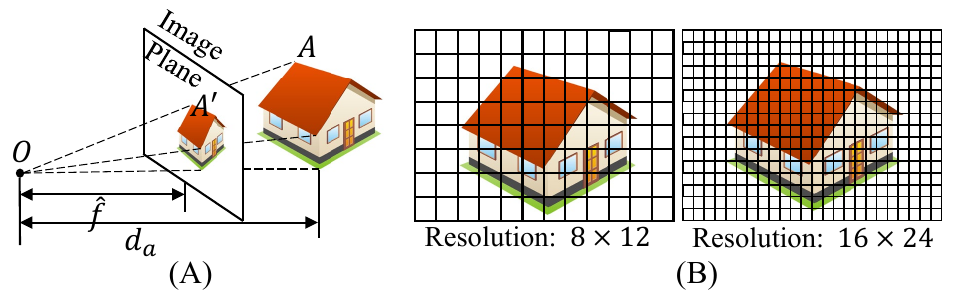}
\caption{\textbf{Pinhole camera model}. (A)Object $A$ positioned at a distance $d_{a}$ undergoes projection onto the image plane. (B) Employing two cameras for capturing an image of the car.  The left one has a larger pixel size. Although the projected imaging sizes are the same, the pixel-represented images (resolution) are different.}
\label{fig: pinhole camera}
\vspace{-0.5em}
\end{figure}

\blue{Figure \ref{fig: pinhole camera} (A) depicts the pin-hole perspective projection, where object $A$ located at distance $d_{a}$ is projected to $A'$. Adhering to the principle of similarity, it is obvious that:
\begin{equation}
\vspace{-1em}
    d_{a} = \hat{S} \Bigl[\frac{\hat{f}}{\hat{S'}}\Bigr]= \hat{S}\cdot \alpha
\label{eq: similarity}
\end{equation}
where $\hat{S}$ and $\hat{S'}$ are the real and \textit{imaging} size respectively. 
The symbol $\hat{\cdot}$ signifies that variables are expressed in physical metrics (e.g., millimeters). To ascertain $d_{a}$ from a single image, one must have access to the focal length, imaging size of the object, and real-world object size. Estimating the focal length from a single image is challenging and inherently ill-posed. Despite numerous methods having been explored~\cite{leres, hold2018perceptual}, achieving satisfactory accuracy remains elusive. Hereby, we simplify the scenario by assuming known focal lengths for the training/test images.}  In contrast, understanding the imaging size is much easier for a neural network. To obtain the real-world object size, a neural network needs to understand the semantic scene layout and the object, at which a neural network excels. We define $\alpha = \nicefrac{\hat{f}}{\hat{S'}} $, showing that $d_{a}$ is proportional to $\alpha$.

\blue{We observe the following regarding sensor size, pixel size, and focal length.}

\noindent
\textbf{O1: Sensor size and pixel size do not affect metric depth estimation.} \blue{Based on perspective projection (Fig.~\ref{fig: pinhole camera} (A)), sensor size only influences the field of view (FOV) and is not relevant to $\alpha$, hence it does not affect metric depth estimation. For pixel size, consider two cameras with different pixel sizes ($\delta_{1} = 2\delta_{2}$) but the same focal length $\hat{f}$, capturing the same object at distance $d_{a}$. Fig.~\ref{fig: pinhole camera} (B) displays their captured images. According to the preliminaries, the pixel-represented focal length is $f_{1} = \frac{1}{2} f_{2}$. Since the second camera has smaller pixel sizes, the resolution of the pixel-represented image is given by $S'_{1} = \frac{1}{2} S'_{2}$, despite both having the same projected imaging size $\hat{S'}$. According to Eq.~\eqref{eq: similarity}, we have $\frac{\hat{f}}{\delta_{1}\cdot S'_{1}} = \frac{\hat{f}}{\delta_{2}\cdot S'_{2} }$, which implies $\alpha_1 = \alpha_2$ and $d_{1} = d_{2}$. This means that variations in camera sensors do not impact the estimation of metric depth.
}

\begin{figure}[!bt]
\centering
\includegraphics[width=0.45\textwidth]{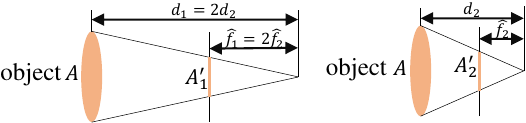}
\caption{\textbf{Illustration of two cameras with different focal length} at different distance. As $f_1=2f_2$ and $d_1=2d_2$, 
$A$ is projected 
to two image planes with the same imaging size (i.e. $A^{'}_1 = A^{'}_2$).
}
\label{fig: confusion}
\vspace{-2em}
\end{figure}

\noindent\textbf{O2: The focal length is vital for metric depth estimation}. 
\blue{Figure \ref{fig: inspiration} shows the challenge of metric ambiguity caused by an unspecified focal length, which is further discussed in Figure \ref{fig: confusion}. In the scenario where two cameras ($\hat{f}_{1} = 2\hat{f}_{2}$) are positioned at distances $d_{1} = 2d_{2}$, the imaging sizes remain consistent for both cameras. As a result, the neural network struggles to distinguish between different supervision labels based solely on visual cues. To address this issue, we propose a canonical camera transformation method to reduce conflicts between supervision requirements and image representations.}

\begin{figure}[h]
	\vspace{-1em}
	\centering
	\includegraphics[width=0.45\textwidth]{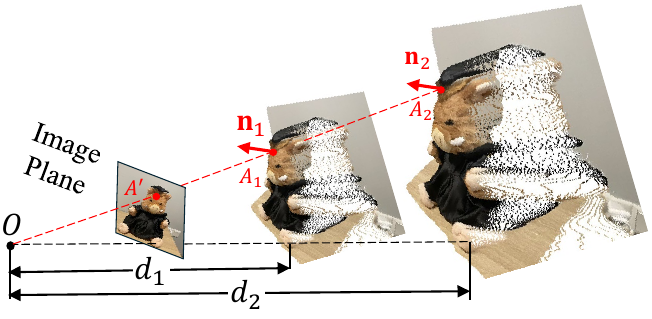}
	\caption{\textbf{The metric-agnostic property of normal.} With differently predicted metrics $d_1$ and $d_2$, the pixel $A^{'}$ on the image will be back-projected to 3D points $A_1$ and $A_2$, respectively. The surface normal $\textbf{n}_1$ at $A_1$ and $\textbf{n}_2$ at $A_2$ remain the same. 
	}
	\label{fig: normal}
	\vspace{-2.5em}
\end{figure}

Unlike depth, surface normal does not have any metric ambiguity problem. In Fig. \ref{fig: normal}, we illustrate this concept with two depth maps at varying scales, denoted as $\mathbf{D}_1$ and $\mathbf{D}_2$, featuring distinct metrics d1 and d2, respectively, where $d_1<d_2$. After upprojecting the depth to the 3D point cloud, the dolls are in different depths $d_1$ and $d_2$. However, the surface normals $\mathbf{n}_1$ and $\mathbf{n}_2$ corresponding to a certain pixel $A' \in \mathbf{I}$ remain the same.

\begin{figure*}[]
	\centering
	\includegraphics[width=0.94\textwidth]{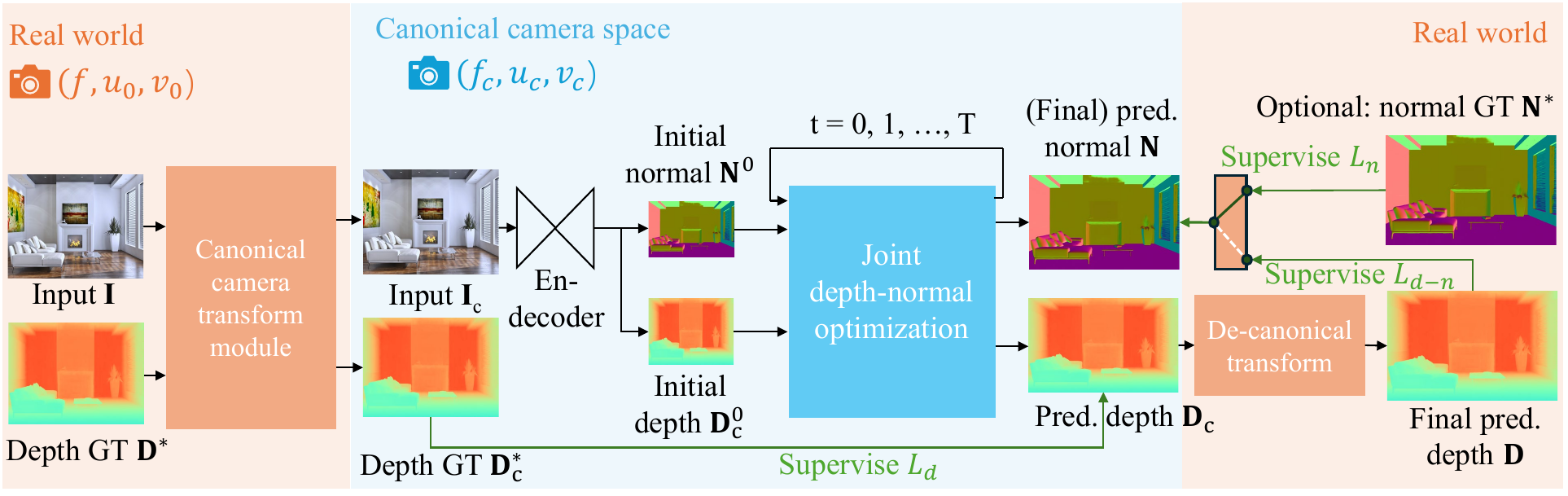}
	\vspace{-0.8 em}
	\caption{\textbf{Pipeline.}
		Given an input image $I$, we first transform it to the canonical space using CSTM. The transformed image $I_c$ is fed into a standard depth-normal estimation model to produce the predicted metric depth $D_c$ in the canonical space and metric-agnostic surface normal $N$. During training, $D_c$ is supervised by a GT depth $D^*_c$ which is also transformed into the canonical space. In inference, after producing the metric depth $D_c$ in the canonical space, we perform a de-canonical transformation to convert it back to the space of the original input $I$. The canonical space transformation and de-canonical transformation are executed using camera intrinsics. The predicted normal $N$ is supervised by depth-normal consistency via the recovered metric depth $D$ as well as GT normal $N^\ast$, if available.}
	\label{fig: pipeline}
	\vspace{-1.8em}
\end{figure*}

\subsection{Canonical Camera Transformation}

\blue{
The fundamental concept entails establishing a canonical camera space ($(f_{x}^{c}, f_{y}^{c})$, with $f_{x}^{c}=f_{y}^{c}=f^{c}$ in experimental settings) and transposing all training data into this designated space. Consequently, all data can be broadly construed as being captured by the canonical camera. 
We propose two transformation methods, i.e. either transforming the input image ($\mathbf{I}\in\mathbb{R}^{H \times W \times 3}$) or the ground-truth (GT) label ($\mathbf{D}\in\mathbb{R}^{H \times W}$). The initial intrinsics are $\{f, u_{0}, v_{0}\}$.}

\noindent\blue{\textbf{Method1: transforming depth labels (CSTM\_label).}
Fig.~\ref{fig: inspiration}'s ambiguity is for depths. 
Consequently, our initial approach directly addresses this issue by transforming the ground-truth depth labels. Specifically, we rescale the ground-truth depth ($\mathbf{D}^{*}$) using the ratio $\omega_d = \frac{f^{c}}{f}$ during training, denoted as $\mathbf{D}^{}_{c} = \omega_d \mathbf{D}^{*}$. The original camera model undergoes transformation to ${f^{c}, u_{0}, v_{0}}$. In inference, the predicted depth ($\mathbf{D}_{c}$) exists in the canonical space and necessitates a de-canonical transformation to restore metric information, expressed as $\mathbf{D} = \frac{1}{\omega_d}\mathbf{D}_{c}$. It is noteworthy that the input $\mathbf{I}$ remains unaltered, represented as $\mathbf{I}_c = \mathbf{I}$.}

\noindent\blue{\textbf{Method2: transforming input images (CSTM\_image).}
From an alternate perspective, the ambiguity arises due to the resemblance in image appearance. Consequently, this methodology aims to alter the input image to emulate the imaging effects of the canonical camera. Specifically, the image $\mathbf{I}$ undergoes resizing using the ratio $\omega_r=\frac{f^{c}}{f}$, denoted as $\mathbf{I}{c} = \mathcal{T}(\mathbf{I}, \omega_r)$, where $\mathcal{T}(\cdot)$ signifies image resizing. As a result of resizing the optical center, the canonical camera model becomes ${f^{c}, \omega_r u_{0}, \omega_r v_{0}}$. The ground-truth labels are resized without scaling, represented as $\mathbf{D}^{*}_{c} = \mathcal{T}(\mathbf{D}^*, \omega_r)$. In inference, the de-canonical transformation involves resizing the prediction to its original dimensions without scaling, expressed as $\mathbf{D} = \mathcal{T}(\mathbf{D}_{c}, \frac{1}{\omega_r})$.}

\red{While similar transformations have been employed in MPSD~\cite{MapillaryPSD} to normalized depth prediction, our approaches apply these modules to predict metric depth directly.}

\blue{Figure \ref{fig: pipeline} shows the pipeline. After adopting either transformation, a patch is randomly cropped for training purposes. This cropping operation solely adjusts the field of view (FOV) and the optical center, thus averting any potential metric ambiguity issues. In the labels transformation approach, $\omega_r = 1$ and $\omega_d=\frac{f^c}{f}$, while in the images transformation method, $\omega_d = 1$ and $\omega_r=\frac{f^c}{f}$. Throughout the training process, the transformed ground-truth depth labels $\mathbf{D}^{*}_{c}$ are employed as supervision. Importantly, since surface normals are not susceptible to metric ambiguity, no transformation is applied to normal labels ${\mathbf{N}}^{*}$.}
\vspace{-1em}

\subsection{Jointly optimizing depth and normal}

\noindent We propose to optimize metric depth and surface normal jointly in an end-to-end manner. This optimization is primarily aimed at leveraging a large amount of annotation knowledge available in depth datasets to improve normal estimation, particularly in outdoor scenarios where depth datasets contain significantly more annotations than normal datasets. In our experiments, we collect from the community $9488$K images with depth annotations across $14$ outdoor datasets while less than $20$K outdoor normal-labeled images, presented in Tab.~\ref{table: datasetsv2}. 

\begin{figure*}[!bt]
	\centering
  	\vspace{-1.3em}
	\includegraphics[width=0.9\textwidth]{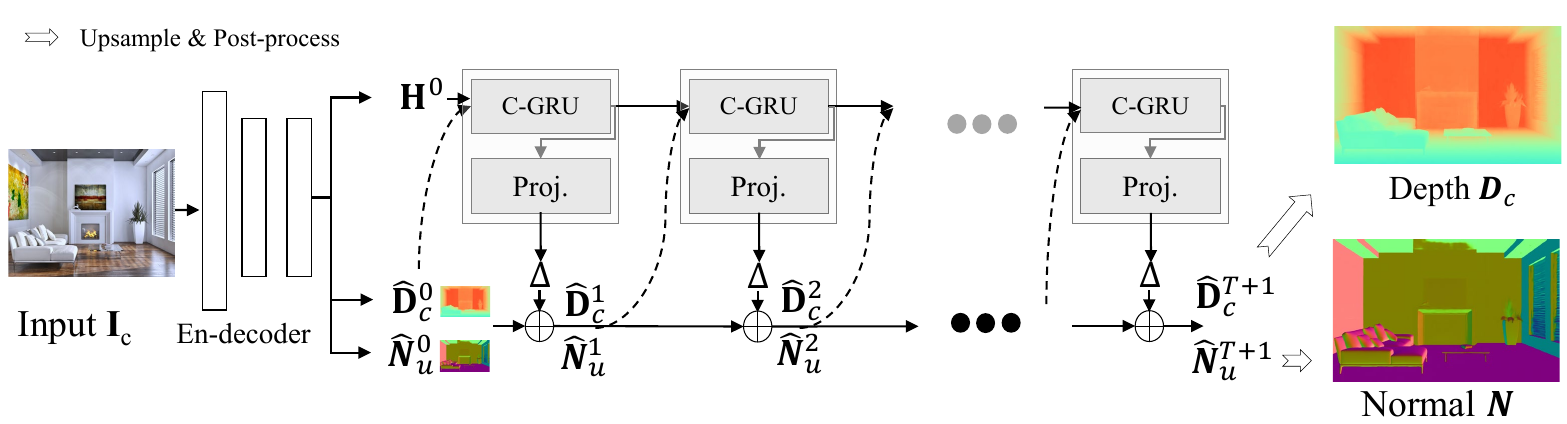}
  	\vspace{-0.5em}
	\caption{\textbf{Joint depth and normal optimization.} In the canonical space, we deploy recurrent blocks composed of ConvGRU sub-blocks (C-RGU) and projection heads (Proj.) to predict the updates $\Delta$. During optimization, intermediate low-resolution depth and normal  $\hat{\mathbf{D}}_c^0$ $\hat{\mathbf{N}}_u^0$ are initially given by the decoder, and then iteratively refined by the predicted updates $\Delta$. After $T+1$ iterations, the optimized intermediate predictions $\hat{\mathbf{D}}_c^{T+1}$ $\hat{\mathbf{N}}_u^{T+1}$ are upsampled and post-processed to obtain the final depth $\mathbf{D}_c$ in the canonical space and the final normal $\mathbf{N}$. 
	}
	\label{fig: optimization}
	\vspace{-1.3em}
\end{figure*}

To facilitate knowledge flow across the depth and normal, we implement the learning-based optimization with recurrent refinement blocks, as depicted in Fig \ref{fig: optimization}. Unlike previous monocular methods \cite{bae2022irondepth, shao2023iebins}, our method updates both depth and normal iteratively through these blocks. Inspired by RAFT \cite{teed2020raft, lipson2021raft}, we iteratively optimize the intermediate low-resolution depth $\hat{\mathbf{D}_c}$ and unnormalized normal $\hat{\mathbf{N}_u}$, where \^{} denotes low resolution prediction $\hat{\mathbf{D}}_c \in \mathbb{R}^{\frac{H}{4} \times \frac{W}{4}}$, and $\hat{\mathbf{N}}_u \in \mathbb{R}^{\frac{H}{4} \times \frac{W}{4} \times 3}$, and the subscript $_c$ means the depth $\hat{\mathbf{D}}_c$ is in canonical space. As sketched in Fig. \ref{fig: optimization}, $\hat{\mathbf{D}}_c^t$ and $\hat{\mathbf{N}}_u^t$ represent the low-resolution depth and normal optimized after step $t$, where $t=0, 1, 2, \dots, T$ denotes the step index. Initially, at step $t=0$, $\hat{\mathbf{D}}_c^0$ and $\hat{\mathbf{N}}_u^0$ are given by the decoder. In addition to updating depth and normal, the optimization module also updates hidden feature maps $\mathbf{H}^t$, which are initialized by the decoder. During each iteration, the learned recurrent block $\mathcal{F}$ output updates $\Delta\hat{\mathbf{D}}_c$, $\Delta\hat{\mathbf{N}}_u$ and renews the hidden features $\mathbf{H}$:
\begin{eqnarray}
 \Delta\hat{\mathbf{D}}_c^{t+1}, \Delta\hat{\mathbf{N}}_u^{t+1}, \mathbf{H}^{t+1} =  \mathcal{F}(\hat{\mathbf{D}}_u^{t}, \hat{\mathbf{N}}_u^{t}, \mathbf{H}^{t}, \mathbf{H}^{0}),
\label{eq: recurrent module}
\end{eqnarray}
The updates are then applied for updating the predictions:
\begin{eqnarray}
\begin{split}
 \hat{\mathbf{D}}^{t+1}_c = \hat{\mathbf{D}}^{t}_c + \Delta\hat{\mathbf{D}}^{t+1}_c , \  \hat{\mathbf{N}}^{t+1}_u = \hat{\mathbf{N}}^{t}_u + \Delta\hat{\mathbf{N}}^{t+1}_u,
\end{split}
\label{eq: update}
\end{eqnarray}
To be more specific, the recurrent block $\mathcal{F}$ comprises a ConvGRU sub-block and two projection heads. First, the ConvGRU sub-block updates the hidden features ${\mathbf{H}}^{t}$ taking all the variables as inputs. Subsequently, the two branched projection heads $\mathcal{G}_d$ and $\mathcal{G}_n$ estimate the updates $\Delta\hat{\mathbf{D}}^{t+1}$ and $\Delta\hat{\mathbf{N}}^{t+1}$ respectively. A more comprehensive representation of Eq. \ref{eq: recurrent module}, therefore, can be written as:
\begin{eqnarray}
\begin{split}
& \quad \mathbf{H}^{t+1} = \mathrm{ConvGRU}(\hat{\mathbf{D}}^{t}, \hat{\mathbf{N}}^{t}, \mathbf{H}^{0}, \mathbf{H}^{t}), \\ & \Delta\hat{\mathbf{D}}^{t+1} = \mathcal{G}_d(\mathbf{H}^{t+1}), \quad \Delta\hat{\mathbf{N}}^{t+1} = \mathcal{G}_n(\mathbf{H}^{t+1}).
\end{split}
\label{eq: detailed recurrent module}
\end{eqnarray}
For detailed structures of the refinement module $\mathcal{F}$, we recommend readers refer to supplementary materials.

After $T+1$ iterative steps, we obtain the well-optimized low-resolution predictions $\mathbf{\hat{D}}^{T+1}_c$ and $\mathbf{\hat{N}}^{T+1}_u$. These predictions are then up-sampled and post-processed to generate the final depth $\mathbf{D}_c$ and surface normal $\mathbf{N}$: 
\begin{eqnarray}
\begin{split}
\mathbf{D}_c = \mathcal{H}_d(\text{upsample}(\hat{\mathbf{D}}^{T+1}_c)) \\
\mathbf{N} = \mathcal{H}_n(\text{upsample}(\hat{\mathbf{N}}^{T+1}_u)),
\end{split}
\label{eq: final_output}
\end{eqnarray}
where $\mathcal{H}_d$ is the ReLU function to guarantee depth is non-negative, and $\mathcal{H}_n$ represents normalization to ensure $\Vert\mathbf{n}\Vert = 1$ for all pixels.

In a general formulation, the end-to-end network in Fig.~\ref{fig: optimization} can be rewritten as:
\begin{equation}
    \mathbf{D}_c, \ \mathbf{N} = \mathcal{N}_{d-n}(\mathbf{I}_{c}, \theta)  
\label{eq: full_network}
\vspace{-0.5 em}
\end{equation} where
$\theta$ is the network's ($\mathcal{N}_{d-n}$) parameters.

\vspace{1em}
\subsection{Supervision}
\noindent The training objective is:
\begin{equation}
    \min_{\theta} L (\mathcal{N}_{d-n}(\mathbf{I}_{c}, \theta), \mathbf{D}^ {*}_{c}, \mathbf{N}^ {*})  
\label{eq: robust metric depth}
\vspace{-0.5 em}
\end{equation}
where $\mathbf{D}^{*}_{c}$ and $\mathbf{I}_{c}$ are transformed ground-truth depth labels and images in the canonical space $c$, $\mathbf{N}^{*}$ denotes normal labels, $L$ is the supervision loss to be illustrated as following.

\noindent\textbf{Random proposal normalization loss.} To boost the performance of depth estimation, we propose a random proposal normalization loss (RPNL). The scale-shift invariant loss~\cite{Ranftl2020, leres} is widely applied for the affine-invariant depth estimation, which decouples the depth scale to emphasize the single image distribution. However, such normalization based on the whole image inevitably squeezes the fine-grained depth difference, particularly in close regions. Inspired by this, we propose to randomly crop several patches ($p_{i(i=0,...,M)} \in \mathbb{R}^{h_i \times w_i}$) from the ground truth $\mathbf{D}^{*}_c$ and the predicted depth $\mathbf{D}_c$. Then we employ the median absolute deviation normalization~\cite{singh2019investigating} for paired patches. By normalizing the local statistics, we can enhance local contrast. The loss function is as follows:
\begin{eqnarray}\nonumber
L_{\RPNL} = \frac{1}{MN} \sum_{p_i}^{M}\sum_{j}^{N} \lvert \frac{d^{*}_{p_i, j} - \mu(d^{*}_{p_i, j})}{\frac{1}{N}\sum_{j}^{N} \left |d^{*}_{p_i, j} - \mu(d^{*}_{p_i, j}) \right |} - \\
\frac{d_{p_i, j} - \mu(d_{p_i, j})}{\frac{1}{N}\sum_{j}^{N} \left | d_{p_i, j} - \mu(d_{p_i, j}) \right |} \rvert
\label{eq: RPNL}
\end{eqnarray}
where $d^*\in \mathbf{D}^*_c$ and $d \in \mathbf{D}_c$ are the ground truth and predicted depth respectively. $\mu(\cdot)$ and is the median of depth. $M$ is the number of proposal crops, which is set to 32. During training, proposals are randomly cropped from the image by $0.125$ to $0.5$ of the original size. Furthermore, several other losses are employed, including the scale-invariant logarithmic loss~\cite{eigen2014depth} $L_{silog}$, pair-wise normal regression loss~\cite{leres}$L_{\PWN}$, virtual normal loss~\cite{yin2021virtual} $L_{\VNL}$. Note $L_{silog}$ is a variant of L1 loss.  The overall losses are as follows.
\begin{eqnarray}
L_{d} = L_{\PWN} + L_{\VNL} + L_{silog} + L_{\RPNL}.
\label{eq: losses}
\end{eqnarray}

\noindent\textbf{Normal loss.} To supervise normal prediction, we employ two distinct loss functions depending on the availability of ground-truth (GT) normals $\mathbf{N}^\ast$. As presented in Fig.~\ref{fig: pipeline}, when GT normals are provided, we utilize an aleatoric uncertainty-aware loss \cite{bae2021estimating} ($L_n(\cdot)$) to supervise prediction $\mathbf{N}$. Alternatively, in the absence of GT normals, we propose a consistency loss $L_{d-n}(\mathbf{D}, \mathbf{N})$ to align the predicted depth and normal. This loss is computed based on the similarity between a pseudo-normal map generated from the predicted depth using the least square method \cite{qi2018geonet}, and the predicted normal itself. Different from previous methods, \cite{bae2021estimating, qi2018geonet}, this loss operates as a self-supervision mechanism, requiring no depth or normal ground truth labels. Note that here we use the depth $\mathbf{D}$ in the real world instead of the one $\mathbf{D}_c$ in the canonical space to calculate depth-normal consistency. %
The overall losses are as follows.
\begin{eqnarray}
L = w_dL_d(\mathbf{D}_c, \mathbf{D}_c^\ast )+w_nL_n(\mathbf{N}, \mathbf{N^\ast})+w_{d-n}L_{d-n}(\mathbf{N}, \mathbf{D})
\label{eq: losses_final}
\end{eqnarray},
where $w_d=0.5$, $w_n=1$, $w_{d-n}=0.01$ serve as weights to balance the loss items.

\vspace{-1.0em}

%% file: exps.tex
\begin{table}[!t]
\caption{Quantitative comparison on NYUv2 and KITTI \textbf{metric depth} benchmarks. Methods \colorbox{gray!25}{overfitting the bench-} \colorbox{gray!25}{mark} are marked with grey, while \colorbox{blue!20}{robust depth estimation} methods are in blue. `ZS' denotes the zero-shot testing, and `FT' means the method is further finetuned on the benchmark. 
Among all zero-shot testing (ZS) results, our methods performs the best and is even better than overfitting methods. Further fine-tuning (FT) helps our method surpass all known methods, ranked by the averaged ranking among all metrics. Best results are in \textbf{bold} and second bests are \underline{underlined}. }
\vspace{0.2 em}
\begin{threeparttable}
\scalebox{0.7}{
\begin{tabular}{r |cccccc}
\toprule[1pt]
\multirow{1}{*}{Method} & $\boldsymbol{\delta_{1}}$$\uparrow$ & $\boldsymbol{\delta_{2}}$$\uparrow$ & $\boldsymbol{\delta_{3}}$$\uparrow$ & \textbf{AbsRel}$\downarrow$ & \textbf{log10}$\downarrow$ & \textbf{RMS}$\downarrow$  \\ \hline
\multicolumn{7}{c}{\textbf{NYUv2 Metric Depth Benchmark}} \\ \hline
\rowcolor{gray!15} Li \etal.~\cite{li2017two}               & $0.788$    & $0.958$    & $0.991$  & $0.143$   & $0.063$    & $0.635$     \\
\rowcolor{gray!15} Laina \etal.~\cite{laina2016deeper}      & $0.811$    & $0.953$    & $0.988$  & $0.127$   & $0.055$    & $0.573$       \\
\rowcolor{gray!15} VNL ~\cite{Yin2019enforcing}            & $0.875$   & $0.976$    & $0.994$  & $0.108$   & $0.048$    & $0.416$    \\ 
\rowcolor{gray!15} TrDepth~\cite{yang2021transformers}     & $0.900$   & $0.983$    & $0.996$  & $0.106$  & $0.045$     & $0.365$   \\
\rowcolor{gray!15} Adabins~\cite{bhat2021adabins}         & $0.903$    & ${0.984}$  & $0.997$  & $0.103$  & $0.044$     & $0.364$    \\
\rowcolor{gray!15} NeWCRFs~\cite{yuan2022new}              
& ${0.922}$  & $0.992$  & $0.998$ 
& $0.095$  & $0.041$     & $0.334$   \\
\rowcolor{gray!15} IEBins~\cite{shao2023iebins}              
& $0.936$  & $0.992$     & $0.998$& $0.087$  & $0.038$  & $0.314$ 
\\

\rowcolor{blue!20}ZeroDepth~\cite{guizilini2023towards}  ZS              
& $0.901$  & $0.961$     & - & $0.100$  & -  & $0.380$    \\
\rowcolor{blue!20}Polymax~\cite{yang2024polymax} ZS
& $0.969$  & ${0.996}$  & ${0.999}$ 
& ${0.067}$   & ${0.029}$   & ${0.250}$  \\

\rowcolor{blue!20}ZoeDepth~\cite{bhat2023zoedepth} FT              
& $0.953$  & $0.995$     & $0.999$ & $0.077$  & $0.033$  & $0.277$ 
  \\
\rowcolor{blue!20}ZeroDepth~\cite{guizilini2023towards} FT              
& $0.954$  & $0.995$     & $1.000$ & $0.074$  & $0.103$  & $0.269$ 
   \\
\rowcolor{blue!20}DepthAnything~\cite{depthanything} FT
& ${0.984}$   & $\boldsymbol{0.998}$   & $\boldsymbol{1.000}$ & ${0.056}$   & ${0.024}$ & ${0.206}$
  \\
\hline
\rowcolor{blue!20}
Ours Conv-L CSTM\_image ZS    
& $0.925$  & $0.983$  & $0.994$ 
& $0.092$   & $0.040$   & $0.341$  \\
\rowcolor{blue!20}
Ours Conv-L CSTM\_label ZS   
& $0.944$  & $0.986$  & $0.995$ 
& $0.083$   & $0.035$   & $0.310$  \\
\rowcolor{blue!20}
Ours ViT-L CSTM\_label ZS  
& ${0.975}$   & $0.994$   & ${0.998}$  & ${0.063}$  & ${0.028}$  & ${0.251}$  \\
\rowcolor{blue!20}
Ours ViT-g CSTM\_label ZS
& ${0.980}$   & ${0.997}$   & ${0.999}$  
& ${0.067}$  & ${0.030}$     & $0.260$ \\
\rowcolor{blue!20}
Ours ViT-L CSTM\_label FT
& $\boldsymbol{0.989}$   & $\boldsymbol{0.998}$   & $\boldsymbol{1.000}$ & $\underline{0.047}$  & $\underline{0.020}$  & $\boldsymbol{0.183}$ 
  \\
  \rowcolor{blue!20}
Ours ViT-g CSTM\_label FT
& $\underline{0.987}$   & ${0.997}$   & $\underline{0.999}$  
& $\boldsymbol{0.045}$  & $\boldsymbol{0.015}$     & $\underline{0.187}$

\\ 
\hline 

\hline  
\multicolumn{7}{c}{\textbf{}} \\ 

\midrule[1pt]
\multirow{1}{*}{Method} & $\boldsymbol{\delta_{1}}$$\uparrow$ & $\boldsymbol{\delta_{2}}$$\uparrow$ & $\boldsymbol{\delta_{3}}$$\uparrow$ & \textbf{AbsRel} $\downarrow$ & \textbf{RMS} $\downarrow$ & \textbf{RMS\_log} $\downarrow$ \\ \hline
\multicolumn{7}{c}{\textbf{KITTI Metric Depth Benchmark}} \\ \hline 

\rowcolor{gray!15}Guo \etal \cite{guo2018learning}  & $0.902$  & $0.969$  & $0.986$ & $0.090$ & $3.258$ & $0.168$    \\
\rowcolor{gray!15}VNL~\cite{Yin2019enforcing} & ${0.938}$   & ${0.990}$   & ${0.998}$  & ${0.072}$  & $3.258$      & ${0.117}$    \\ 
\rowcolor{gray!15}TrDepth~\cite{yang2021transformers}   & $0.956$  & $0.994$  & $0.999$   & $0.064$  & $2.755$  & $0.098$  \\
\rowcolor{gray!15}Adabins~\cite{bhat2021adabins} & $0.964$  & $0.995$  & $0.999$   & $0.058$  & $2.360$  & $0.088$  \\
\rowcolor{gray!15}NeWCRFs~\cite{yuan2022new} 
& $0.974$  & $0.997$  & $0.999$   & $0.052$  & $2.129$  & $0.079$  \\
\rowcolor{gray!15}IEBins~\cite{shao2023iebins}              
& $0.978$  & $0.998$     & $0.999$  & ${0.050}$  & $2.011$  & $0.075$ 
\\
\rowcolor{blue!20} ZeroDepth~\cite{guizilini2023towards} ZS              
& $0.910$  & $0.980$     & $0.996$   & ${0.102}$  & $4.044$  & $0.172$ \\

\rowcolor{blue!20} ZoeDepth~\cite{bhat2023zoedepth} FT              
& $0.971$  & $0.996$     & $0.999$ & ${0.057}$  & $2.281$  & $0.082$ 
    \\
\rowcolor{blue!20} ZeroDepth~\cite{guizilini2023towards} FT              
& $0.968$  & $0.995$     & $0.999$   & $0.053$  & $2.087$  & $0.083$ 
 \\
\rowcolor{blue!20} DepthAnything~\cite{depthanything} FT  
  & ${0.982}$  & $\boldsymbol{0.998}$  & $\boldsymbol{1.000}$ & ${0.046}$   & $\underline{1.869}$   & ${0.069}$
\\ \hline
\rowcolor{blue!20}
Ours Conv-L CSTM\_image ZS & 
${0.967}$   & ${0.995}$   & ${0.999}$  
& $0.060$  & ${2.843}$     & ${0.087}$    \\ 
\rowcolor{blue!20}
Ours Conv-L CSTM\_label ZS
& ${0.964}$   & ${0.993}$   & ${0.998}$  
& $0.058$  & ${2.770}$     & ${0.092}$ \\
\rowcolor{blue!20}
Ours ViT-L CSTM\_label ZS
& ${0.974}$  & ${0.995}$     & ${0.999}$ & ${0.052}$   & ${2.511}$   & ${0.074}$   \\
\rowcolor{blue!20}
Ours ViT-g CSTM\_label ZS
& ${0.977}$   & ${0.996}$   & ${0.999}$  
& ${0.051}$  & ${2.403}$     & ${0.080}$ \\
\rowcolor{blue!20}
Ours ViT-L CSTM\_label FT
& $\underline{0.985}$   & $\boldsymbol{0.998}$   & $\underline{0.999}$  
& $\boldsymbol{0.044}$  & ${1.985}$     & $\underline{0.064}$\\
\rowcolor{blue!20}
Ours ViT-g CSTM\_label FT
& $\boldsymbol{0.989}$   & $\boldsymbol{0.998}$   & $\boldsymbol{1.000}$  
& $\boldsymbol{0.039}$  & $\boldsymbol{1.766}$     & $\boldsymbol{0.060}$
 \\ \hline
\toprule[1pt]
\end{tabular}\newline}
\end{threeparttable}
\label{table:errors cmp on NYUD-V2}
\vspace{-2 em}
\end{table}

\noindent\textbf{Dataset details.}
\label{sec:data}
We have meticulously assembled a comprehensive dataset incorporating 16 publicly available RGB-D datasets, comprising a cumulative total of over 16 million data points specifically intended for training purposes. This dataset encompasses a diverse array of both indoor and outdoor scenes. Notably, approximately 10 million frames within the dataset are annotated with normals, with a predominant focus on annotations relating to indoor scenes. It is noteworthy to highlight that all datasets have provided camera intrinsic parameters. Additionally, beyond the test split of training datasets, we have procured 7 previously unobserved datasets to facilitate robustness and generalization evaluations. Detailed descriptions of the utilized training and testing data are provided in Table \ref{table: datasetsv2}.

\begin{table}[t]
	\caption{Quantitative comparison of surface normals on \textbf{NYUv2, ibims-1, and ScanNet normal} benchmarks. `ZS' means zero-shot testing and `FT' performs post fine-tuneing on the target dataset. Methods \colorbox{gray!25}{trained only on NYU} are highlighted with grey. Best results are in \textbf{bold} and second bests are \underline{underlined}. Our method ranks first over all benchmarks.}
	\vspace{0.2 em}
	\begin{threeparttable}
	\scalebox{0.7}{
		\begin{tabular}{r |cccccc}
			\toprule[1pt]
			\multirow{1}{*}{Method} & $\boldsymbol{11.25^{\circ}}$$\uparrow$ & $\boldsymbol{22.5^{\circ}}$$\uparrow$ & $\boldsymbol{30^{\circ}}$$\uparrow$ & \textbf{mean}$\downarrow$ & \textbf{median}$\downarrow$ & \textbf{RMS\_normal}$\downarrow$  \\ \hline			
			\multicolumn{7}{c}{\textbf{NYUv2 Normal Benchmark}} \\ \hline			
			\hline
			\rowcolor{gray!20} Ladicky \etal ~\cite{ladicky2014discriminatively}           & 0.275    & 0.490    & 0.587  & 33.5   & 23.1    & -     \\
			\rowcolor{gray!20} Fouhey \etal.~\cite{fouhey2014unfolding}      &  0.405    & 0.541    & 0.589  & 35.2   & 17.9    & -       \\
			\rowcolor{gray!20} Deep3D ~\cite{wang2015designing}            & 0.420   & 0.612    & 0.682  & 20.9   & 13.2    & -    \\ 
			\rowcolor{gray!20} Eigen \etal ~\cite{eigen2015predicting}     & 0.444   & 0.672    & 0.759  & 20.9  & 13.2     & -  \\
			\rowcolor{gray!20} SkipNet ~\cite{bansal2016marr}          & 0.479   & 0.700  & 0.778  & 19.8  & 12.0     & 28.2   \\
			\rowcolor{gray!20} SURGE ~\cite{wang2016surge}              
			& 0.473  & 0.689 & 0.766 
			& 20.6  &   12.2   & -   \\
			\rowcolor{gray!20} GeoNet ~\cite{qi2018geonet}              
			& 0.484  & 0.715  & 0.795 
			& 19.0  & 11.8     & 26.9   \\
			\rowcolor{gray!20} PAP ~\cite{zhang2019pattern}              
			& 0.488  & 0.722  & 0.798 
			&  18.6 & 11.7     & 25.5   \\
			\rowcolor{gray!20} GeoNet++ ~\cite{qi2020geonet++}    
			& $0.502$  & $0.732$  & $0.807$ 
			& $18.5$   & $11.2$   & $26.7$  \\
			\rowcolor{gray!20}  Bae \etal~\cite{bae2021estimating} 
			& 0.622   & 0.793   & 0.852 & 14.9  & 7.5  &  23.5 \\
			FrameNet~\cite{huang2019framenet} ZS              
			& $0.507$ & $0.720$ & $0.795$ & $18.6$  & $11.0$ & $26.8$   \\
			VPLNet~\cite{wang2020vplnet} ZS    
			& $0.543$   & $0.738$   & $0.807$ & $18.0$  & $9.8$  & -  \\
			TiltedSN~\cite{do2020surface} ZS    
			& $0.598$   & $0.774$   & $0.834$ & $16.1$  & $8.1$  & $25.1$  \\
			\red{Omnidata\cite{eftekhar2021omnidata} ZS} 
			& \red{$0.577$}   & \red{$0.777$}   & \red{$0.838$} & \red{$16.7$}  & \red{$9.6$}  & \red{25.0} \\
			Bae \etal~\cite{bae2021estimating} ZS 
			& $0.597$   & $0.775$   & $0.837$ & $16.0$  & $8.4$  & $24.7$  \\
			Polymax~\cite{yang2024polymax} ZS
			& ${0.656}$   & ${0.822}$  & ${0.878}$ & $\underline{13.1}$  & ${7.1}$  & ${20.4}$ \\					
			\hline
			Ours ViT-L CSTM\_label ZS    
			& ${0.662}$  & ${0.831}$  & ${0.881}$ & $\underline{13.1}$   & ${7.1}$   & ${21.1}$  \\
			Ours ViT-g CSTM\_label ZS 
			& $\underline{0.664}$    & ${0.831}$    & ${0.881}$   
			& ${13.3}$   & $\underline{7.0}$      & $21.3$  \\	
			
			Ours ViT-L CSTM\_label FT    
			& $\boldsymbol{0.688}$   & $\boldsymbol{0.849}$   & $\boldsymbol{0.898}$ 
			& $\boldsymbol{12.0}$   & $\boldsymbol{6.5}$    & $\boldsymbol{19.2}$   \\
			Ours ViT-g CSTM\_label FT 
			& 0.662  & $\underline{0.837}$   & $\underline{0.889}$ 
			& 13.2  & 7.5    & $\underline{20.2}$ \\
			\hline 
			
			\hline  
			\multicolumn{7}{c}{\textbf{}}		
			\\			
			
			\midrule[1pt]
			\multicolumn{7}{c}{\textbf{ibims-1 Normal Benchmark}} \\ \hline 
			VNL~\cite{Yin2019enforcing} ZS & 0.179   &  0.386  & 0.494  & 39.8 & 30.4  & 51.0    \\
			BTS~\cite{bts} ZS   & 0.130  & 0.295  & 0.400  & 44.0 & 37.8  & 53.5  \\
			Adabins~\cite{bhat2021adabins} ZS & 0.180  & 0.387  & 0.506   & 37.1  & 29.6  & 46.9  \\
			IronDepth~\cite{bae2022irondepth} ZS              
			& 0.431  & 0.639  & 0.716   & 25.3  & 14.2  & 37.4 \\
            \red{Omnidata\cite{kar20223d} ZS} 
			& \red{$0.647$}   & \red{$0.734$}   & \red{$0.768$} & \red{$20.8$}  & \red{$7.7$}  & \red{\underline{35.1}} \\
   
			\hline
			Ours ViT-L CSTM\_label ZS    
			& $\underline{0.694}$  & $\underline{0.758}$   & $\underline{0.785}$ & $\boldsymbol{19.4}$  & $\boldsymbol{5.7}$ &  $\boldsymbol{34.9}$ \\
			Ours ViT-g CSTM\_label ZS 
			& $\boldsymbol{0.697}$   & $\boldsymbol{0.762}$  & $\boldsymbol{0.788}$  
			& $\underline{19.6}$ & $\boldsymbol{5.7}$    & $35.2$ \\
			\hline 

			\hline  
			\multicolumn{7}{c}{\textbf{}}		
			\\				
			\midrule[1pt]		 
			\multicolumn{7}{c}{\textbf{ScanNet Normal Benchmark}} \\ \hline 
   			\red{Omnidata\cite{kar20223d}} 
			& \red{$0.629$}   & \red{$0.806$}   & \red{$0.847$} & \red{$15.1$}  & \red{$8.6$}  & \red{23.1} \\
			FrameNet~\cite{huang2019framenet}              
			& $0.625$  & $0.801$  & $0.858$ 
			& $14.7$  & $7.7$     & $22.8$   \\
			VPLNet~\cite{wang2020vplnet}    
			& $0.663$  & $0.818$  & $0.870$ 
			& $12.6$   & $6.0$   & $21.1$  \\
			TiltedSN~\cite{do2020surface}    
			& $0.693$  & $0.839$  & $0.886$ 
			& $12.6$   & $6.0$   & $21.1$  \\
			Bae \etal~\cite{bae2021estimating} 
			& $0.711$  & $0.854$  & $0.898$ 
			& $11.8$   & $5.7$   & $20.0$	
			\\ \hline
			Ours ViT-L CSTM\_label 
			 & $\underline{0.760}$   & $\underline{0.885}$   & $\underline{0.923}$  
 			& $\underline{9.9}$  & $\underline{5.3}$    & $\underline{16.4}$
			 \\
			Ours ViT-g CSTM\_label 
			& $\boldsymbol{0.778}$   & $\boldsymbol{0.901}$   & $\boldsymbol{0.935}$ 
			& $\boldsymbol{9.2}$   & $\boldsymbol{5.0}$   & $\boldsymbol{15.3}$ 
			\\ \hline
			\toprule[1pt]			
		\end{tabular}\newline}
\end{threeparttable}
	\label{table:errors cmp on NYUN-V2}
	\vspace{-2 em}
\end{table}

\noindent\textbf{Implementation details.}
In our experiments, we employ different network architectures and aim to provide diverse choices for the community, including convnets and transformers. For convnets, we employ an UNet architecture with the ConvNext-large~\cite{liu2022convnet} backbone. ImageNet-22K pre-trained weights are used for initialization. For transformers, we apply DINO v2-reg \cite{oquab2023dinov2, darcet2023vision} vision transformers \cite{dosovitskiy2020an} (ViT) as backbones, DPT \cite{ranftl2021vision} as decoders.%

We use AdamW with a batch size of $192$, an initial learning rate $0.0001$ for all layers, and the polynomial decaying method with the power of $0.9$. We train our models on $48$ A100 GPUs for %
$800$k iterations. Following the DiverseDepth~\cite{yin2021virtual}, we balance all datasets in a mini-batch to ensure each dataset accounts for an almost equal ratio. During training, images are processed by the canonical camera transformation module, flipped horizontally with a $50\%$ chance, and then randomly cropped into 
$512 \times 960$ pixels for convnets and $616 \times 1064$ for vision transformers. In the ablation experiments, training settings are different as we sample $5000$ images from each dataset for training. We trained on $8$ GPUs for $150$K iterations. Details of networks architectures, training setups, and efficiency analysis are presented in the supplementary materials. Fine-tuning experiments on KITTI and NYU are conducted on $8$ GPUs with $20$K further steps. 

\begin{table*}[]
\centering
 \caption{Quantitative comparison with SoTA metric depth methods on $5$ unseen benchmarks. For SoTA methods, we use their NYUv2 and KITTI models for indoor and outdoor scene evaluation respectively, while we use the same model for all zero-shot testing.}
 \vspace{0.2 em}
 \begin{threeparttable}
 \resizebox{0.95\linewidth}{!}{%
\begin{tabular}{l|l|ll|lll}
\toprule[0.2pt]
\multirow{2}{*}{Method} & \multirow{2}{*}{Metric Head}       & DIODE(Indoor) & iBIMS-1      & DIODE(Outdoor)      & ETH3D      & NuScenes     \\
        & & \multicolumn{2}{c|}{Indoor scenes (AbsRel$\downarrow$/RMS$\downarrow$)}  & \multicolumn{3}{c}{Outdoor scenes (AbsRel$\downarrow$/RMS$\downarrow$)} \\ \hline
Adabins~\cite{bhat2021adabins}  
         & KITTI or NYU \tnote{\dag}
         &  0.443 / 1.963       
         &0.212 / 0.901         
         &0.865 / 10.35                     
         &1.271 / 6.178            
         &0.445 / 10.658              \\
NewCRFs~\cite{yuan2022new}  
         & KITTI or NYU \tnote{\dag}
         &0.404 / 1.867               
         &0.206 / 0.861         
         &0.854 / 9.228                     
         &0.890 / 5.011            
         &0.400 / 12.139              \\ 

ZoeDepth~\cite{bhat2023zoedepth}  
& KITTI and NYU \tnote{\ddag}
&0.400 / 1.581               
&0.169 / 0.711         
&0.269 / 6.898                     
&0.545 / 3.112            
&0.504 / 7.717              \\ \hline

Ours Conv-L CSTM\_label    
        & Unified
        &{0.252} / {1.440}               
        & \underline{0.160} / \textbf{0.521}         
        &{0.414} / {6.934}                     
        & {0.416} / {3.017}          
        & 0.154 / 7.097             \\
Ours Conv-L CSTM\_image    
        & Unified
        & 0.268 / 1.429         
        & \textbf{0.144} / {0.646}        
        & {0.535} / {6.507}                    
        & \textbf{0.342} / \textbf{2.965}           
        & \underline{0.147} / \textbf{5.889}    \\ 
        
Ours ViT-L CSTM\_image    
& Unified
& \textbf{0.093} / \underline{0.389}         
& 0.185 / \underline{0.592}        
& \underline{0.221}  / \underline{3.897}                    
& \underline{0.357} / \underline{2.980}           
& {0.165} / {9.001}    \\ 
Ours ViT-g CSTM\_image    
& Unified
& \underline{0.081} / \textbf{0.359}         
& {0.249} / {0.611}        
& \textbf{0.201} / \textbf{3.671}                    
& 0.363 / 2.999           
& \textbf{0.129} / \underline{6.993}    \\ 
           \bottomrule
\end{tabular}}
\begin{tablenotes}
\scriptsize
\item[\dag]: Two different metric heads are trained on KITTI and NYU respectively. \ddag: Both metric heads are ensembled by an additional router.
\end{tablenotes}
 \end{threeparttable}
\label{table: metric eval on more datasets.}
\vspace{-1.5 em}
\end{table*}

\begin{table*}[t]
\centering
\caption{
Comparison with SoTA affine-invariant depth methods on 5 zero-shot transfer benchmarks.
Our model significantly outperforms previous methods and sets new state-of-the-art. Following the benchmark setting, all methods have manually aligned the scale and shift. 
}
\vspace{0.2 em}
\begin{threeparttable}
\setlength{\tabcolsep}{2pt}
\resizebox{0.96\linewidth}{!}{%
\begin{tabular}{ r |ll|ll|ll|ll|ll|ll|ll}
\toprule[1pt]
\multirow{2}{*}{Method} & \multirow{2}{*}{Backbone} & \multirow{2}{*}{\#Params} &\multicolumn{2}{c|}{\#Data} & \multicolumn{2}{c|}{NYUv2} & \multicolumn{2}{c|}{KITTI} & \multicolumn{2}{c|}{DIODE(Full)} & \multicolumn{2}{c|}{ScanNet} & \multicolumn{2}{c}{ETH3D} \\
&  &  & Pretrain & Train & AbsRel$\downarrow$     & $\delta_{1}$$\uparrow$     & AbsRel$\downarrow$      & $\delta_{1}$$\uparrow$      & AbsRel$\downarrow$      & $\delta_{1}$$\uparrow$      &AbsRel$\downarrow$      & $\delta_{1}$$\uparrow$       &AbsRel$\downarrow$     & $\delta_{1}$$\uparrow$  \\ \hline
DiverseDepth~\cite{yin2021virtual}& ResNeXt50~\cite{xie2017aggregated}& 25M  
& 1.3M &320K 
&0.117 &0.875 
&0.190 &0.704 
&0.376 &0.631 
&0.108 &0.882 
&0.228 &0.694  \\
MiDaS~\cite{Ranftl2020}& ResNeXt101&  88M %
& 1.3M & 2M 
&0.111 &0.885 
&0.236 &0.630 
&0.332 &0.715 
&0.111 &0.886  
&0.184 &0.752 \\
Leres~\cite{leres} & ResNeXt101&  %
& 1.3M & 354K 
&0.090  &0.916  
&0.149 &0.784 
&0.271 &0.766 
&0.095 &0.912 
&0.171 &0.777  \\
Omnidata~\cite{eftekhar2021omnidata} & ViT-Base& %
& 1.3M & 12.2M 
& 0.074 & 0.945 
& 0.149 & 0.835
& 0.339 & 0.742 
& 0.077 & 0.935 
& 0.166 & 0.778  \\
HDN~\cite{zhang2022hierarchical} & ViT-Large~\cite{dosovitskiy2020an}&  306M  
& 1.3M & 300K 
&0.069 &0.948  
&0.115 &0.867 
&0.246 &0.780 
&0.080 &0.939 
&0.121 &0.833 \\
DPT-large~\cite{ranftl2021vision} & ViT-Large& %
& 1.3M & 188K 
& 0.098 & 0.903 
& 0.100 & 0.901
& 0.182 & 0.758 
& 0.078 & 0.938 
& 0.078 & 0.946 \\
DepthAnything~\cite{ranftl2021vision} & ViT-Large& %
& 142M & 63.5M
& 0.043 & \textbf{0.981} 
& 0.076 & 0.947
& 0.277 & 0.759 
& \textbf{0.042} & \textbf{0.980} 
& 0.127 & 0.882  \\
Marigold~\cite{ranftl2021vision} & Latent diffusion V2\cite{rombach2022high}& 899M %
& 5B & 74K
& 0.055 & 0.961 
& 0.099 & 0.916
& 0.308 & 0.773
& \underline{0.064} & \underline{0.951} 
& 0.065 & 0.960  \\
\hline
Ours CSTM\_label & ViT-Small &  22M
& 142M & 16M
& 0.056 & 0.965   
& 0.064 & 0.950
& 0.247 & 0.789 
& \st{0.033}\tnote{\dag} & \st{0.985}\tnote{\dag}
& 0.062 & 0.955 \\
Ours CSTM\_image & ConvNeXt-Large~\cite{liu2022convnet}&  198M 
& 14.2M & 8M
&0.058 &0.963  
&0.053 &0.965 
&0.211 &0.825 
&0.074 &0.942
&0.064 &0.965  \\ 
Ours CSTM\_label & ConvNeXt-Large&   
& 14.2M & 8M
&0.050 &0.966  
&0.058 &0.970 
&0.224 &0.805 
&0.074 &0.941 
&0.066 &0.964  \\
Ours CSTM\_label & ViT-Large& 306M
& 142M & 16M
&\textbf{0.042} &\underline{0.980}  
&\underline{0.046} &\underline{0.979} 
&\underline{0.141} &\underline{0.882} 
&\st{0.021}\tnote{\dag} &\st{0.993}\tnote{\dag} 
&\textbf{0.042} &\textbf{0.987}  \\
Ours CSTM\_label & ViT-giant\cite{zhai2022scaling}& 1011M  
& 142M & 16M
&\underline{0.043} &\textbf{0.981} 
&\textbf{0.044} &\textbf{0.982} 
&\textbf{0.136} &\textbf{0.895} 
&\st{0.022}\tnote{\dag} &\st{0.994}\tnote{\dag} 
&\textbf{0.042} &\underline{0.983}  \\

 \toprule[1pt]
\end{tabular}}
\begin{tablenotes}
\scriptsize
\item[\dag]: ScanNet is partly annotated with normal \cite{huang2019framenet}. For samples without normal annotations, these models use depth labels to facilitate normal learning.
\end{tablenotes}
\end{threeparttable}
\label{Table: generalization evaluation.}
\vspace{-1.5em}
\end{table*}

\noindent\textbf{Evaluation details for monocular depth and normal estimation.}
\blue{a) To demonstrate the robustness of our metric depth estimation method, we evaluate on 7 zero-shot benchmarks, including NYUv2, KITTI~\cite{geiger2012we}, ScanNet~\cite{dai2017scannet}, NuScenes~\cite{caesar2020nuscenes}, iBIMS-1\cite{koch2018evaluation}, and DIODE~\cite{vasiljevic2019diode} (both indoor and outdoor). Following previous studies, we use metrics such as absolute relative error (AbsRel), accuracy under threshold ($\delta_{i} < 1.25^{i}, i=1, 2, 3$), root mean squared error (RMS), root mean squared error in log space (RMS\_{log}), and log10 error (log10). We report results for zero-shot and fine-tuning testing on the KITTI and NYU benchmarks.}
b) For normal estimation tasks and ablations,  employ several error metrics to assess performance. Specifically, we calculate the mean (\textbf{mean}), median (\textbf{median}), and rooted mean square  (\textbf{RMS\_normal}) of the angular error as well as the accuracy under threshold of \{$11.25^\circ$, $22.5^\circ$, $30.0^\circ$\} consistent with methodologies established in previous studies ~\cite{bae2021estimating}. We conduct in-domain evaluation using the Scannet dataset, while the NYU and iBIMS-1 datasets are reserved for zero-shot generalization testing. 
c) Furthermore, %
we also follow current affine-invariant depth benchmarks~\cite{leres, zhang2022hierarchical} (Tab. \ref{Table: generalization evaluation.}) to evaluate the generalization ability on $5$ zero-shot datasets, \textit{i.e.},  NYUv2, DIODE, ETH3D, ScanNet~\cite{dai2017scannet}, and KITTI. We mainly compare with large-scale data trained models. Note that in this benchmark we follow existing methods to apply the scale shift alignment before evaluation. 

We report results with different canonical transformation methods (CSTM\_lable and CSTM\_image) on the ConvNext-Large model (Conv-L in Tab.~\ref{table:errors cmp on NYUD-V2} and Tab.~\ref{table:errors cmp on NYUN-V2}). As CSTM\_label is slightly better, more results using this method from multi-size ViT-models (ViT-S for Small, ViT-L for Large, ViT-g for giant2) are reported. Note that all models for zero-shot testing use the same checkpoints except for fine-tuning experiments.

\noindent\textbf{Evaluation details for reconstruction and SLAM.}
a) To evaluate our metric 3D reconstruction quality, we randomly sample 9 unseen scenes from NYUv2 and use colmap~\cite{schoenberger2016mvs} to obtain the camera poses for multi-frame reconstruction. Chamfer $l_1$ distance and the F-score~\cite{knapitsch2017tanks} are used to evaluate the reconstruction accuracy. 
b) In dense-SLAM experiments, following Li~\etal~\cite{li2021generalizing}, we test on the KITTI odometry benchmark~\cite{Geiger2013IJRR} and evaluate the average translational RMS($\%, t_{rel}$) and rotational RMS (${}^{\circ}/100m, r_{rel}$) errors~\cite{Geiger2013IJRR}.

\noindent\textbf{Evaluation on metric depth benchmarks.} To evaluate the accuracy of predicted metric depth, firstly,  we compare with state-of-the-art (SoTA) metric depth prediction methods on NYUv2~\cite{silberman2012indoor}, KITTI~\cite{geiger2012we}.
We use the same model to do %
all evaluations. %
Results are reported in Tab.~\ref{table:errors cmp on NYUD-V2}. Firstly, comparing with existing overfitting methods, which are trained on benchmarks for hundreds of epochs, our zero-shot testing (`ZS' in the table) without any fine-tuning or metric adjustment already achieves comparable or even better performance on some metrics. Then comparing with robust monocular depth estimation methods, such as Zerodepth~\cite{guizilini2023towards} and ZoeDepth~\cite{bhat2023zoedepth}, our zero-shot testing is also better than them. Further post finetuning (`FT in the table') lifts our method to the 1st rank. %

Furthermore, We collect $5$ unseen datasets to do more metric accuracy evaluation. These datasets contain a wide range of indoor and outdoor scenes, including rooms, buildings, and driving scenes. The camera models are also varied.
We mainly compare with the SoTA metric depth estimation methods and take their NYUv2 and KITTI models for indoor and outdoor scene evaluation respectively. From Tab. \ref{table: metric eval on more datasets.}, we observe that although %
NuScenes is similar to KITTI, existing methods face a noticeable performance decrease. In contrast, our model is more robust. %
 
\noindent\textbf{Generalization over diverse scenes.}
Affine-invariant depth benchmarks decouple the scale's effect, which aims to evaluate the model's generalization ability to diverse scenes. Recent impact works, such as MiDaS, LeReS, DPT, \red{Marigold, and DepthAnything} achieved promising performance on them. Following them, we test on 5 datasets and manually align the scale and shift to the ground-truth depth before evaluation. Results are reported in Tab.~\ref{Table: generalization evaluation.}. Although our method enforces the network to recover the more challenging metric, our method outperforms them on all datasets. 

\begin{figure*}[]
	\centering
	\includegraphics[width=0.98\textwidth]{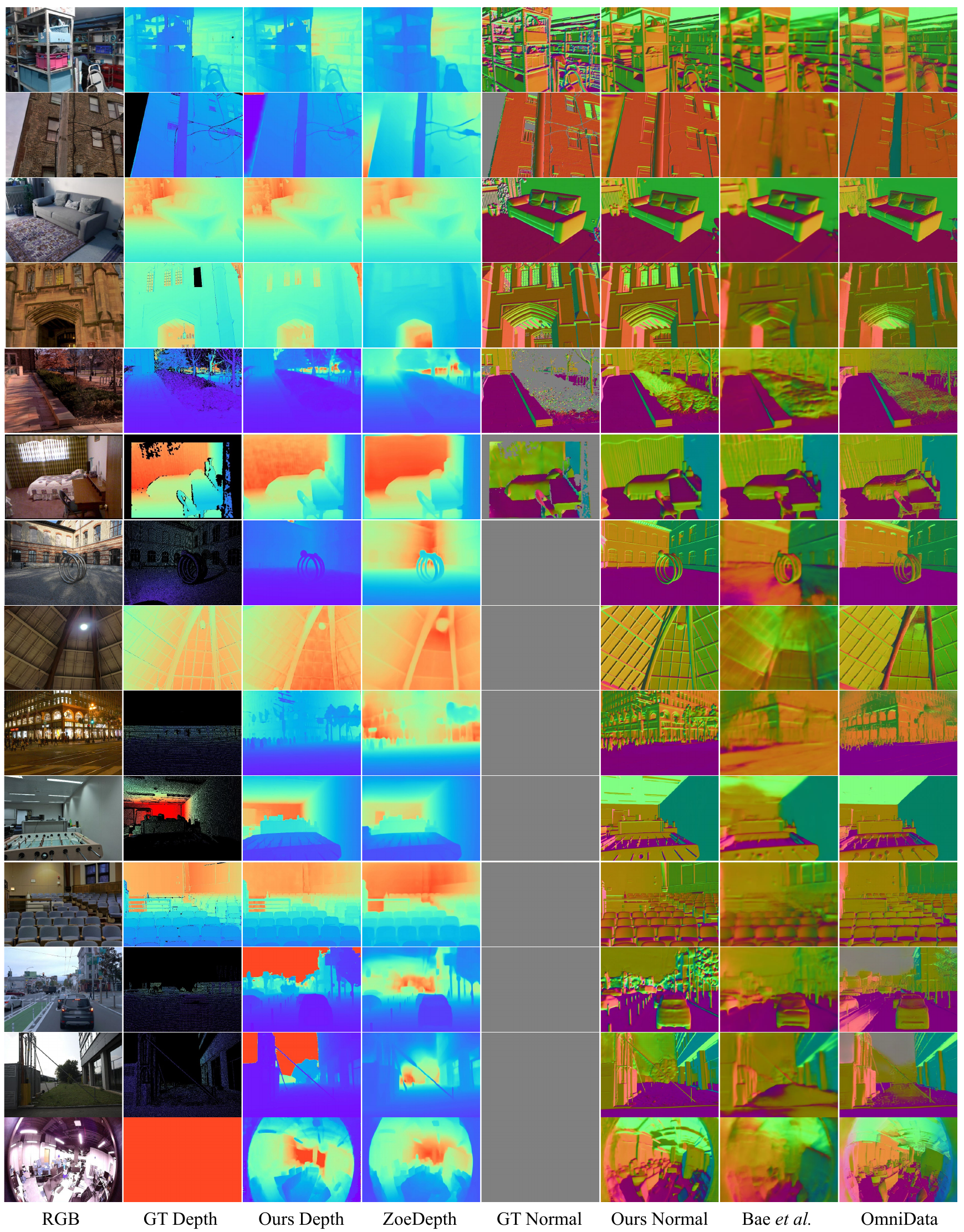}
	\vspace{-0.5 em}
	\caption{\textbf{Qualitative comparisons of metric depth and surface normals for iBims, DIODE, NYU, Eth3d, Nuscenes, and self-collected drone datasets.} We present visualization results of our predictions (`Ours Depth' / `Ours Normal'), groundtruth labels (`GT Depth' / `GT Normal') and results from other metric depth (`ZoeDepth'\cite{bhat2023zoedepth}) and surface normal methods (`Bae \etal' \cite{bae2021estimating} and `OmniData'\cite{eftekhar2021omnidata}).}
	\label{fig: visualization.}
	\vspace{-1em}
\end{figure*}

\begin{figure*}[]
	\centering
	\includegraphics[width=0.90\textwidth]{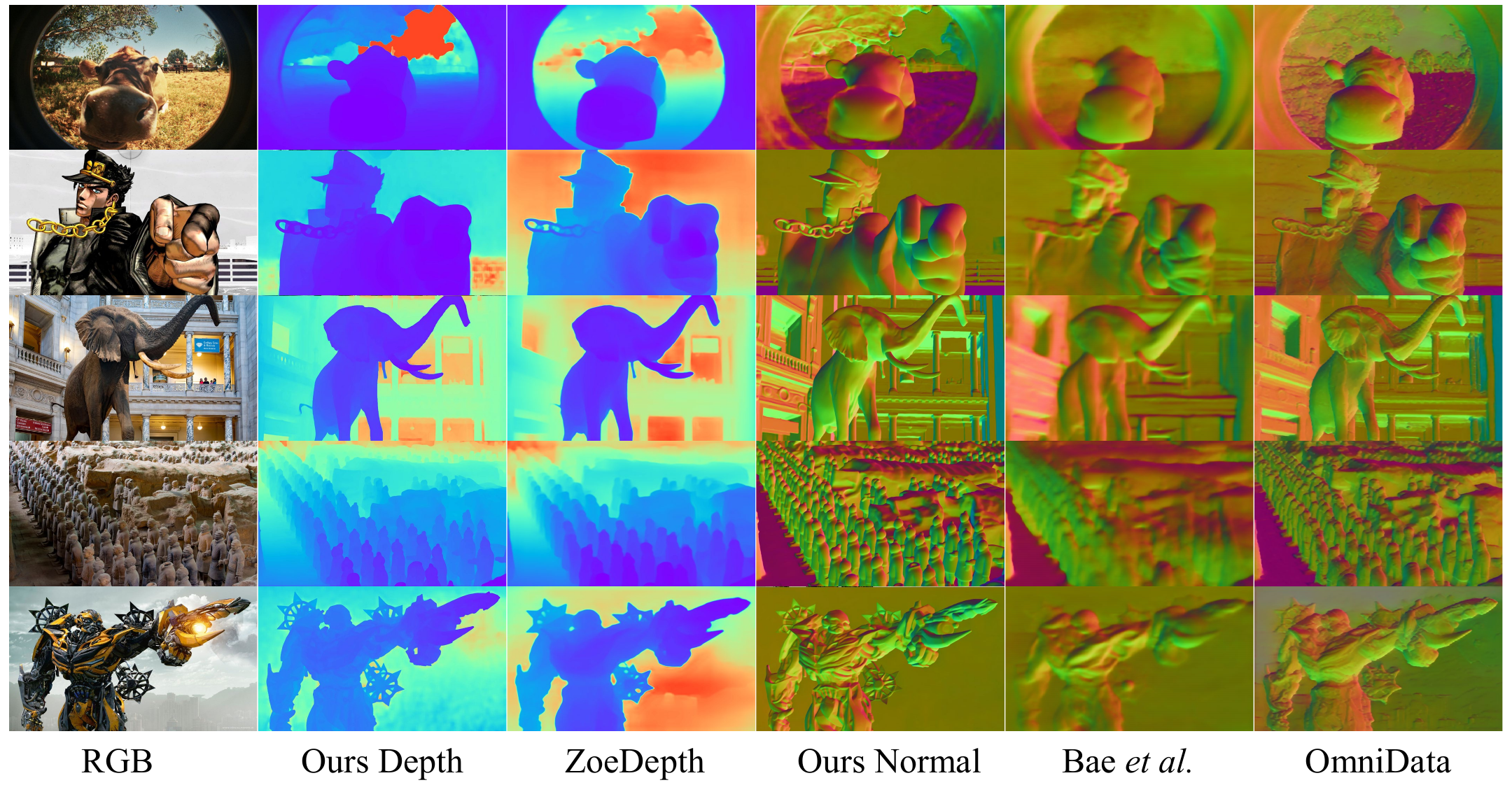}
	\vspace{-1em}
	\caption{\textbf{Qualitative comparisons of metric depth and surface normals in the wild.} We present visualization results of our predictions (`Ours Depth' / `Ours Normal') and results from other metric depth (`ZoeDepth'\cite{bhat2023zoedepth}) and surface normal methods (`Bae \etal' \cite{bae2021estimating} and `OmniData'\cite{eftekhar2021omnidata}).}
	\label{fig: visualization wild.}
	\vspace{-1em}
\end{figure*}

\noindent\textbf{Evaluation on surface normal benchmarks.} We evaluate our methods on ScanNet, NYU, and iBims-1 surface normal benchmarks. Results are reported in Tab.~\ref{table:errors cmp on NYUN-V2}. %
Firstly, we organize a zero-shot testing benchmark on NYU dataset, see methods denoted with `ZS' in the table. We compare with existing methods which are trained on ScanNet or Taskonomy and have achieved promising performance on them, such as Polymax~\cite{yang2024polymax} and Bae~\etal~\cite{bae2021estimating}. 
Our method surpasses them over most metrics. Cmparing with methods that have been overfitted the NYU data domain for hundreds of epochs (marked with blue), our zero-shot testing outperforms them on all metrics. Our post-finetuned models (`FT' marks) further boost the performance. 
Similarly, we also achieve SoTA performance on iBims-1 and Scannet benchmarks. For the iBims-1 dataset, we follow IronDepth\cite{bae2022irondepth} to generate the ground-truth normal annotations.

\vspace{-1em}

\begin{table}[!b]
\vspace{-2 em}
	\caption{Training and testing datasets used for experiments.}
	\centering
	\begin{threeparttable}
		\scalebox{0.7}{
			\begin{tabular}{ r llllll}
				\toprule[1pt]
				\multicolumn{1}{l|}{Datasets}                     & \multicolumn{1}{l|}{Scenes}    & \multicolumn{1}{l|}{Source}     & \multicolumn{1}{l|}{%
					Label}     &     \multicolumn{1}{l|}{Size}     & \#Cam.          \\ \hline
				\multicolumn{6}{c}{Training Data}                                                                                                                                                          \\ \hline
				\multicolumn{1}{l|}{DDAD~\cite{packnet}}                           & \multicolumn{1}{l|}{Outdoor}     &  \multicolumn{1}{l|}{Real-world}      &  \multicolumn{1}{l|}{Depth}                 & \multicolumn{1}{l|}{$\sim$80K}     & 36+           \\
				\multicolumn{1}{l|}{Lyft~\cite{lyftl5preception}}                           & \multicolumn{1}{l|}{Outdoor}        & 
				\multicolumn{1}{l|}{Real-world}      &  \multicolumn{1}{l|}{Depth}                 & \multicolumn{1}{l|}{$\sim$50K} & 6+              \\
				\multicolumn{1}{l|}{Driving Stereo (DS)~\cite{yang2019drivingstereo}}                 & \multicolumn{1}{l|}{Outdoor}        & \multicolumn{1}{l|}{Real-world}      & \multicolumn{1}{l|}{Depth}       & \multicolumn{1}{l|}{$\sim$181K}  & 1              \\
				\multicolumn{1}{l|}{DIML~\cite{cho2021diml}}                           & \multicolumn{1}{l|}{Outdoor}        & \multicolumn{1}{l|}{Real-world}      & \multicolumn{1}{l|}{Depth}       & \multicolumn{1}{l|}{$\sim$122K}       & 10        \\
				\multicolumn{1}{l|}{Arogoverse2~\cite{Argoverse2}}                    & \multicolumn{1}{l|}{Outdoor}        & \multicolumn{1}{l|}{Real-world}      & \multicolumn{1}{l|}{Depth}                 & \multicolumn{1}{l|}{$\sim$3515K}  & 6+           \\
				\multicolumn{1}{l|}{Cityscapes~\cite{Cordts2016Cityscapes}}                     & \multicolumn{1}{l|}{Outdoor}        & \multicolumn{1}{l|}{Real-world}      & \multicolumn{1}{l|}{Depth}       & \multicolumn{1}{l|}{$\sim$170K}      & 1         \\
				\multicolumn{1}{l|}{DSEC~\cite{Gehrig21ral}}                           & \multicolumn{1}{l|}{Outdoor}        & \multicolumn{1}{l|}{Real-world}      & \multicolumn{1}{l|}{Depth}                 & \multicolumn{1}{l|}{$\sim$26K} & 1                \\
				\multicolumn{1}{l|}{Mapillary PSD~\cite{MapillaryPSD}} & \multicolumn{1}{l|}{Outdoor}   &  \multicolumn{1}{l|}{Real-world}      & \multicolumn{1}{l|}{Depth} & \multicolumn{1}{l|}{750K} & 1000+               \\
				\multicolumn{1}{l|}{Pandaset~\cite{itsc21pandaset}}                       & \multicolumn{1}{l|}{Outdoor}        & \multicolumn{1}{l|}{Real-world}      & \multicolumn{1}{l|}{Depth}   & \multicolumn{1}{l|}{$\sim$48K} & 6                      \\
				\multicolumn{1}{l|}{UASOL~\cite{bauer2019uasol}}                          & \multicolumn{1}{l|}{Outdoor}        & \multicolumn{1}{l|}{Real-world}      & \multicolumn{1}{l|}{Depth}       & \multicolumn{1}{l|}{$\sim$1370K}          & 1             \\
				\multicolumn{1}{l|}{Virtual KITTI~\cite{cabon2020virtual}}                      & \multicolumn{1}{l|}{Outdoor}         & \multicolumn{1}{l|}{Synthesized}      & \multicolumn{1}{l|}{Depth}                 & \multicolumn{1}{l|}{37K}  & 2   \\
				\multicolumn{1}{l|}{Waymo~\cite{sun2020scalability}}                      & \multicolumn{1}{l|}{Outdoor}         & \multicolumn{1}{l|}{Real-world}      & \multicolumn{1}{l|}{Depth}                 & 				\multicolumn{1}{l|}{$\sim${1M}} & 5 \\
				\multicolumn{1}{l|}{Matterport3d~\cite{zamir2018taskonomy}}                      & \multicolumn{1}{l|}{In/Out}         & \multicolumn{1}{l|}{Real-world}      & \multicolumn{1}{l|}{Depth + Normal}                 & \multicolumn{1}{l|}{144K}  & 3 \\
				\multicolumn{1}{l|}{Taskonomy~\cite{zamir2018taskonomy}}                      & \multicolumn{1}{l|}{Indoor}         & \multicolumn{1}{l|}{Real-world}      & \multicolumn{1}{l|}{Depth + Normal}                 & \multicolumn{1}{l|}{$\sim$4M} & $\sim$1M                \\
				\multicolumn{1}{l|}{Replica~\cite{straub2019replica}}                      & \multicolumn{1}{l|}{Indoor}         & \multicolumn{1}{l|}{Real-world}      & \multicolumn{1}{l|}{Depth + Normal}                 & \multicolumn{1}{l|}{$\sim$150K} & 1 \\
				\multicolumn{1}{l|}{ScanNet\tnote{\dag}~\cite{dai2017scannet}}                      &   \multicolumn{1}{l|}{Indoor}         & \multicolumn{1}{l|}{Real-world}      & \multicolumn{1}{l|}{Depth + Normal}                 & \multicolumn{1}{l|}{$\sim$2.5M} & 1 \\
				\multicolumn{1}{l|}{HM3d~\cite{ramakrishnan2021habitat}}                      & \multicolumn{1}{l|}{Indoor}         & \multicolumn{1}{l|}{Real-world}      & \multicolumn{1}{l|}{Depth + Normal}                 &\multicolumn{1}{l|}{$\sim$2000K}  & 1
				\\
				\multicolumn{1}{l|}{Hypersim~\cite{roberts2021hypersim}}                      & \multicolumn{1}{l|}{Indoor}         & \multicolumn{1}{l|}{Synthesized}      & \multicolumn{1}{l|}{Depth + Normal}                 &\multicolumn{1}{l|}{54K} & 1 				
				\\ \hline
				\multicolumn{5}{c}{Testing Data}                                                                                                                                                           \\ \hline
				  \multicolumn{1}{l|}{NYU~\cite{silberman2012indoor}}                            & \multicolumn{1}{l|}{Indoor}      & \multicolumn{1}{l|}{Real-world}         & \multicolumn{1}{l|}{Depth+Normal}                & \multicolumn{1}{l|}{654} & 1                      \\
				\multicolumn{1}{l|}{KITTI~\cite{Geiger2013IJRR}}                          & \multicolumn{1}{l|}{Outdoor}        & \multicolumn{1}{l|}{Real-world}      & \multicolumn{1}{l|}{Depth}                 & \multicolumn{1}{l|}{652} & 4                   \\
				\multicolumn{1}{l|}{ScanNet\tnote{\dag}~\cite{dai2017scannet}}                        & \multicolumn{1}{l|}{Indoor}         & \multicolumn{1}{l|}{Real-world}      & \multicolumn{1}{l|}{Depth+Normal}                & \multicolumn{1}{l|}{700} & 1                      \\
				\multicolumn{1}{l|}{NuScenes (NS)~\cite{caesar2020nuscenes}}                       & \multicolumn{1}{l|}{Outdoor}        & \multicolumn{1}{l|}{Real-world}      & \multicolumn{1}{l|}{Depth}                 & \multicolumn{1}{l|}{10K} & 6                        \\
				\multicolumn{1}{l|}{ETH3D~\cite{schops2017multi}}                          & \multicolumn{1}{l|}{Outdoor}        & \multicolumn{1}{l|}{Real-world}      & \multicolumn{1}{l|}{Depth}                 & \multicolumn{1}{l|}{431} & 1                        \\
				\multicolumn{1}{l|}{DIODE~\cite{vasiljevic2019diode}}                          & \multicolumn{1}{l|}{In/Out} & \multicolumn{1}{l|}{Real-world}      & \multicolumn{1}{l|}{Depth}                 & \multicolumn{1}{l|}{771} & 1                     \\
				\multicolumn{1}{l|}{iBims-1~\cite{koch2018evaluation}}      & \multicolumn{1}{l|}{Indoor}                        & \multicolumn{1}{l|}{Real-world}         & \multicolumn{1}{l|}{Depth}                & \multicolumn{1}{l|}{100} & 1                        \\
				\toprule[1pt]
		\end{tabular}}
		\begin{tablenotes}
		\scriptsize
		\item[\dag] ScanNet is a non-zero-shot testing dataset for our ViT models.
		\end{tablenotes}
		\vspace{-0.2 em}
	\end{threeparttable}
	\label{table: datasetsv2}
\end{table}
\subsection{Zero-shot Generalization %
}

\noindent\textbf{Qualitative comparisons of surface normals and depths.} We visualize our predictions in Fig. \ref{fig: visualization.}. A comparison with another widely used generalized metric depth method, ZoeDepth \cite{bhat2023zoedepth}, demonstrates that our approach produces depth maps with superior details on fine-grained structures (objects in row1, suspension lamp in row4, beam in row8), and better foreground/background distinction (row 11, 12). In terms of surface normal prediction, our normal maps exhibits significantly finer details compared to Bae. \etal \cite{bae2021estimating} and can handle some cases where their method fail (row7, 8, 9). Our method not only generalizes well across diverse scenarios but can also be directly applied to unseen camera models like the fisheye camera shown in row 12. More visualization results for in-the-wild images are presented in Fig.~\ref{fig: visualization wild.}, including comic-style (Row 2) and CG(computer graphics)-generated objects (Row5)
\vspace{-1em}

\begin{table*}[]
\renewcommand\arraystretch{1.1}
\caption{Quantitative comparison of 3D scene reconstruction with LeReS~\cite{leres}, DPT~\cite{ranftl2021vision}, %
RCVD~\cite{kopf2021rcvd}, 
SC-DepthV2~\cite{bian2021tpami}, and two learning-based MVS methods (DPSNet~\cite{im2019dpsnet}, SimpleRecon~\cite{sayed2022simplerecon}) on 9 unseen NYUv2 scenes. Apart from the MVS approaches and ours, other methods have to align the scale with ground truth depth for each frame. As a result, our reconstructed 3D scenes achieve the best performance.}
\vspace{0.2 em}
\centering
\resizebox{.95\linewidth}{!}{%
  \centering
  \small 
  \setlength{\tabcolsep}{0.5mm}{\begin{tabular}{@{} r |rc|rc|rc|rc|rc|rc|rc|rc|rc@{}}
    \toprule
    \multirow{2}{*}{Method} & \multicolumn{2}{c|}{Basement\_0001a} & \multicolumn{2}{c|}{Bedroom\_0015} & \multicolumn{2}{c|}{Dining\_room\_0004} & \multicolumn{2}{c|}{Kitchen\_0008} & \multicolumn{2}{c|}{Classroom\_0004} & \multicolumn{2}{c|}{Playroom\_0002}  & \multicolumn{2}{c|}{Office\_0024} & \multicolumn{2}{c|}{Office\_0004} & \multicolumn{2}{c}{Dining\_room\_0033}\\
      & C-$l_1$$\downarrow$ & F-score $\uparrow$ & C-$l_1$$\downarrow$ & F-score $\uparrow$ & C-$l_1$$\downarrow$ & F-score $\uparrow$ & C-$l_1$$\downarrow$ & F-score $\uparrow$ & C-$l_1$$\downarrow$ & F-score $\uparrow$ & C-$l_1$$\downarrow$ & F-score $\uparrow$ &
      C-$l_1$$\downarrow$ & F-score $\uparrow$ & C-$l_1$$\downarrow$ & F-score $\uparrow$ & C-$l_1$$\downarrow$ & F-score $\uparrow$\\ \hline
    RCVD~\cite{kopf2021rcvd} & 0.364 & 0.276 
            & 0.074 & 0.582 &
             0.462 & 0.251 &
             0.053 & 0.620 &
              0.187 & 0.327 &
             0.791 & 0.187 &
             0.324 & 0.241  &
             0.646 & 0.217 &
             0.445 & 0.253 \\

    SC-DepthV2~\cite{bian2021tpami}  & 0.254 & 0.275 &
             0.064 & 0.547 &
             0.749 & 0.229 &
             0.049 & 0.624 &
              0.167 & 0.267 &
             0.426 & 0.263 &
             0.482 & 0.138  &
             0.516 & 0.244 &
             0.356 &0.247 \\

    DPSNet~\cite{im2019dpsnet} & 0.243 & 0.299 &
             0.195 & 0.276 &
             0.995 & 0.186 &
             0.269 & 0.203 &
             0.296 & 0.195 &
             0.141 & 0.485 &
             0.199 & 0.362  &
             0.210 & 0.462 &
             0.222 & 0.493 \\
              
    DPT~\cite{leres} & 0.698 & 0.251 &
             0.289 & 0.226 &
             0.396 & 0.364 &
             0.126 & 0.388 &
             0.780 & 0.193 & 
             0.605 & 0.269 &
             0.454 & 0.245  &
             0.364 & 0.279 &
             0.751 & 0.185 \\  
    LeReS~\cite{leres} & 0.081 & 0.555 &
             0.064 & 0.616 &
             0.278 & 0.427 &
             0.147 & 0.289 &
             0.143 & \textbf{0.480} &
             0.145 & 0.503 &
             0.408 & 0.176  &
             0.096 & 0.497 &
             0.241 & 0.325 \\  
    \red{SimpleRecon}~\cite{sayed2022simplerecon} & \red{0.068} & \red{0.695} 
            & \red{0.086} & \red{0.449} &
             \red{0.199} & \red{0.413} &
             \red{0.055} & \red{0.624} &
              \red{\textbf{0.142}} & \red{0.461} &
             \red{0.092} & \red{0.517} &
             \red{\textbf{0.054}} & \red{\textbf{0.638}}  &
             \red{0.051} & \red{0.681} &
             \red{0.165} & \red{0.565} \\               
             \hline
    Ours & \textbf{0.042} & \textbf{0.736} &
             \textbf{0.059} & \textbf{0.610} &
             \textbf{0.159} & \textbf{0.485} &
             \textbf{0.050} & \textbf{0.645} &
             0.145 & 0.445 &
             \textbf{0.036} & \textbf{0.814} &
             0.069 & \textbf{0.638}  &
             \textbf{0.045} & \textbf{0.700} &
             \textbf{0.060} & \textbf{0.663} \\

    \bottomrule
  \end{tabular}}}
  \vspace{-1em}
  \label{tab: NYUD reconstruction cmp.}
\end{table*}

\subsection{Applications Based on Our Method}
We apply the CSTM\_image model to various tasks. 

\begin{figure*}[]
\centering
\includegraphics[width=0.85\textwidth]%
{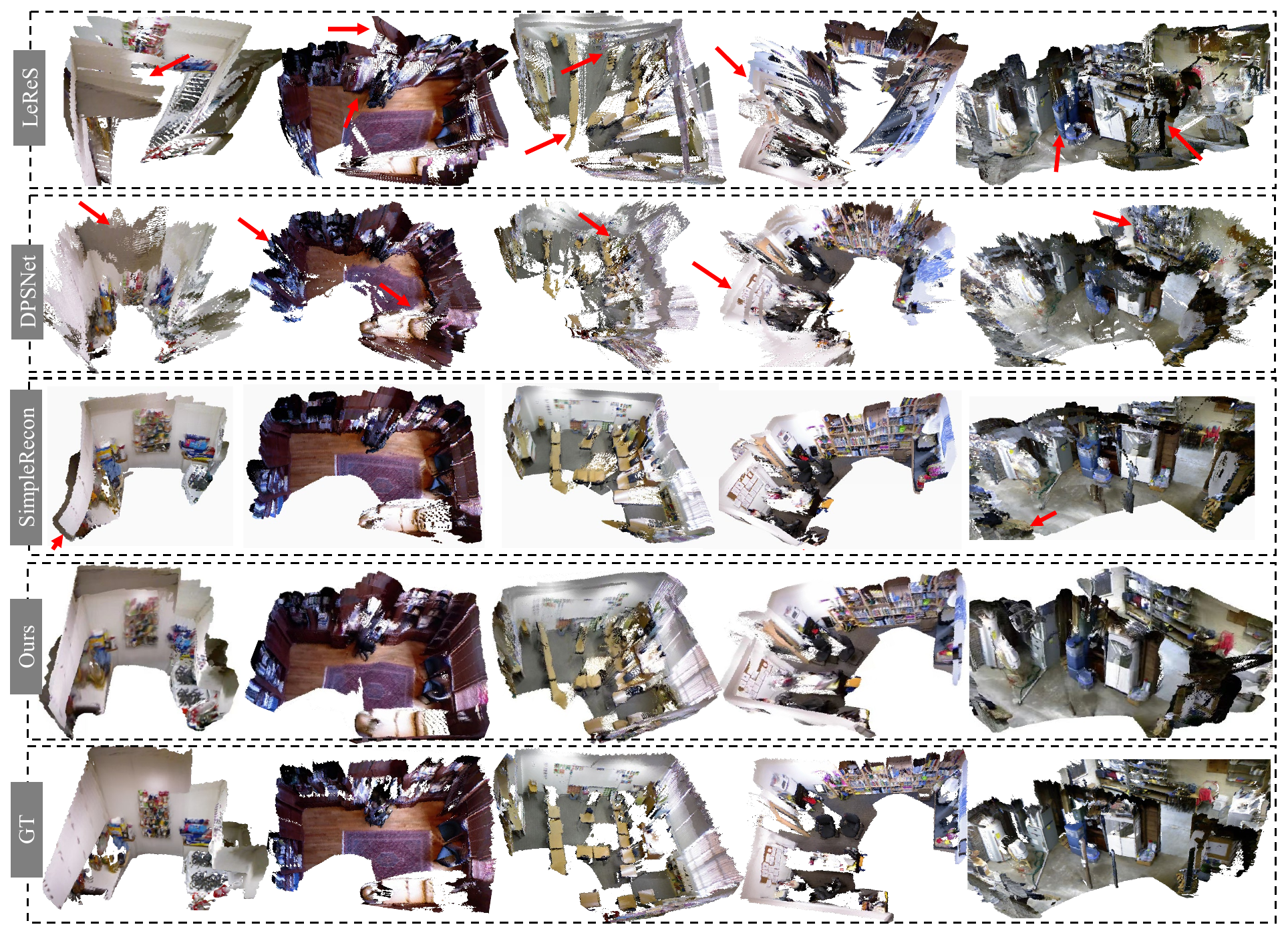}
\vspace{-0.5 em}
\caption{\textbf{Reconstruction of zero-shot scenes with multiple views.} We sample several NYUv2 scenes for 3D reconstruction comparison. As our method can predict accurate metric depth, thus all frame's predictions are fused directly for reconstruction. By contrast, LeReS~\cite{leres}'s depth is up to an unknown scale and shift, causing noticeable distortions. For MVS methods, DPSNet~\cite{im2019dpsnet} fails on low-texture backgrounds, while SimpleRecon~\cite{sayed2022simplerecon} distorts regions without sufficient observations.}
\vspace{-2em}
\label{fig: visual nyud reconstruction cmp.}
\end{figure*}

\noindent\textbf{3D scene reconstruction .}
\blue{
To present our method's ability to recover real-world metric 3D, we first conduct a quantitative comparison on 9 unseen NYUv2 scenes. We predict per-frame metric depth and fuse these with the provided camera poses, with results detailed in Table \ref{tab: NYUD reconstruction cmp.}. We compare our approach to several methods: the video consistent depth prediction method (RCVD~\cite{kopf2021rcvd}), unsupervised video depth estimation (SC-DepthV2~\cite{bian2021tpami}), 3D scene shape recovery (LeReS~\cite{leres}), affine-invariant depth estimation (DPT~\cite{ranftl2021vision}), and multi-view stereo reconstruction (DPSNet~\cite{im2019dpsnet}, SimpleRecon~\cite{sayed2022simplerecon}). Except for the multi-view approaches and our method, all others require aligning scales with ground truth depth for each frame. While our approach is not specifically designed for video or multi-view reconstruction, it demonstrates promising frame consistency and significantly more accurate 3D scene reconstructions in these zero-shot scenarios. Qualitative comparisons in Fig.~\ref{fig: visual nyud reconstruction cmp.} reveal that our reconstructions exhibit considerably less noise and fewer outliers.
}

\noindent\textbf{Dense-SLAM mapping.}
\blue{
Monocular SLAM is a key robotic application that uses a single video input to create trajectories and dense 3D maps. However, due to limited photometric and geometric constraints, existing methods struggle with scale drift in large scenes and fail to recover accurate metric information. Our robust metric depth estimation serves as a strong depth prior for the SLAM system. To demonstrate this, we input our metric depth into the state-of-the-art SLAM system, Droid-SLAM~\cite{teed2021droid}, and evaluate the trajectory on KITTI without any tuning. Results are shown in Table \ref{tab: KITTI SLAM.}. With access to accurate per-frame metric depth, Droid-SLAM experiences a significant reduction in translation drift ($t_{rel}$). Additionally, our depth data enables Droid-SLAM to achieve denser and more precise 3D mapping, as illustrated in Fig.~\ref{Fig: first page fig.} and detailed in the supplementary materials.}

\blue{We also tested on the ETH3D SLAM benchmarks, with results in Table \ref{Tab: ETH3D SLAM}. Using our metric depth predictions, Droid-SLAM shows improved performance, although the gains are less pronounced in the smaller indoor scenes of ETH3D compared to KITTI.
}

\begin{table}[]
\vspace{-0.5em}
\renewcommand\arraystretch{1.1}
\caption{Comparison with SoTA SLAM methods on KITTI. We input predicted metric depth to the Droid-SLAM~\cite{teed2021droid} (`Droid+Ours'), which outperforms others by a large margin on trajectory accuracy.}
\centering
\resizebox{.95\linewidth}{!}{%
  \centering
  \small 
  \setlength{\tabcolsep}{0.5mm}{\begin{tabular}{@{} l |c|c|c|c|c|c|c@{}}
    \toprule
    \multirow{2}{*}{Method} & Seq 00 & Seq 02 & Seq 05 & Seq 06 & Seq 08 & Seq 09 & Seq 10 \\ \cline{2-8} 
      & \multicolumn{7}{c}{Translational RMS drift ($t_{rel}, \downarrow$) / Rotational RMS drift ($r_{rel}, \downarrow$)} \\ \hline
    GeoNet~\cite{yin2018geonet} & 27.6/5.72 & 42.24/6.14 & 
             20.12/7.67 & 
             9.28/4.34 & 
             18.59/7.85 & 
             23.94/9.81 & 
             20.73/9.1  \\
    VISO2-M~\cite{song2015high}  & 12.66/2.73 & 
             9.47/1.19 & 
             15.1/3.65 & 
             6.8/1.93 & 
             14.82/2.52 & 
             3.69/1.25 & 
             21.01/3.26  \\

    ORB-V2~\cite{murORB2} & 11.43/0.58 & 
             10.34/0.26 &
             9.04/0.26 & 
             14.56/0.26 & 
             11.46/0.28 & 
             9.3/0.26 & 
             2.57/0.32    \\
              
    Droid~\cite{teed2021droid} & 33.9/\textbf{0.29} & 
             34.88/\textbf{0.27} & 
             23.4/0.27 & 
             17.2/0.26 & 
             39.6/0.31 & 
             21.7/0.23 & 
             7/0.25   \\  \hline
             
    Droid+Ours & \textbf{1.44}/0.37 & 
             \textbf{2.64}/0.29 & 
             \textbf{1.44}/\textbf{0.25} & 
             \textbf{0.6}/\textbf{0.2} & 
             \textbf{2.2}/\textbf{0.3} & 
             \textbf{1.63}/\textbf{0.22} & 
             \textbf{2.73}/\textbf{0.23}    \\

    \bottomrule
  \end{tabular}}}
  \vspace{-2em}
  \label{tab: KITTI SLAM.}
\end{table}

\begin{table}[]
\vspace{0em}
\caption{
Comparison of VO error on ETH3D benchmark. Droid SLAM system is input with our depth (`Droid + Ours'), and ground-truth depth (`Droid + GT'). The average trajectory error is reported.
}
 \resizebox{.9\linewidth}{!}{%
\begin{tabular}{l|llllll}
\hline
             & Einstein & \multirow{2}{*}{Manquin4} & \multirow{2}{*}{Motion1} & Plant- & sfm\_house & sfm\_lab \\
             & \_global &  &  & scene3 & \_loop & \_room2 \\ \hline
             
& \multicolumn{6}{c}{Average trajectory error ($\downarrow$)}  \\ \hline
Droid        & 4.7                               & 0.88     & 0.83    & 0.78        & 5.64             & 0.55            \\ 
Droid + Ours & 1.5                               & 0.69     & 0.62    & 0.34        & 4.03             & 0.53            \\ 
Droid + GT   & 0.7                               & 0.006    & 0.024   & 0.006       & 0.96             & 0.013           \\ \hline
\end{tabular}}
\label{Tab: ETH3D SLAM}
\vspace{-1 em}
\end{table}

\begin{figure}[!bth]
\centering
\includegraphics[width=0.45\textwidth]{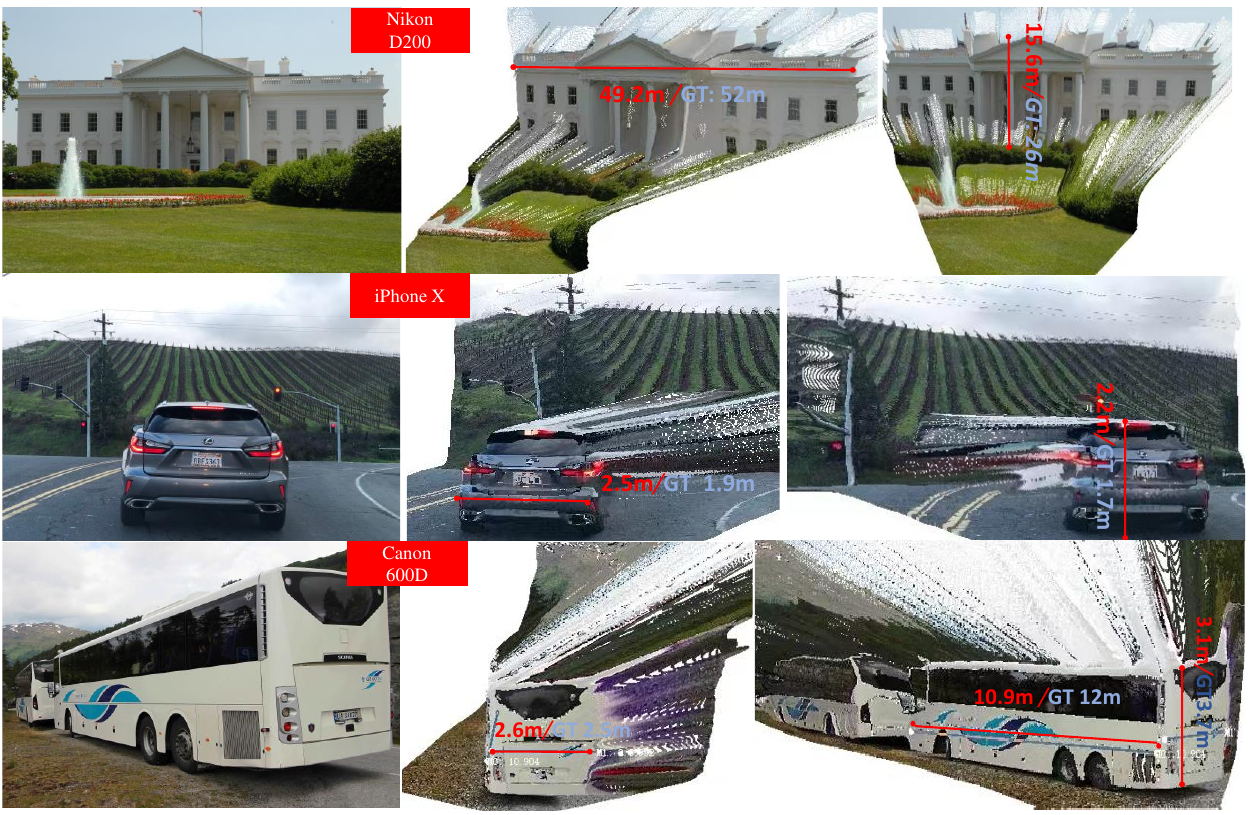}
\vspace{-0.8 em}
\caption{\textbf{Metrology of in-the-wild scenes.} We collect several Flickr photos, which are captured by various cameras. With photos' metadata, we reconstruct the 3D metric shape and measure structures' sizes. Red and blue marks are ours and ground-truth sizes respectively. }
\label{fig: reconstruction in the wild.}
\vspace{-1em}
\end{figure}

\noindent\textbf{Metrology in the wild.} \blue{To demonstrate the robustness and accuracy of our recovered metric 3D shapes, we downloaded Flickr photos taken by various cameras and extracted coarse camera intrinsic parameters from their metadata. We utilized our CSTM\_image model to reconstruct the metric shapes and measure the sizes of structures (marked in red in Fig.~\ref{fig: reconstruction in the wild.}), with ground-truth sizes shown in blue. The results indicate that our measured sizes closely align with the ground-truth values.}

\begin{figure}[!bth]
\centering
\includegraphics[width=0.45\textwidth]{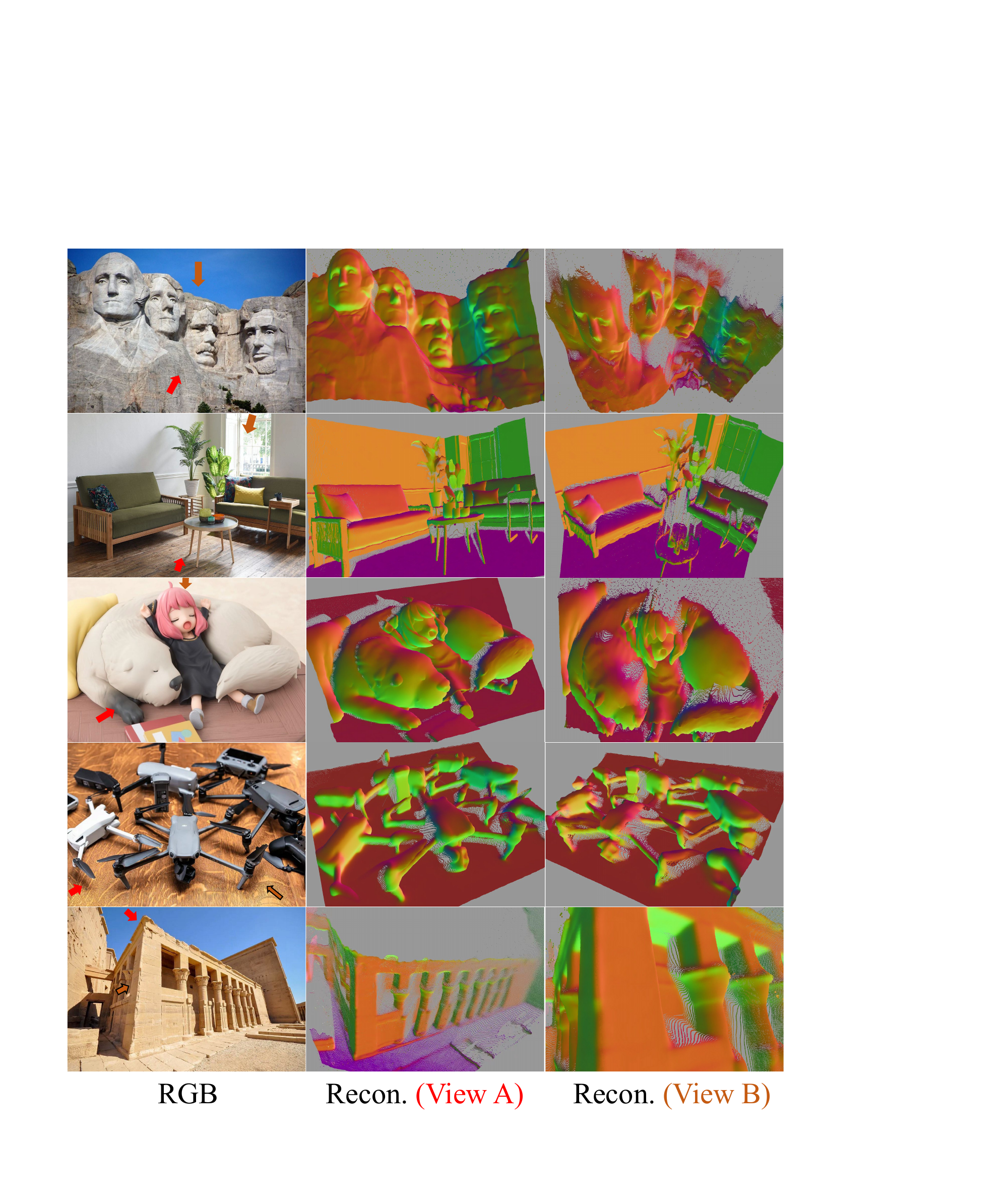}
\vspace{-0.5 em}
\caption{\textbf{Reconstruction from in-the-wild single images.} We collect web images and select proper focal lengths. The reconstructed pointclouds are colorized by normals.}
\label{fig: new reconstruction in the wild.}
\vspace{-2.0em}
\end{figure}

\noindent\textbf{Monocular reconstruction in the wild.} To further visualize the reconstruction quality of our recovered metric depth, we randomly collect images from the internet and recover their metric 3D and normals. As there is no focal length provided, we select proper focal lengths according to the reconstructed shape and normal maps. The reconstructed pointclouds are colorized by their corresponding normals (Different views are marked by red and orange arrays in Fig.~\ref{fig: new reconstruction in the wild.}). \vspace{-1em}

\vspace{-0.5em}
\subsection{Ablation Study}
\noindent\textbf{Ablation on canonical transformation.}
\blue{We examine the impact of our proposed canonical transformations for input images (CSTM\_input) and ground-truth labels (CSTM\_output). Results are presented in Table \ref{table: importance of camera model.}. We trained the model on a mixed dataset of 90,000 images and tested it across six datasets. A naive baseline (Ours w/o CSTM) removes the CSTM modules, enforcing the same supervision as our approach. Without CSTM, the model struggles to converge on mixed metric datasets and fails to achieve metric predictions on zero-shot datasets. This limitation is why recent mixed-data training methods often resort to learning affine-invariant depth to sidestep metric challenges. In contrast, both of our CSTM methods enable the model to attain metric prediction capabilities and achieve comparable performance. Table \ref{table:errors cmp on NYUD-V2} confirms this comparable performance. Thus, adjusting supervision and the appearance of input images during training effectively addresses metric ambiguity issues. Additionally, we compared our approach with CamConvs~\cite{facil2019cam}, which incorporates the camera model in the decoder using a 4-channel feature. While CamConvs uses the same training schedule, model, and data, it relies on the network to implicitly learn various camera models from image appearance, linking image size to real-world dimensions. We believe this approach strains data diversity and network capacity, resulting in lower performance.}

\begin{table}[]
\caption{Effectiveness of our CSTM. CamConvs~\cite{facil2019cam} directly encodes various camera models in the network, while we perform a simple yet effective transformation to solve the metric ambiguity. Without CSTM, the model achieve transferable metric prediction ability.}
\vspace{0.2 em}
\scalebox{0.8}{
\begin{tabular}{l|lll|lll}
\toprule[1pt]
\multirow{2}{*}{Method}        & DDAD & Lyft & DS & NS & KITTI & NYU \\ 
        &\multicolumn{3}{c|}{Test set of train. data (AbsRel$\downarrow$)}     & \multicolumn{3}{c}{Zero-shot test set (AbsRel$\downarrow$)} \\  \hline 
w/o CSTM &$0.530$ &$0.582$  &$0.394$  &$1.00$ & $0.568$      &$0.584$  \\
CamConvs~\cite{facil2019cam}  &$0.295$ &$0.315$  &$0.213$ &$0.423$  &$0.178$      &$0.333$   \\
Ours CSTM\_image &$0.190$ &$0.235$  &$0.182$  &$0.197$ & $0.097$      &$0.210$ \\
Ours CSTM\_label &$0.183$ &$0.221$  &$0.201$  &$0.213$ & $0.081$      &$0.212$  \\
\toprule[1pt]
\end{tabular}}
\label{table: importance of camera model.}
\vspace{-1 em}
\end{table}

\noindent\textbf{Ablation on canonical space.}
\blue{We investigate the impact of the canonical camera, specifically the canonical focal length. The model is trained on a small sampled dataset and evaluated on both the training and validation sets. We calculate the average Absolute Relative (AbsRel) error for three different focal lengths:} \red{250,} \blue{500, 1000, 1500,} \red{and 2500}. \blue{Our experiments indicate that a focal length of 1000 yields slightly better performance than the others; further details can be found in the supplementary materials.} %

\noindent\textbf{Effectiveness of 
the random proposal normalization loss.}
\blue{To demonstrate the effectiveness of our random proposal normalization loss (RPNL), we conducted experiments on a sampled small dataset, with results shown in Table 10. We tested on DDAD, Lyft, DrivingStereo (DS), NuScenes (NS), KITTI, and NYUv2. The 'baseline' includes all losses except RPNL, which we compare to 'baseline + RPNL' and 'baseline + SSIL \cite{Ranftl2020}'. Our RPNL significantly enhances performance, while the scale-shift invariant loss \cite{Ranftl2020}, which normalizes the entire image, offers slight improvements.}

\begin{table}[]
\vspace{-1em}
\caption{Effectiveness of random proposal normalization loss. Baseline is supervised by `$L_{\PWN} + L_{\VNL} + L_{silog}$'. SSIL is the scale-shift invariant loss proposed in ~\cite{Ranftl2020}.}
\vspace{0.2 em}
\scalebox{0.8}{
\begin{tabular}{l|lll|lll}
\toprule[1pt]
\multirow{2}{*}{Method}        & DDAD & Lyft & DS & NS & KITTI & NYUv2 \\ 
        &\multicolumn{3}{c|}{Test set of train. data (AbsRel$\downarrow$)}     & \multicolumn{3}{c}{Zero-shot test set (AbsRel$\downarrow$)} \\  \hline 
baseline  &$0.204$ &$0.251$  &$0.184$  &$0.207$ &$0.104$      &$0.230$     \\
baseline + SSIL~\cite{Ranftl2020} &$0.197$ &$0.263$  &$0.259$  &$0.206$ & $0.105$      &$0.216$     \\
baseline + RPNL   &$\textbf{0.190}$  &$\textbf{0.235}$  &$\textbf{0.182}$  &$\textbf{0.197}$ &$\textbf{0.097}$      &$\textbf{0.210}$     \\  \toprule[1pt]
\end{tabular}}
\label{table: effectiveness of rpnl.}
\vspace{-1.0 em}
\end{table}

\begin{figure}[]
	\centering
	\includegraphics[width=0.45\textwidth]{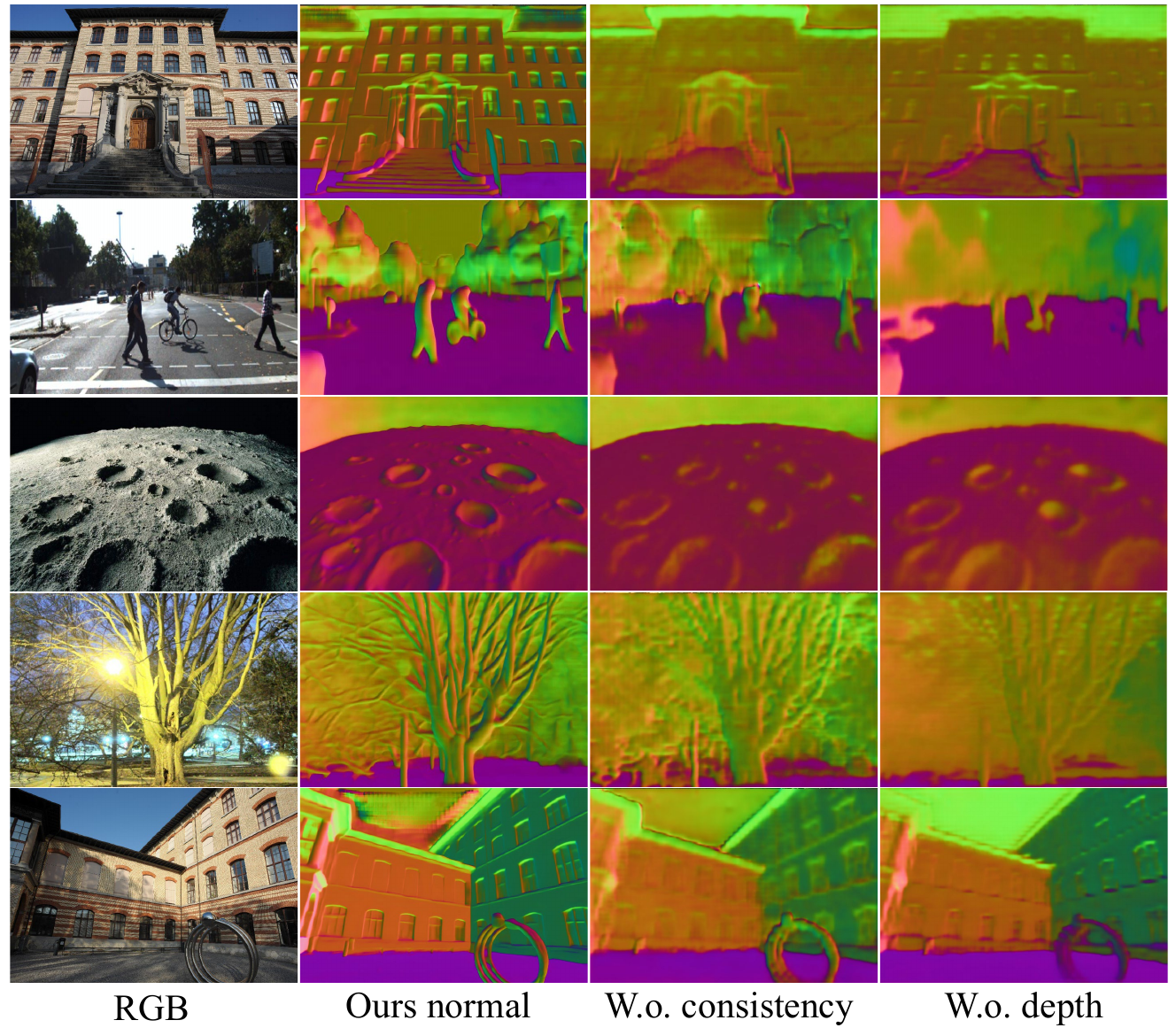}
	\vspace{-0.2 em}
	\caption{Effect of joint depth-normal optimization. We compare normal maps learned by different strategies on several outdoor examples. Learning normal only `without depth' leads to flattened surfaces, since most of the normal labels lie on planes. In addition, `without consistency' imposed between depth and normal, the predictions become much coarser.}
	\label{fig: normal_abl.}
	\vspace{-1em}
\end{figure}

\noindent\textbf{Effectiveness of joint optimization.} We assess the impact of joint optimization on both depth and normal estimation using small datasets sampled with ViT-small models over a 4-step iteration. The evaluation is conducted on the NYU indoor dataset and the DIODE outdoor dataset, both of which include normal labels for the convenience of evaluation. In Tab. \ref{table: effectiveness of joint opt.}, we start by training the same-architecture networks `without depth' or `without normal' prediction. Compared to our joint optimization approach, both single-modality models exhibit slightly worse performance.To further demonstrate the benefit of joint optimization and the incorporation of large-scale outdoor data prior to normal estimation, we train a model using only the Taskonomy dataset (i.e., 'W.o. mixed datasets'), which shows inferior results on DIODE(outdoor). We also verify the effectiveness of the recurrent blocks and the consistency loss. Removing either of them (`W.o. consistency' / `W.o. recurrent block') could lead to drastic performance degradation for normal estimation, particularly for outdoor scenarios like DIODE(Outdoor). Furthermore, we present some visualization comparisons in Fig ~\ref{fig: normal_abl.}. Training surface normal and depth together without the consistency loss ('W.o. consistency') results in notably poorer predicted normals compared to our full method ('Ours normal'). Additionally, if the model learns the normal individually ('W.o. depth'), the performance also degrades. The efficiency analysis of the joint optimization module is presented in the supplementary materials.

\begin{table}[]
\vspace{-1.0em}
	\caption{Effectiveness of joint optimization. Joint optimization surpasses independent estimation. For outdoor normal estimation, this module introduces geometry clues from large-scale depth data. The proposed recurrent block and depth-normal consistency constraint are essential for the optimization}
	\vspace{0.2 em}
	\scalebox{0.8}{
		\begin{tabular}{l|ll|ll}
			\toprule[1pt]
			\multirow{2}{*}{Method}        & DIODE(Outdoor) & NYU & DIODE(Outdoor) & NYU \\ 
			&\multicolumn{2}{c|}{Depth (AbsRel$\downarrow$)}     & \multicolumn{2}{c}{Normal (Med. error$\downarrow$)} \\  \hline 
			W.o. normal &\quad \quad    0.315 &0.119  & \quad-  & -     \\
			W.o. depth    &\quad \quad  -  & -  &\quad16.25  &8.78  \\
			W.o mixed datasets &\quad \quad$0.614$  &$0.116$  &\quad$18.94$  &$9.50$  \\
			W.o. recurrent block   &\quad \quad $0.309$  &$0.127$  &\quad$16.51$  &$9.31$  \\
			W.o. consistency &\quad \quad $0.310$ &$0.121$    &\quad$16.45$  &$9.72$  \\
			Ours  &\quad \quad $\textbf{0.304}$ &$\textbf{0.114}$  &\quad$\textbf{14.91}$  &$\textbf{8.77}$      \\
			\toprule[1pt]
	\end{tabular}}
	\label{table: effectiveness of joint opt.}
	\vspace{-2.0 em}
\end{table}

\noindent\textbf{Selection of intermediate normal representation.} During optimization, unnormalized normal vectors are utilized as the intermediate representation. Here we explore three additional representations (1) A vector defined in $\mathfrak{so3}$ representing 3D rotation upon a reference direction. We implement this vector by lietorch\cite{teed2021droid}. (2) An azimuthal angle and a polar angle. (3) A 2D homogeneous vector \cite{zhao2021confidence}. All the representations investigated are additive and can be surjectively transferred into surface normal. In this experiment, we only change the representations and compare the performances. Surprisingly, according to Table \ref{table: selection of normal.}, the naive unnormalized normal performs the best. We hypothesize that this simplest representation reduces the learning difficulty.

\begin{table}[h]
	\vspace{-1.0 em}
	\caption{More selection of intermediate normal representation. }
	\scalebox{0.8}{
		\begin{tabular}{l|ll|ll}
			\toprule[1pt]
			\multirow{2}{*}{Representation}  & Taskonomy & Scannet  & DIODE(Outdoor) & NYU  \\ 
			&\multicolumn{2}{c|}{Test set (Med. err.$\downarrow$)}     & \multicolumn{2}{c}{Zero-shot testing (Med. err.$\downarrow$)}  \\  \hline 
			3D rotation vector   &\quad  $5.28$ & $8.92$  & \quad $16.00$ & $9.45$ \\
			Azi. and polar angles &\quad    $5.34$ & $9.01$ & \quad $15.21$  & $9.21$     \\
			Homo. 2D vector \cite{zhao2021confidence}  &\quad  $5.02$  &  $8.50$  & \quad ${15.40}$ & $8.79$ \\ 
			Ours (unnormalized 3D)   &\quad  $\textbf{5.01}$  & $\textbf{8.41}$  & \quad $\textbf{14.91}$ & $\textbf{8.77}$ \\ 
			\toprule[1pt]
	\end{tabular}}
	\label{table: selection of normal.}
	\vspace{-0.5 em}
\end{table}

\noindent\textbf{Best optimizing steps} To determine the optimal number of optimization steps for various ViT models, we vary different steps to refine depth and normal. Table \ref{table: step.} illustrates that increasing the number of iteration steps does not consistently improve results. Moreover, the ideal number of steps may differ based on the model size, with larger models generally benefiting from more extensive optimization.

\vspace{-1.6 em}
\begin{table}[h]
	\caption{Select the best joint optimizing steps for different ViT models. We find the best step varying with model size. All models are trained following the settings in Tab.~\ref{table: effectiveness of joint opt.}}
	\vspace{0.0 em}
	\scalebox{0.8}{
		\begin{tabular}{c|ccc}
			\toprule[1pt]
			\multirow{2}{*}{\diagbox{\# Steps}{Backbone}}  & ViT-Small & \quad\quad\quad  ViT-Large  & ViT-giant  \\ 
			& \multicolumn{3}{c}{KITTI Depth (AbsRel$\downarrow$) / NYU v2 Normal (Med. err.$\downarrow$)} \\   \hline 
			2   & {0.102/9.01}  & {0.070/8.40} & {0.069/8.25}  \\
			4 & \textbf{0.088/8.77} & {0.067/8.24}  & {0.067/8.23}       \\
			8   & {0.090/8.80} & \textbf{0.065/8.21}  &  \textbf{0.064/8.22} \\ 
			16  & {0.095/8.79}  &  {0.068/8.30} & {0.065/8.27} \\ 
			\toprule[1pt]
	\end{tabular}}
	\label{table: step.}
	\vspace{-1.8 em}
\end{table}

\section{Conclusion} In this paper, we introduce a family of geometric foundation models for zero-shot monocular metric depth and surface normal estimation. We propose solutions to address challenges in both metric depth estimation and surface normal estimation. To resolve depth ambiguity caused by varying focal lengths, we present a novel canonical camera space transformation method. Additionally, to overcome the scarcity of outdoor normal data labels, we introduce a joint depth-normal optimization framework that leverages knowledge from large-scale depth annotations.

Our approach enables the integration of millions of data samples captured by over 10,000 cameras to train a unified metric-depth and surface-normal model. To enhance the model's robustness, we curate a dataset comprising over 16 million samples for training. Zero-shot evaluations demonstrate the effectiveness and robustness of our method. For downstream applications, our models are capable of reconstructing metric 3D from a single view, enabling metrology on randomly collected internet images and dense mapping of large-scale scenes. With their precision, generalization, and versatility, Metric3D v2 models serve as geometric foundational models for monocular perception.